\documentclass{article}

\usepackage[preprint]{neurips_2026}

\usepackage[utf8]{inputenc} 
\usepackage[T1]{fontenc}    
\usepackage{hyperref}       
\usepackage{url}            
\usepackage{booktabs}       
\usepackage{amsfonts}       
\usepackage{nicefrac}       
\usepackage{microtype}      
\usepackage{xcolor}
\usepackage{pifont}
\newcommand{\cmark}{\textcolor{green!60!black}{\ding{51}}}
\newcommand{\xmark}{\textcolor{red!75!black}{\ding{55}}}
\usepackage{graphicx} 
\usepackage{longtable}
\usepackage{multirow}
\usepackage[table]{xcolor}
\usepackage{amsmath}
\usepackage{float}
\usepackage{arydshln}
\usepackage{graphicx} 
\usepackage{wrapfig}

\newcommand{\iconFrozen}{\includegraphics[height=0.9em]{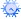}}
\newcommand{\iconTrainable}{\includegraphics[height=0.9em]{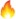}}

\usepackage{titletoc}
\newcommand{\pmstd}[1]{_{\scalebox{0.65}{$\scriptstyle ( #1)$}}}
\title{Parcel2Progression: An Anatomy-aware Longitudinal Framework for Alzheimer’s Disease Diagnosis}

\author{%
  Madhumitha Venkatesh\thanks{Indian Institute of Technology Hyderabad, India}  \hspace{3em}
  Shanawaj S Madarkar\footnotemark[1]  \hspace{3em}
  Konda Reddy Mopuri\footnotemark[1] \\ 
}

\begin{document}

\maketitle

\begin{abstract}
  Alzheimer's disease (AD) progression is a longitudinal process with subtle pathological cues in the early stages. Yet, computational constraints have limited most neuroimaging models to either compromise spatial information or limit the number of longitudinal\footnote{a sequence of temporal scans of a subject collected over time} scans. We aim to overcome this bottleneck and fully leverage high-resolution, variable-length T1w structural MRI (4D sMRI) scan sequences. We introduce \textbf{Parcel2Progression (P2P)}, a Longitudinal Transformer Framework which tackles this challenge using an Atlas-guided\footnote{A brain atlas is a standardized reference map that uses ``parcels" to define distinct anatomical or functional brain regions} \textbf{Parcel Encoder} that tokenizes 3D scans into a set of richer anatomically grounded representations. A \textbf{Longitudinal Transformer} then integrates irregular, arbitrary-length longitudinal visits with patient age. This synergy delivers two key advantages:  (1) parcel-specific interpretability, and (2) computational tractability for long-term analysis, which scales linearly with the number of scans compared to a naive quadratic 4D ViT cost. P2P outperforms prior works and baselines in both MCI (Mild Cognitive Impairment) to AD conversion prediction and AD vs. CN (Cognitively Normal) classification tasks across ADNI, AIBL, and MIRIAD datasets. Leveraging longitudinal scans boosts performance over single-scan baselines by up to $5\%$ and $7\%$ in balanced accuracy for AD classification and MCI conversion prediction tasks, respectively. Interpretability analysis using parcel saliencies and attention rollouts reveals clinically consistent atrophy\footnote{Progressive shrinkage of brain tissues due to widespread loss and death of neurons} patterns in AD and MCI subjects. We also demonstrate the frameworks' reliability in anomaly detection using a synthetic dataset, and test the model's generalizability for other neurodegenerative diseases like Frontotemporal Dementia.
\end{abstract}

\section{Introduction}

\textit{Neurodegeneration is the tale of a mind reshaped, each lost neuron casting ripples. Understanding these subtle waves into a measurable pattern has been a timeless quest.}
Neurodegenerative diseases (NDDs) do not arrive with a single event~\cite{Apostolova2012,Barnes2005,Braak1991,Braak2006}; they advance quietly, re-shaping neural circuits and, over time, the very structure of the brain.
The NIA-AA AT(N)\footnote{The NIA-AA AT(N) is a research framework from the National Institute on Aging and the Alzheimer's Association.} framework definition of Alzheimer’s disease (AD), the most prevalent NDD, groups biomarkers into $\beta$-amyloid (A)~\cite{DeTure2019}, pathologic tau ($\tau$)~\cite{Bateman2012}, and neurodegeneration (N)~\cite{Planche2022,Jack2018}.
In this work, we focus on \emph{N} because neurodegeneration leaves structural fingerprints \textit{i.e.}, regional atrophy, shape change, and tissue loss that structural MRI (sMRI) can directly capture~\cite{Vemuri2010,Whitwell2007,Frisoni2010,Jack1999}. We therefore aim to exploit the diagnostic potential of T1w sMRI, a non-invasive technique, by extracting high-dimensional features without the aid of complementary imaging modalities or clinical scores, given its broader availability in routine clinical practice.

\begin{figure}[t]
  \centering
   \includegraphics[width=\linewidth]{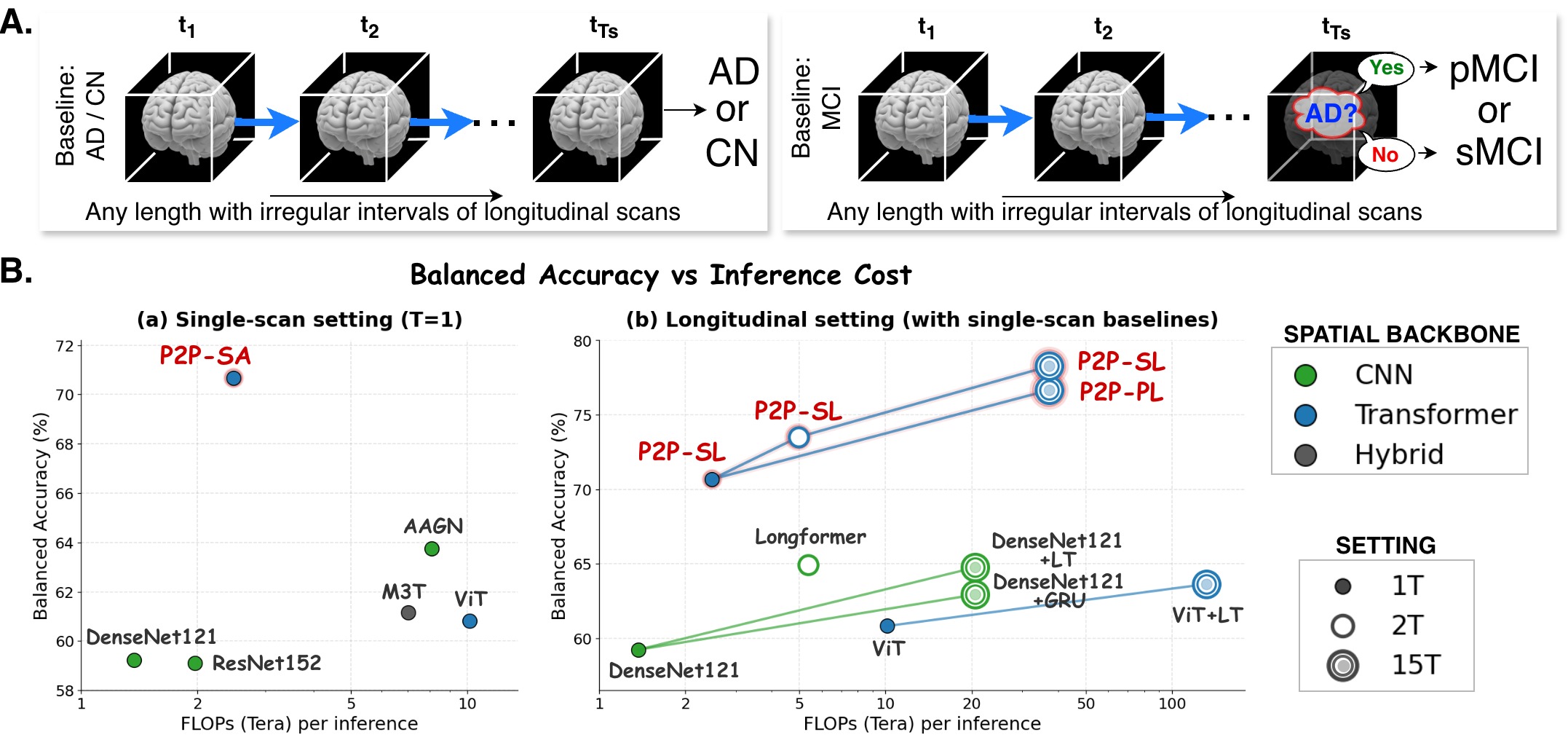}
\vspace{-1em}
    \caption{A) Problem overview: Analyze high-resolution, arbitrary‐length longitudinal T1w sMRI (4D) scans to predict (i) Alzheimer’s disease and (ii) MCI‐to‐AD conversion, while keeping explanations anatomically grounded. B) The plots shows P2P's superiority in the compute vs. performance tradeoff at inference time compared to other SOTA models and baselines on ADNI for the harder task of sMCI vs. pMCI under both single-scan and longitudinal (max T scans) settings.}
   \label{fig:overview}
   \vspace{-1.5em}
\end{figure}

AD follows a characteristic spatial trajectory, beginning in medial temporal lobe structures, such as the hippocampus, before spreading to associated cortices~\cite{jack2013tracking,Braak1991,Ferreira2017,blinkouskaya2021brain,Planche2022,Jellinger2022,McKhann2011,Nestor2008,Rauchmann2025}. In early stages such as Mild Cognitive Impairment (MCI), morphological signatures of stable and progressive MCI subjects (sMCI \& pMCI) frequently overlap in cross-sectional data, making single time-point snapshots insufficient for differentiation~\cite{Zhang2017LandmarkLongitudinal}. Longitudinal data resolves this ambiguity by capturing the spatiotemporal pattern and rate of neurodegeneration~\cite{blinkouskaya2021brain,Ferreira2017,Qiu2022,Apostolova2012,Planche2022,Rauchmann2025}. This temporal context enables the differentiation of stable versus progressive trajectories for robust MCI-to-AD conversion predictions, while increasing diagnostic confidence in AD vs. CN classification by decoupling neurodegeneration from healthy aging.


However, building models that truly exploit longitudinal signal in practice introduces several challenges that must be addressed \emph{simultaneously}, and this coupled problem setting still remains under-explored in the longitudinal sMRI literature.
\textbf{(i) High-resolution} 3D sMRI enables the capture of reliable fine-grained anatomical cues required for faithful reasoning in subtle cases like MCI~\cite{Frisoni2010StructuralMRIAD}. This necessity extends to longitudinal studies, where preserving spatial resolution is crucial for tracking disease progression. 
\textbf{(ii) Variable lengths:} Real-world clinical data is characterized by irregular visit intervals and variable-length patient histories. Longitudinal models must accommodate these inconsistencies to be clinically viable.
\textbf{(iii) Compute cost} of jointly modeling high-resolution 3D structures across multiple visits scales rapidly with both volumetric resolution and sequence length. Consequently, existing approaches typically sacrifice one of these dimensions: either aggressively reducing spatial dimensionality to support temporal reasoning or limiting temporal sequence to maintain spatial fidelity, often necessitating dataset truncation. Such trade-offs inevitably prevent models from simultaneously capturing fine-grained anatomical changes and long-horizon clinical trajectories, undermining its utility in diverse clinical settings.
Finally, AD does not progress uniformly across individuals: cohort studies report distinct atrophy subtypes (e.g., limbic-predominant vs.\ hippocampal-sparing)~\cite{Ferreira2017,Rauchmann2025}, implying that effective models must remain sensitive to \emph{where} degeneration manifests and \emph{how} it evolves over time, rather than relying on undifferentiated whole-brain features from a single visit~\cite{Jellinger2022,blinkouskaya2021brain,jack2013tracking,Planche2022}.
Taken together, these realities motivate models that (i) read structure region-by-region, aligned to clinically coherent anatomy, and (ii) aggregate evidence over arbitrary-length follow-up histories to learn subject-specific progression pathways, supporting both robust AD diagnosis and challenging prognostic MCI tasks like sMCI vs.\ pMCI (Fig.~\ref{fig:overview}).

Current methods treat anatomy-aware model design and longitudinal modelling to capture progression as separate subdomains. We argue that they are two facets of a single \emph{region-specific progression} phenomenon. We introduce \textbf{Parcel2Progression (P2P)}, a longitudinal transformer framework that unifies these perspectives to address the challenges of high-resolution 3D sMRIs and variable-length clinical histories. The plot in Fig~\ref{fig:overview} shows that P2P demonstrates better tradeoff between performance and compute. Our contributions are:
~\label{contributions}

\vspace{-4.5pt}
\begin{enumerate}
\item \textbf{High-dimensional anatomical grounding:} A \emph{Parcel Encoder} encodes high-resolution 3D sMRI scans into atlas-grounded parcel representations. Thus, P2P preserves fine-grained regional specificity and provides a richer spatial context than a standard global embedding.
\vspace{-1pt}
\item \textbf{Computationally efficient temporal scaling:} P2P significantly reduces self-attention overhead by coupling parcel-specific encoding with an age-aware \emph{Longitudinal transformer} through a two-stage approach. This enables modeling of irregular visit intervals while scaling linearly with the number of visits for computational tractability.
\vspace{-1pt}
\item \textbf{Superior diagnostic and prognostic performance:} P2P achieves SOTA performance across ADNI/AIBL/MIRIAD datasets and demonstrates strong generalisation on cross-datasets. Notably, P2P achieves significant longitudinal gains on the challenging MCI-to-AD conversion prediction task, with balanced accuracy improvements up to 8\% \& AUC by 6\%.

\end{enumerate}

\section{Related Works}
Deep learning (DL) for AD diagnosis and MCI's prognostic task from structural T1w MRI has has evolved through complementary paradigms, each addressing key limitations. 


\noindent
\textit{\textbf{3D Single-scan models:}}
Early DL methods moved from 2D slice inputs~\cite{Gao2023AHM} to consuming full 3D scans to preserve volumetric context and spatial continuity. 3D CNNs~\cite{Folego2020WholeBrain3DCNN,Oh2019SciRep,Qiu2020InterpretableDL} were extensively used due to their strong inductive bias under limited data~\cite{Khatri2024AlzheimersDA}, but their interpretations are often clinically opaque and can be driven by shortcut cues. Recently, ViTs~\cite{Kun_Training_MICCAI2024} and CNN-ViT hybrids~\cite{Altay2021PreclinicalAD,zhu_midl,Li2022TransResNet,Wu2022AMSNet} have been explored to better model large global 3D context. Region-discovery approaches like~\cite{Lian2020HFCN,zhu_midl} attempt to localize discriminative patterns via patch-based aggregation, but arbitrary patching can still encourage learning shortcut cues. Multi-plane/ multi-slice tokenization works~\cite{Khatri2024AlzheimersDA} like M3T (2022,\cite{Jang2022}) adopted a hybrid backbone, improving efficiency but compromising true 3D continuity. To incorporate neurobiological grounding, anatomy-aware designs~\cite{RelSA2025,Jiang2024AAGN} inject priors through gating parcel regions or attention biases over predefined regions. Yet, AAGN’s (2024,\cite{Jiang2024AAGN}) reliance on ROI-feature extraction from overlapping CNN feature maps limits true parcel-specific representations. While anatomical grounding is especially valuable for clinically critical pMCI vs. sMCI conversion \cite{Guan2023,Jiang2024AAGN}, these methods remain structurally cross-sectional (single scan), leaving progression modeling and longitudinal validity largely unaddressed.
\noindent
\textit{\textbf{4D Longitudinal modelling:}}
Classical longitudinal pipelines model disease progression via various techniques. Voxel-level pipelines~\cite{davatzikos2009spareAD,Wang2014HMMMRI,huang2017longitudinal} model progression from voxel-wise morphometry, but are context-poor and unstable. ROI-descriptor pipelines~\cite{zhu2021longrange,Li2011,Farzan2015LongitudinalAtrophy}, by contrast, model trajectories from regional summaries, but remain feature-space bound. A separate line of ROI-centric works~\cite{li2019hippocampal,sarasua2022hippocampal,Olaoluwa_2026_WACV} (focused on specific structures e.g  hippocampus) perform shape-based longitudinal learning, yet they remain tool-dependent, narrow the hypothesis space, and do not generalize naturally to whole-brain multi-region progression modeling. Recent DL models increasingly aim to avoid these brittle preprocessing dependencies by learning task-relevant representations through a data-driven approach~\cite{Oh2019SciRep}.
Methods like~\cite{cui2019rnn,Jomeiri2024,agh2020pami} often compress each 3D visit into a single scan-level embedding using CNNs and then aggregate histories via RNNs (typically with fixed-interval assumptions~\cite{cui2019rnn} or capped histories~\cite{Jomeiri2024}), collapsing regional structure and weakening sensitivity to subtle, localized progression. For computational tractability, these CNN+RNN methods and techniques performing self-/weakly-supervised normative learning~\cite{Ouyang2023LSOR,zhao2021lssl,ouyang2021lne,ouyang2022disentangle} further rely on aggressive input downsampling (64×64×64) or curated 2D slice-selection~\cite{Hu2023VGGTswinformer}, compromising fine-grained anatomical detail. Recently, Longformer (2024)~\cite{chen2024longformer}  improved sensitivity to spatial cues by modeling current and prior scan (@1.75mm resolution) differences via optical flow, but is restricted to max 2 visits and incurs substantial inference overhead. Similarly, AFSNet’s design requires fixed scan pairs, thus truncating dataset (dropping scans without baseline visits)\cite{Liu2025AttentionGuided3DCNN}. 

In summary, existing methods typically limit longitudinal modeling by compromising spatial fidelity, or restricting temporal context (detailed comparison in Table~\ref{tab:p2p_baseline_comparison}).
P2P, a spatio-temporal transformer, mitigates this by learning a set of parcel-granular anatomical embeddings from high-resolution (1mm; $182\times218\times182$) scans, and couples them with age-aware,variable-length temporal learning. This enables anatomically grounded region-wise progression modelling whilst being computationally efficient (Fig~\ref{fig:overview}B).

\begin{figure*}[t]
  \centering
   \includegraphics[width=\linewidth]{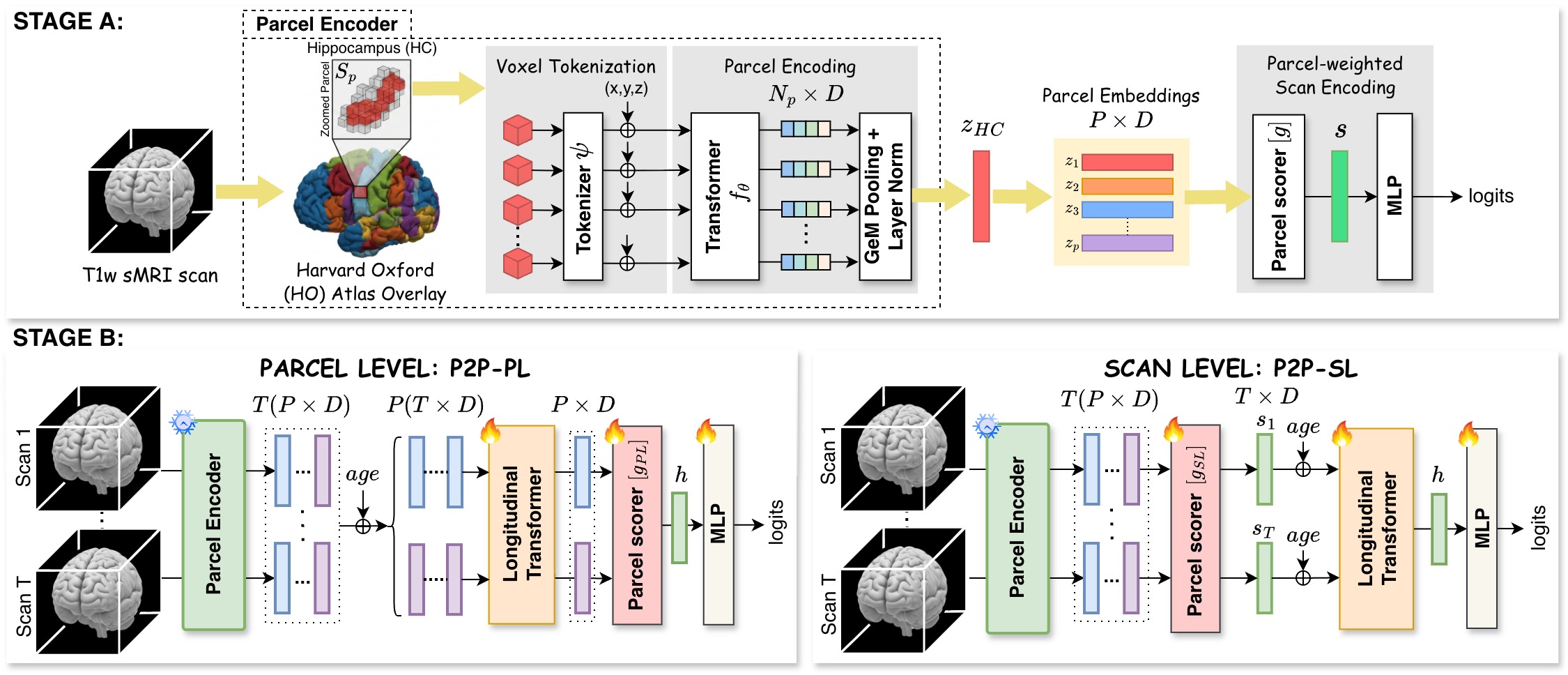}
   \vspace{-1em}
   \caption{\textbf{P2P framework.} Stage A (P2P-SA): An Atlas-guided Parcel Encoder framework for single-scan predictions by encoding a scan into parcel embeddings (with Ex. Hippocampus) using the Harvard-Oxford brain atlas. Stage B: Age-aware Longitudinal Transformer variants, P2P-PL (parcel-level) \& P2P-SL (scan-level), for longitudinal classification. [\iconFrozen{}: Frozen, \iconTrainable{}: Trainable parameters]}
   \label{fig:spatial_trans}
\end{figure*}


\section{Parcel2Progression (P2P) Framework}
We propose a two-stage pipeline for longitudinal (4D) T1-weighted sMRI classification.
Stage~A learns to encode anatomically anchored parcel embeddings from a single high-resolution 3D scan via an \emph{Atlas-guided Parcel Encoder}.
Stage~B then performs evidence aggregation over longitudinal histories using an \emph{Age-aware Longitudinal Transformer} operating on frozen parcel embeddings.
This decoupling at train-time preserves anatomical resolution while enabling scalable sequence modeling.

\subsection{Stage A (P2P-SA): Atlas-guided Parcel Encoding of single-scan}

\paragraph{\textbf{Parcel Encoder:}}
\noindent Each preprocessed scan is partitioned into a set of $N$ non-overlapping cubic patches of edge length $v$. A shared tokenizer $\psi$ (e.g., an MLP or DenseNet) projects each patch into a $D$-dimensional embedding, $\psi: \mathbb{R}^{v \times v \times v} \to \mathbb{R}^D$. For a given scan, the resulting patch embeddings are $X = [x_1, \dots, x_N]^\top \in \mathbb{R}^{N \times D}$, where $x_i \in \mathbb{R}^D$ is the embedding for the $i$-th patch.
To preserve 3D structural context, spatial information is injected via 3D coordinates ($c_i \in \mathbb{R}^3$) using an MLP $\phi_s(\mathbb{R}^3\!\to\!\mathbb{R}^D$): 
$
\bar{x}_i = x_i + \phi_s(c_i)$. Then, $ \bar{X} = [\bar{x}_1, \dots, \bar{x}_N]^\top.$

We register the Harvard–Oxford probabilistic atlas~\cite{SCR_001476} to each scan and obtain soft parcel memberships for $P$ anatomical parcels (here, $P{=}69$). Let $M \in \mathbb{R}_{\ge 0}^{N\times P}$ be the membership matrix with entries $M_{ip}$. For \emph{brain} tokens, the soft assignments satisfy:
$\sum_{p=1}^{P} M_{ip}=1$ and $M_{ip}\ge 0$,
whereas \emph{background} tokens satisfy $\sum_{p=1}^{P} M_{ip}=0$. Using $M$, \emph{parcel token sets} are formed by selecting tokens with non-zero membership. For each parcel $p\in\{1,..,P\}$, the sets are defined as:
$
\mathbf{S}_p = \big\{\, \bar{x}_i : M_{ip} > 0 \,\big\},\; 
\text{where} \; N_p = |\mathbf{S}_p|.
$
Equivalently, stacking tokens yields a variable-size matrix per parcel (as parcels differ in volume/shape),
$
\mathbf{S}_p = 
\begin{bmatrix}
\bar{x}_{i_1} \hdots \bar{x}_{i_{N_p}}
\end{bmatrix}^\top
\in \mathbb{R}^{N_p \times D},
$ where $\{i_1,..,i_{N_p}\}$ are the indices with $M_{i_k p}>0$. Due to soft assignments, a token may appear in multiple $\mathbf{S}_p$ (overlapping parcel sets), improving robustness to boundary uncertainty.

To avoid cross-parcel information leakage and the ensuing collapse, encoding via self-attention is performed \emph{independently within each parcel set} using a shared Transformer encoder $f_\theta$~\cite{Dosovitskiy2020}:
\vspace{-0.7em}
\[
\hat{X}^{(p)} \;=\; f_\theta\!\big(\mathbf{S}_p\big) \;=\; 
\begin{bmatrix}
\hat{x}^{(p)}_{1} \hdots \hat{x}^{(p)}_{N_p}
\end{bmatrix}^\top
\in\mathbb{R}^{N_p\times D}
\]
This yields parcel-specific contextualization while remaining computationally efficient relative to global attention over all $N$ tokens.
The tokens are then aggregated within parcel $p$ via Generalized Mean (GeM) pooling~\cite{Gem} to emphasize consistently strong responses followed by LayerNorm:

\vspace{-1em}
\[
u_p =\; \Big(\tfrac{1}{N_p}\!\sum_{k=1}^{N_p} \big(\hat{x}^{(p)}_{k}\big)^{r}\Big)^{\!1/r},
\quad
z_p =\; \operatorname{LN}(u_p)\in\mathbb{R}^{D}
\]
where $r\!\ge\!1$ (here $r=2$, applied component-wise). This produces a single embedding per parcel, independent of its volume and shape, and prevents dilution of small critical structures (e.g., amygdala).

\noindent
\textit{\textbf{Parcel scorer:}} A learnable \emph{Parcel Scorer} assigns diagnostic importance scores to parcels, facilitating interpretability by quantifying a parcel's relative regional contribution to the model's output. Aggregating these parcel embeddings via a convex combination prioritizes pathological regions, yielding a discriminative scan-level embedding with rich anatomical context.
A shared MLP $g(\cdot)$ maps each parcel embedding ($z_p$) to a scalar score $\mathfrak{s}_p=g(z_p)$. To prevent over-reliance on a small subset of regions, parcel-dropout is introduced as a regularizer with rate $\rho\in[0,1)$ during training: sample $d_p\sim\text{Bernoulli}(\rho)$ and set $k_p=1-d_p$. Then, softmax is applied over parcels using temperature-scaled scores with $\tau>0$:

\vspace{-1.3em}
\[
\alpha_p \;=\; \dfrac{\exp\!\big(\tilde{\mathfrak{s}}_p/\tau\big)}{\sum_{q=1}^{P}\exp\!\big(\tilde{\mathfrak{s}}_q/\tau\big)}, 
\qquad
\tilde{\mathfrak{s}}_p =
\begin{cases}
\mathfrak{s}_p, & k_p=1,\\
-\infty, & k_p=0,
\end{cases}
,\qquad 
\sum_{p=1}^{P}\alpha_p=1,\ \alpha_p\ge 0.
\]
 
\noindent
The scan-level embedding, $s$ is convex combination of parcel embeddings,
$
s=\sum_{p=1}^{P}\alpha_p\,z_p,\; s\in \mathbb{R}^D.$Finally, an MLP head returns logits:$\hat{y}=\operatorname{softmax}\!\big(MLP(s)\big)$. P2P-SA is trained end-to-end with cross-entropy loss, $\mathcal{L}_{\text{CE}}(y,\hat{y})$ for per-scan predictions. Fig~\ref{fig:spatial_trans} illustrates the P2P-SA framework.

\vspace{-0.2em}
\paragraph{Why group \emph{before} attention?} Avoid ``collapse'' due to parcel information leakage.
Grouping tokens \emph{prior} to attention constraints information mixing within each parcel set, ensuring the resulting parcel embeddings remain identifiable and meaningful. Refer Appendix Sec~\ref {Asec:collapse} for a detailed justification and Table~\ref{tab:combined_adni_ablations} for collapse scenario results.

\subsection{Stage B: Age-aware Longitudinal Transformer for longitudinal classification}
Stage~B is designed for heterogeneous longitudinal histories, where subjects have variable length scans and irregular acquisition times. Let a subject have $T_s$ visits with corresponding ages $\{A_s(1),\ldots,A_s(T_s)\}$.
At a prediction time $t$, the model consumes the available history up to $t$. To reduce memory footprint from high-dimensional 3D inputs and enable training over longer histories, Stage~B leverages the pre-trained Stage~A Parcel encoder to precompute and operate on \emph{frozen} parcel embeddings $\{z_{p,t}\in\mathbb{R}^D\}$ for each visit. This parcel-level representation enables temporal learning to model longitudinal neurodegeneration at the level of spatially localized anatomical regions. This preserves region-specific progression patterns and facilitates richer spatio-temporal learning, rather than relying on compressed scan-level summaries that may obscure such detail. Visit ages of subject are injected as additive temporal embeddings via an MLP, $\phi_t:\mathbb{R}^1\!\to\!\mathbb{R}^D$. We design two variants for the Longitudinal Transformer framework: i) \emph{parcel-level} variant and ii) \emph{scan-level} variant, and are illustrated in Fig~\ref{fig:spatial_trans}. Both variants are trained with cross-entropy as the loss function.

\noindent
\textit{\textbf{Parcel level (P2P-PL)}: Per-parcel longitudinal modeling with late fusion.}
This variant is designed to learn the disease-progression dynamics \emph{per} anatomical region followed by parcel selection after temporal aggregation. For each parcel $p$, a time series of $\{z_{p,1},\ldots,z_{p,T}\}$ is formed and encoded with a shared Longitudinal transformer $\mathcal{T}_{PL}(\cdot)$ to obtain a parcel trajectory summary $v_p$:
\[
\tilde{z}_{p,t} = z_{p,t} + \phi_t(A_s(t)),
\quad
v_p = \mathcal{T}_{PL}\big(\tilde{z}_{p,1:T_s}\big)\in\mathbb{R}^{D}\]
Parcels are then scored \emph{after} longitudinal aggregation (by mean pooling) using a shared MLP parcel scorer $g_{PL}$, followed by softmax to obtain: $
\beta_p = \text{softmax}\big(g_{PL}(v_p)\big)$. A convex combination aggregates these embeddings to get the complete progression summary:
$
h \;=\; \sum_{p=1}^{P}\beta_p\,v_p \;\in\mathbb{R}^{D}. 
$
Finally, logits are predicted as: $\hat{y}_{PL}=\text{softmax}\big(\text{MLP}(h)\big)$.


\noindent
\textit{\textbf{Scan level (P2P-SL)}: Longitudinal modeling over scan embeddings.}
This variant first aggregates parcels within each visit to obtain a discriminative scan embedding, then models progression across visits. This makes it preferable for long histories or in the case of constrained compute. At visit $t$, a shared MLP parcel scorer, $g_{SL}$ is used to obtain: $\beta_{p,t}= \text{softmax}\big(g_{SL}(z_{p,t})\big)$ and the weighted scan embedding, $s_t= \sum_{p=1}^{P}\beta_{p,t}\,z_{p,t} \;\in\mathbb{R}^{D}$. After adding age embeddings, the sequence $(\tilde{s}_1,\ldots,\tilde{s}_T)$ is encoded using a Longitudinal transformer $\mathcal{T}_{SL}(\cdot)$:
\[
\tilde{s}_t = s_t + \phi_t(A_s(t)), 
\quad
h = \mathcal{T}_{SL}\big(\tilde{s}_{1:T_s}\big)\in\mathbb{R}^{D}
\]
The output $h$ (after mean pooling of tokens) is passed to an MLP classifier which maps it to logits: $\hat{y}_{SL}=\text{softmax}\big(\text{MLP}(h)\big)$.

\subsection{Efficiency Analysis}
As a spatio-temporal reference, a 4D ViT with joint attention over all $T \times N$ tokens serves as the logically consistent baseline to quantify the computational bottlenecks of global attention that P2P explicitly targets.
Consider $N$ total tokens per scan, $N_b$ brain tokens, $T$ visits, $P>1$ atlas parcels with per–parcel sizes $\{N_p\}_{p=1}^P$. Then, the dominant \emph{self-attention} complexity is:

\vspace{-0.5em}
\begin{itemize}
\item Naïve 4D ViT: for all tokens = $O\!\big(T^2 N^2\big)$ \& for only brain tokens = $O\!\big(T^2 N_b^2\big)$
\item P2P-SA (single scan): Parcel grouping yields:
$O\!\Big(\textstyle\sum_{p=1}^{P} N_p^2\Big)
\;<\;
O(N_b^2).$
\item P2P (Stage A + Stage B):
$
O\!\big(T\cdot \textstyle\sum_{p=1}^{P} N_p^2\big)\;+\;
\begin{cases}
O(PT^2) & \text{P2P-PL},\\
O(T^2) & \text{P2P-$\text{SL}$}.
\end{cases}
$
\end{itemize}

Overall, P2P replaces quadratic temporal scaling of naïve 4D attention with a spatially factored form with dominant term as:
$\mathbf{O\!\big(T\cdot \sum_{p=1}^{P} N_p^2\big)}$,
making the temporal cost \emph{linear} in $T$. Consequently, the efficiency advantage \emph{increases linearly with sequence length}.
Under the HO atlas configuration ($P{=}69$; $N_b$$\approx$$6{,}300$; $N{\approx}14{,}812$), parcel grouping reduces per-scan spatial self-attention cost from $O(N_b^{2})$ to $\sum_{p=1}^{P} N_p^{2}$, yielding $\sim$$70\%$ self-attention compute reduction relative to brain-only ViT ($\sim$$5\times$ fewer ops). 


\section{Experiments and Results}
\label{sec:exp}

\subsection{Datasets and Preprocessing}
We utilize 3 \textit{longitudinal T1w} sMRI datasets: \textbf{ADNI}~\cite{Jack2008} from all the phases 1/GO/2/3/4 (10,935 scans with AD:493, pMCI:291, sMCI:396, CN:1210 subjects), \textbf{AIBL}~\cite{Fowler2021AIBL} (1,084 scans with AD: 104; CN:500 subjects), and \textbf{MIRIAD}~\cite{Malone2013} (708 scans with AD: 46, CN:23 subjects). To test the generalizability to other Neurodegenerative diseases, we use the \textbf{NIFD}~\cite{Knopman2014FTLDNI} dataset (962 scans with FTD:214, CN:118 subjects). These datasets collectively span broad variability in age, sex, acquisition site, field strength (1.5T/3T), and scanner manufacturer. All scans were preprocessed and registered to MNI152 (1\,mm isotropic resolution)~\cite{Mazziotta2001} template space using a unified preprocessing pipeline. Finally, padded to get a common grid of $184\times224\times184$ voxels. For the AD/CN task, all scans of subjects with AD or CN baseline diagnoses were utilized. For conversion prediction, pre-conversion scans from MCI subjects who convert to Dementia over their longitudinal visits are labeled pMCI; otherwise, sMCI. Further cohort statistics, conversion time labels and full pipeline specifics are discussed in Appendix Sec.~\ref{Asec:dataset} and Table~\ref{tab:cohort_adni_stats}. To rule out registration-driven confounding, we performed patch-level parcel coverage QC and robustness to registration analysis, discussed in Appendix Sec~\ref{supp:registration_qc},~\ref{supp:robustness}. 

\paragraph{Controlled synthetic dataset.}
To directly evaluate whether P2P can identify the anatomical regions responsible for an abnormal prediction, we construct a controlled 3D Phantom dataset with known parcel-level ground truth. Each $96^3$ volume contains 16 spatially separated parcels represented by simple spherical or cuboidal structures, and is divided into non-overlapping $8^3$ patches following the same patch-based formulation used by P2P for brain MRI. We generate 10,000 samples, balanced between control (CN; $n=5{,}000$) and abnormal (ATR; $n=5{,}000$) cases, simulating single scan cases only. CN samples preserve the underlying parcel structure with only weak nuisance-level variability, whereas ATR samples contain strong synthetic alterations restricted to a small subset of parcels. Since the location and magnitude of every induced alteration are known by construction, this dataset provides explicit parcel-level ground truth for testing whether the representations and importance estimates learned by P2P localize the regions that actually drive the abnormality (Fig.~\ref{fig:synthetic_control}A). The dataset specifics are discussed in detail in Appendix Sec.~\ref{supp:syth_data}.


\begin{table*}[t]
\caption{Prior-work and baseline comparison across ADNI, AIBL, MIRIAD and NIFD. Values are mean$_{\scriptscriptstyle \text{(std)}}$ over 5 folds. Best results are \textbf{bold} for single \& paired and  $\mathbf{\mathcolor{red}{\text{best}}}$ \& $\underline{\text{second best}}$ for longitudinal inputs. (LT.: Longitudinal Transformer, Inf.: Inferencing w/ ADNI ckpt, FT: Finetuning w/ ADNI ckpt, S: single scan, 2T: pair of scans, nT: all longitudinal scans)}
\centering
\resizebox{\textwidth}{!}{
\setlength{\tabcolsep}{1.1pt}
\begin{tabular}{lcccccccccc}
\toprule
\multirow{2}{*}{\textbf{Model}} 
& \multicolumn{2}{c}{\textbf{ADNI [AD/CN]}}
& \multicolumn{2}{c}{\textbf{ADNI [sMCI/pMCI]}}
& \multicolumn{2}{c}{\textbf{AIBL [AD/CN]}}
& \multicolumn{2}{c}{\textbf{MIRIAD [AD/CN]}}
& \multicolumn{2}{c}{\textbf{NIFD [FTD/CN]}}
\\
\cmidrule(lr){2-3}
\cmidrule(lr){4-5}
\cmidrule(lr){6-7}
\cmidrule(lr){8-9}
\cmidrule(lr){10-11}

& \textbf{BAcc(\%)} & \textbf{AUC(\%)}
& \textbf{BAcc(\%)} & \textbf{AUC(\%)}
& \textbf{BAcc(\%)} & \textbf{AUC(\%)}
& \textbf{BAcc(\%)} & \textbf{AUC(\%)} 
& \textbf{BAcc(\%)} & \textbf{AUC(\%)} \\
\midrule

\textit{\textbf{Single scan}}
& & & & & & & & & & \\

3D ResNet-152~\cite{He2016} &  $77.12\pmstd{3.13}$ & $85.98\pmstd{4.33}$
& $59.10\pmstd{2.61}$ & $64.26\pmstd{1.32}$
& $64.21\pmstd{2.20}$ & $79.42\pmstd{3.06}$
& $83.73\pmstd{4.24}$ & $90.26\pmstd{4.12}$ & $64.71\pmstd{3.43}$ & $75.14\pmstd{2.27}$\\

3D DenseNet-121~\cite{huang2017densely} & $77.37\pmstd{2.68}$ & $85.80\pmstd{3.54}$
& $59.23\pmstd{3.96}$ & $65.02\pmstd{4.86}$
& $64.89\pmstd{3.74}$ & $79.34\pmstd{3.09}$
& $85.13\pmstd{4.43}$ & $90.54\pmstd{3.50}$ & $76.86\pmstd{2.80}$ & $83.51\pmstd{3.86}$ \\

3D ViT~\cite{Dosovitskiy2020} & $78.06\pmstd{3.56}$ & $83.63\pmstd{2.61}$
& $60.82\pmstd{2.80}$ & $64.70\pmstd{4.93}$
& $63.26\pmstd{2.58}$ & $78.26\pmstd{3.00}$
& $82.29\pmstd{4.84}$ & $84.52\pmstd{4.29}$ & $52.96\pmstd{1.98}$ & $60.68\pmstd{7.12}$ \\

M3T~\cite{Jang2022} 
& $80.72\pmstd{1.16}$ & $87.57\pmstd{3.85}$
& $61.15\pmstd{3.20}$ & $67.13\pmstd{2.45}$
& $66.50\pmstd{3.86}$ & $83.06\pmstd{3.35}$
& $83.02\pmstd{2.28}$ & $91.03\pmstd{5.35}$ & $78.94\pmstd{3.22}$ & $84.09\pmstd{3.25}$ \\




AAGN~\cite{Jiang2024AAGN} 
& $81.84\pmstd{1.71}$ & $89.80\pmstd{1.20}$
& $63.75\pmstd{2.64}$ & $72.75\pmstd{2.67}$
& $65.34\pmstd{2.50}$ & $79.49\pmstd{3.84}$
& $84.06\pmstd{4.92}$ & $86.52\pmstd{4.50}$ & $80.81\pmstd{3.28}$ & $85.94\pmstd{1.74}$\\


\rowcolor{blue!8} \textbf{P2P-SA (ours)} 
& $\mathbf{84.21\pmstd{3.84}}$ & $\mathbf{91.18\pmstd{2.86}}$
& $66.47\pmstd{3.14}$ & $78.89\pmstd{2.06}$
& $68.18\pmstd{0.90}$ & $79.57\pmstd{4.95}$
& $87.26\pmstd{4.31}$ & $86.96\pmstd{4.29}$ & $80.98\pmstd{3.12}$ & $86.86\pmstd{3.45}$ \\

\rowcolor{blue!8} \textbf{P2P-SA} (Inf.) 
& $\text{--}$ & $\text{--}$
& $\text{--}$ & $\text{--}$
& $82.62\pmstd{4.46}$ & $91.22\pmstd{3.86}$
& $89.66\pmstd{4.18}$ & $97.08\pmstd{3.35}$ & $\text{--}$ & $\text{--}$\\

\rowcolor{blue!8} \textbf{P2P-SA} (FT)
& $\text{--}$ & $\text{--}$
& $\mathbf{70.67\pmstd{2.13}}$ & $\mathbf{80.19\pmstd{2.09}}$
& $\mathbf{84.03\pmstd{2.21}}$ & $\mathbf{93.55\pmstd{1.60}}$
& $\mathbf{92.90\pmstd{4.76}}$ & $\mathbf{97.53\pmstd{4.89}}$ & 
$\mathbf{81.95\pmstd{2.83}}$ & $\mathbf{87.16\pmstd{3.24}}$\\

\rowcolor{black!4}
\textcolor{blue}{\textit{FT gains over P2P-SA}}
& \textit{--} & \textit{--}
& \textcolor{blue}{\textit{\(\uparrow\,4.20\)}} & \textcolor{blue}{\textit{\(\uparrow\,1.30\)}}
& \textcolor{blue}{\textit{\(\uparrow\,1.41\)}} & \textcolor{blue}{\textit{\(\uparrow\,3.24\)}}
& \textcolor{blue}{\textit{\(\uparrow\,3.24\)}} & \textcolor{blue}{\textit{\(\uparrow\,0.45\)}} & \textcolor{blue}{\textit{\(\uparrow\,0.07\)}} & \textcolor{blue}{\textit{\(\uparrow\,0.30\)}}\\

\midrule

\textit{\textbf{Paired scans}}
& & & & & & & & & & \\

Longformer~\cite{chen2024longformer} 
& $82.84\pmstd{2.61}$ & $91.20\pmstd{3.32}$
& $64.91\pmstd{3.06}$ & $77.13\pmstd{2.54}$
& $67.47\pmstd{3.19}$ & $84.71\pmstd{3.46}$
& $85.57\pmstd{3.32}$ & $90.07\pmstd{3.38}$ &
$80.17\pmstd{3.36}$ & $86.14\pmstd{3.17}$ \\

\rowcolor{blue!8} \textbf{P2P-SL (ours)} 
& $\mathbf{86.12\pmstd{3.44}}$ & $\mathbf{93.06\pmstd{2.58}}$
& $\mathbf{73.50\pmstd{2.29}}$ & $\mathbf{81.53\pmstd{1.55}}$
& $\mathbf{85.69\pmstd{3.99}}$ & $\mathbf{88.29\pmstd{4.24}}$
& $\mathbf{95.50\pmstd{3.54}}$ & $\mathbf{96.64\pmstd{4.19}}$ & $\mathbf{84.98\pmstd{2.34}}$ & $\mathbf{88.73\pmstd{2.47}}$ \\

\midrule

\multicolumn{10}{l}{
\textit{\textbf{Any length histories with any interval}}}
 \\

3D DenseNet+GRU~\cite{dey2017gate_variants_gru} 
& $79.37\pmstd{3.64}$ & $87.64\pmstd{2.03}$
& $62.93\pmstd{1.51}$ & $75.57\pmstd{1.06}$
& $65.04\pmstd{3.87}$ & $80.26\pmstd{2.35}$
& $83.82\pmstd{3.71}$ & $89.86\pmstd{2.20}$ & $78.19\pmstd{2.54}$ & $85.62\pmstd{3.03}$ \\

3D DenseNet+LT. 
& $79.12\pmstd{3.55}$ & $87.62\pmstd{2.17}$
& $64.75\pmstd{2.43}$ & $73.04\pmstd{1.82}$
& $65.77\pmstd{4.03}$ & $81.86\pmstd{2.86}$
& $84.77\pmstd{3.66}$ & $91.64\pmstd{3.01}$ & $79.19\pmstd{3.14}$ & $84.91\pmstd{2.93}$ \\

3D ViT+LT. 
& $80.97\pmstd{2.49}$ & $87.98\pmstd{1.55}$
& $63.63\pmstd{2.26}$ & $73.31\pmstd{1.09}$
& $64.26\pmstd{2.57}$ & $81.66\pmstd{2.16}$
& $85.55\pmstd{5.75}$ & $92.58\pmstd{2.04}$ & $78.01\pmstd{3.45}$ & $82.60\pmstd{3.83}$ \\


\rowcolor{blue!8} \textbf{P2P-SL} (Inf.)
& $\text{--}$ & $\text{--}$
& $\text{--}$ & $\text{--}$
& $82.75\pmstd{3.07}$ & $90.58\pmstd{2.18}$
& $90.83\pmstd{4.19}$ & $93.42\pmstd{3.62}$ & $\text{--}$ & $\text{--}$ \\

\rowcolor{blue!8} \textbf{P2P-SL} (e2e) 
& $\underline{88.78\pmstd{1.98}}$ & $\mathbf{\mathcolor{red}{95.47\pmstd{1.94}}}$
& $72.66\pmstd{2.28}$ & $82.31\pmstd{2.72}$
& $85.60\pmstd{2.98}$ & $\underline{91.62\pmstd{3.21}}$
& $96.02\pmstd{2.60}$ & $97.53\pmstd{1.74}$ & $84.13\pmstd{3.04}$ & $89.70\pmstd{3.01}$\\

\rowcolor{blue!8} \textbf{P2P-$\textbf{SL}$} 
& $\mathbf{\mathcolor{red}{89.20\pmstd{2.21}}}$ & $\underline{94.84\pmstd{1.66}}$
& $\mathbf{\mathcolor{red}{78.26\pmstd{0.91}}}$ & $\mathbf{\mathcolor{red}{85.96\pmstd{1.18}}}$
& $\underline{86.86\pmstd{3.47}}$ & $90.89\pmstd{4.69}$
& $\mathbf{\mathcolor{red}{96.90\pmstd{3.55}}}$ & $\underline{98.05\pmstd{3.22}}$ & $\mathbf{\mathcolor{red}{85.30\pmstd{2.58}}}$ & $\mathbf{\mathcolor{red}{90.01\pmstd{2.02}}}$\\

\rowcolor{blue!8} \textbf{P2P-$\textbf{PL}$} 
& $88.27\pmstd{3.80}$ & $94.58\pmstd{1.57}$
& $\underline{76.65\pmstd{1.90}}$ & $\underline{85.86\pmstd{0.60}}$
& $\mathbf{\mathcolor{red}{88.38\pmstd{3.09}}}$ & $\mathbf{\mathcolor{red}{92.21\pmstd{3.06}}}$
& $\underline{96.54\pmstd{3.58}}$ & $\mathbf{\mathcolor{red}{98.33\pmstd{3.00}}}$ & $\underline{84.50\pmstd{3.31}}$ & $\underline{89.63\pmstd{2.57}}$\\

\rowcolor{black!4}
\textcolor{blue}{\textit{Best longitudinal gains}}
& \textcolor{blue}{\textit{\(\uparrow\,4.99\)}} & \textcolor{blue}{\textit{\(\uparrow\,4.29\)}}
& \textcolor{blue}{\textit{\(\uparrow\,7.59\)}} & \textcolor{blue}{\textit{\(\uparrow\,5.77\)}}
& \textcolor{blue}{\textit{\(\uparrow\,4.35\)}} & \textcolor{blue}{\textit{\(\downarrow\,1.34\)}}
& \textcolor{blue}{\textit{\(\uparrow\,4.04\)}} & \textcolor{blue}{\textit{\(\uparrow\,1.13\)}} & 
\textcolor{blue}{\textit{\(\uparrow\,3.35\)}} & 
\textcolor{blue}{\textit{\(\uparrow\,2.85\)}} \\

\bottomrule
\end{tabular}
}
\vspace{-1.0em}
\label{tab:sota_comparison_adni_split}
\end{table*}

\paragraph{Experimental Protocol:}
\label{Experimental Protocol}
We follow diagnosis-stratified, 5-fold cross-validation with \emph{subject-wise} split per fold to prevent any leakage and 10\% val split. We report $mean_{\pm std}$ results of Balanced Accuracy (BACC), and AUC on the $20\%$ held-out test set within each fold. We evaluate AD vs.\ CN on ADNI/AIBL/MIRIAD, and study MCI-to-AD conversion prediction (sMCI vs.\ pMCI) on ADNI, which provides richer longitudinal MCI coverage. We further evaluate FTD vs.\ CN for other disease generalizability and ATR vs.\ CN for the synthetic phantom dataset. All experiments were conducted on 4 NVIDIA RTX A6000s GPUs with Intel CPU (1TB RAM). 
Notably, \textit{P2P exhibits architectural flexibility}, as its parameter count (shared weights) is invariant to number of parcels($P$) and visits($T$), enabling scalability to variable-length visits. Further training details are in Appendix Sec~\ref{Asec:train_details}.

\subsection{Prior works and Baselines}
To contextualize performance under a unified protocol, we retrain SOTA architectures and representative baselines that maintain the flexibility to utilize the full longitudinal dataset without data exclusion/ truncation unlike~\cite{Liu2025AttentionGuided3DCNN,Ouyang2023LSOR} or constrain resolutions unlike~\cite{zhao2021lssl,ouyang2021lne,Ouyang2023LSOR,ouyang2022disentangle}. We evaluate them using the same 5-fold splits as P2P, ensuring a comprehensive and unbiased comparison. We compare P2P against conventional methods widely used for AD classification and MCI prognosis tasks~\cite{Folego2020WholeBrain3DCNN,Korolev2017Residual,Ruiz2020DenseNet,Zhang2021DenseConnWiseAttn,Lu2025RanComViT} like 3D ResNet-152~\cite{He2016}, 3D DenseNet-121~\cite{huang2017densely} \& 3D ViT~\cite{Dosovitskiy2020}. We design longitudinal baselines~\cite{cui2019rnn,agh2020pami,Jomeiri2024} by pairing 3D encoders with GRU or P2P's Longitudinal Transformer (LT), along with visit age as temporal encoding (DenseNet121+GRU, DenseNet121+LT, ViT+LT). We ensure these longitudinal baselines consume irregular-interval, variable-length inputs, similar to P2P. We implemented single-scan SOTA methods such as 
M3T~\cite{Jang2022} and AAGN~\cite{Jiang2024AAGN} to compare against P2P-SA. We also implemented longitudinal SOTA work like Longformer~\cite{chen2024longformer}, which does not restrict the dataset size. To better evaluate performance against Longformer, we provide P2P-SL results using the same input settings by considering only current and prior scans (Table~\ref{tab:sota_comparison_adni_split}). A detailed comparison of these baselines and prior works are presented in Appendix Table~\ref{tab:p2p_baseline_comparison}.

\subsection{P2P Model}

\textit{\textbf{Stage A \& B experiments:}}
P2P-SA (Stage A) utilizes an MLP tokenizer with a ViT-Base architecture for the Parcel Encoder. 
Stage~B employs a lightweight longitudinal module comprising 2 Transformer blocks. We report results of P2P-SA and both the longitudinal variants (P2P-PL and P2P-SL) on ADNI, AIBL, and MIRIAD for both AD vs. CN \& sMCI vs. pMCI tasks in Table~\ref{tab:sota_comparison_adni_split}. P2P-SA outperforms all single-scan prior works; P2P-SL \& P2P-PL further outperform all longitudinal baselines. For completeness, we additionally report an end-to-end scan-level variant (e2e) using batch size 1. The e2e model achieves performance comparable to the proposed two-stage strategy. This suggests that the frozen two-stage formulation preserves most of the spatiotemporal predictive benefit while being substantially more compute-efficient and easier to optimize than full joint end-to-end training. 
A detailed analysis for confounders like manufacturer, sex, age and field strength is presented in Appendix Sec~\ref{Asec:confounders}.
We present additional parcel-level variant and report results of replacing the MLP tokenizer with a 3D DenseNet-121 in Appendix Sec~\ref{Asec:long_variantB} and Table~\ref{tab:dataset_detailed_metrics_models_tok_tasks}, which also shows consistently higher performance.



\textit{\textbf{Cross-cohort generalization:}} A key challenge in neuroimaging is distribution shift across cohorts (scanner/site/protocol variations). To test this, we assess P2P's generalization capabilities on cross-datasets (AIBL/ MIRIAD) by using the best ADNI checkpoint in two modes: (i) inference-only and (ii) fine-tuning. The results in Table~\ref{tab:sota_comparison_adni_split} indicate P2P-SA and P2P-SL generalize effectively. Thus, P2P captures cohort-stable anatomical signals while remaining adaptable to target-domain differences through fine-tuning.

\begin{wrapfigure}{r}{0.48\textwidth}
  \centering
  \vspace{-1.5em}  
  \includegraphics[width=0.48\textwidth]{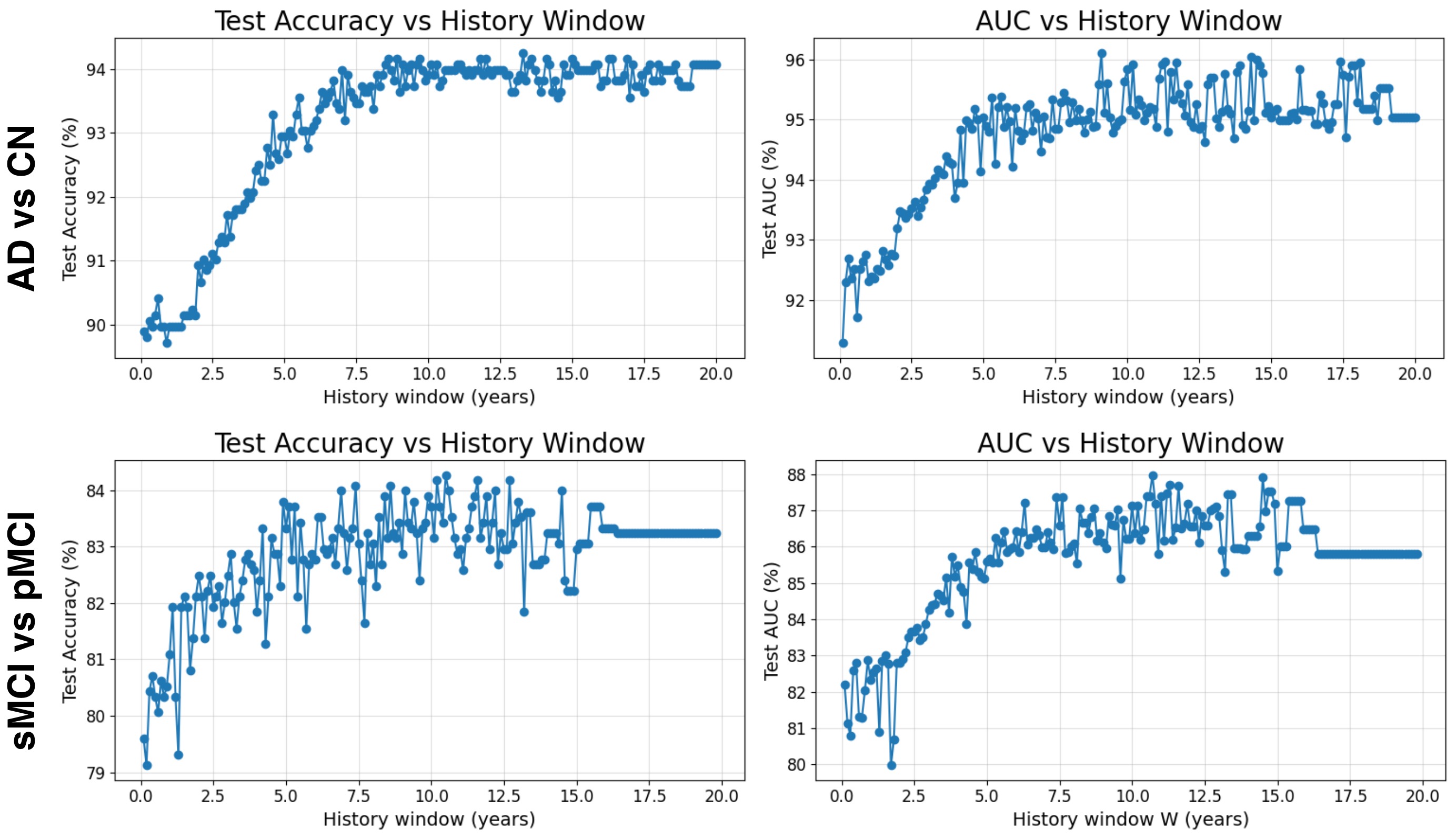}
  \vspace{-1.5em}
  \caption{\textbf{Performance improvements with longitudinal data} using P2P-SL on the ADNI dataset as a function of the history window (0$\rightarrow$20y). Gains start saturating from the 5-year history window, indicating that shorter histories are sufficient for improved performance.}
  \label{fig:long_gain}
  \vspace{-1.5em}
\end{wrapfigure}
\textit{\textbf{Longitudinal gains on AD diagnosis and MCI prognosis:}}
Incorporating longitudinal context yields consistent gains over the best single-scan P2P-SA configuration across all datasets and tasks as reported in Table~\ref{tab:sota_comparison_adni_split}, and Fig~\ref{fig:long_gain}. Notably, P2P-SL achieves $\sim$$7\%$ BACC improvement for the MCI-conversion prediction task. 
Also, a maximum of $\sim$$4\%$ BACC improvements can be noted across ADNI/AIBL/MIRIAD datasets for AD vs. CN task. 
We also observe complementary strengths of the two variants. 
P2P-SL is particularly effective when longer histories are available (e.g., ADNI/MIRIAD), since scan-level sequences provide a compact and stable trajectory representation. 
P2P-PL can be favorable when per-parcel dynamics are critical, and histories are shorter (e.g., AIBL).

\begin{wrapfigure}{r}{0.49\textwidth}
  \centering
  \vspace{-1.3em}
  \includegraphics[width=0.49\textwidth]{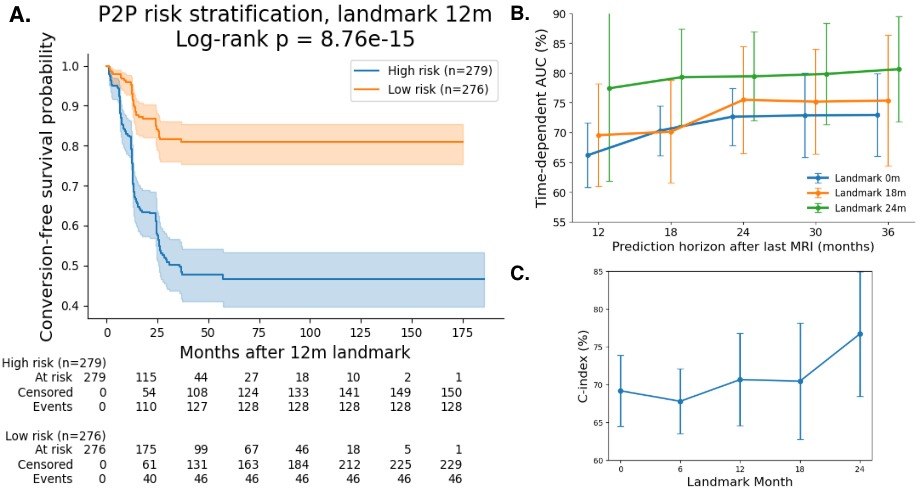}
  \vspace{-1.3em}
  \caption{\textbf{A.} Kaplan-Meier risk stratification~\cite{kaplan1958nonparametric}. \textbf{B.} Time-dependent AUCs across varying landmarks for increasing prediction horizons and \textbf{C.} C-index changes with landmark months.}
  \label{fig:risk_plot}
  \vspace{-1.5em}
\end{wrapfigure}

\textit{\textbf{MCI prognosis and risk stratification:}}
Beyond binary sMCI vs.\ pMCI classification, we fit a Cox proportional hazards model over frozen 
P2P parcel embeddings for MCI-to-AD conversion, treating 
observed conversions as events and subjects without observed 
conversion as right-censored cases (details in Appendix Sec~\ref{Asec:survival}). P2P effectively stratifies MCI subjects into distinct future conversion-risk groups, with significant separation in conversion-free survival (Fig.~\ref{fig:risk_plot}). Notably, meaningful prognostic discrimination is observed at early landmarks, while the improved AUC and C-index~\cite{harrell1982evaluating} at later landmarks indicate that longer MRI histories refine risk estimation by capturing richer progression dynamics.

\begin{wraptable}{r}{0.5\linewidth}
\vspace{-1.9em}
\centering
\caption{Performance using the Neuromorphometrics atlas}
\label{tab:Other_atlas}
\resizebox{\linewidth}{!}{
\begin{tabular}{lcccc}
\toprule
\multirow{2}{*}{\textbf{Model}} 
& \multicolumn{2}{c}{\textbf{ADNI [AD/CN]}}
& \multicolumn{2}{c}{\textbf{ADNI [sMCI/pMCI]}}
\\
\cmidrule(lr){2-3}
\cmidrule(lr){4-5}

& \textbf{BAcc(\%)} & \textbf{AUC(\%)}
& \textbf{BAcc(\%)} & \textbf{AUC(\%)} \\
\midrule
3D DenseNet+GRU
& $79.37\pmstd{3.64}$
& $87.64\pmstd{2.03}$
& $62.93\pmstd{1.51}$
& $75.57\pmstd{1.06}$ \\
3D DenseNet+LT
& $79.12\pmstd{3.55}$
& $87.62\pmstd{2.17}$
& $64.75\pmstd{2.43}$
& $73.04\pmstd{1.82}$ \\
3D ViT+LT
& $80.97\pmstd{2.49}$
& $87.98\pmstd{1.55}$
& $63.63\pmstd{2.26}$
& $73.31\pmstd{1.09}$ \\
\rowcolor{blue!8} P2P-SA
& $80.23\pmstd{2.38}$
& $84.14\pmstd{2.17}$
& $62.11\pmstd{3.45}$
& $75.67\pmstd{3.45}$ \\
\rowcolor{blue!8} P2P-SL
& $\mathbf{84.56}\pmstd{3.83}$
& $\mathbf{89.35}\pmstd{3.24}$
& $\mathbf{73.02}\pmstd{3.07}$
& $\mathbf{81.04}\pmstd{3.66}$ \\
\rowcolor{blue!8} P2P-PL
& $83.98\pmstd{3.04}$
& $89.04\pmstd{3.12}$
& $72.24\pmstd{3.33}$
& $80.08\pmstd{3.51}$ \\

\rowcolor{black!4} Max gain over P2P-SA
& $\mathbf{+4.33}$
& $\mathbf{+5.21}$
& $\mathbf{+10.13}$
& $\mathbf{+5.37}$ \\
\bottomrule
\vspace{-1em}
\end{tabular}
}
\end{wraptable}
\textit{\textbf{Other Atlas:}}
P2P’s framework is atlas-compatible: any atlas can be incorporated by replacing the parcel-membership matrix while retaining parcel encoder and longitudinal modules. However, atlas selection in sMRI cannot be arbitrary. Atlases differ in granularity and, more importantly, in the anatomical, disease-relevant structures they cover. For AD, cortical regions, hippocampal and parahippocampal areas, amygdala, and ventricles are clinically relevant. Therefore, replacing HO with atlases such as AAL3 or Schaefer, which do not provide the same coverage of all relevant structures, including the ventricles, and then comparing task performance would be unfair. We evaluate P2P with an anatomically comparable atlas based on coverage using the Neuromorphometrics atlas, which covers 136 parcels, and report the results in Table~\ref{tab:Other_atlas}.

Introduction of the Neuromorphometric atlas shows that: (i) The longitudinal gains over P2P-SA are preserved and consistent for P2P-SL and P2P-PL. (ii) The prognostic gain is larger under Neuromorphometrics than under Harvard-Oxford, indicating that the benefit of parcel-resolved longitudinal modeling is not an artifact of the HO atlas. (iii) P2P-SL and P2P-PL continue to outperform all longitudinal baselines under the new atlas, and preserve the ranking of methods under a different parcellation.
These gains are non-trivial and can be directly attributed to the P2P framework’s design. We note that absolute performance is lower than under the Harvard-Oxford atlas. We attribute this to granularity: a 136-parcel partition yields fewer patches per parcel and more parcel-specific representations to fit from the same cohort, so reliable representation learning at this resolution likely requires more training data.

\subsection{Ablations}


\begin{wraptable}{r}{0.55\textwidth}
\vspace{-2em}
\caption{Ablation study across Stage~A (single scan) and Stage~B (longitudinal modeling) for ADNI on AD vs. CN task. Values are mean$\pmstd{\text{std}}$ over 5 folds.}
\centering
\scriptsize
\setlength{\tabcolsep}{1pt}
\renewcommand{\arraystretch}{1}
\begin{tabular}{clcc}
\toprule
\textbf{Stage} & \textbf{Model Configuration} & \textbf{BAcc (\%)} & \textbf{AUC (\%)} \\
\midrule

\multirow{9}{*}{A}
& 3D ViT with CLS
& $78.06\pmstd{3.56}$ & $83.63\pmstd{2.61}$ \\

& 3D ViT with Mean token pool
& $78.54\pmstd{1.51}$ & $84.62\pmstd{2.58}$ \\

& 3D ViT with late grouping + Parcel scorer
& $53.23\pmstd{1.20}$ & $65.82\pmstd{2.35}$ \\



& PE with parcel CLS + Mean parcel
& $73.54\pmstd{2.89}$ & $78.70\pmstd{5.63}$ \\

& PE with Gem pool (p=1) + Mean parcel
& $74.41\pmstd{3.01}$ & $77.34\pmstd{5.25}$ \\

& PE with Gem pool (p=2) + Mean parcel
& $76.01\pmstd{2.56}$ & $81.43\pmstd{3.27}$ \\

& PE with Gem pool (p=1) + Parcel scorer
& $81.42\pmstd{4.32}$ & $89.85\pmstd{3.00}$ \\

& \shortstack[l]{PE with Gem pool (p=2) + Parcel scorer\\[-2pt] without Parcel dropout ($\rho=0.00$)}
& $\underline{83.15\pmstd{2.14}}$ & $\underline{90.16\pmstd{2.15}}$ \\

& \cellcolor{blue!8} \shortstack[l]{\textbf{PE with Gem pool (p=2) + Parcel scorer}\\[-2pt] \textbf{with Parcel dropout ($\mathbf{\rho=0.10}$)}}
& \cellcolor{blue!8} $\mathbf{84.21\pmstd{3.88}}$ & \cellcolor{blue!8} $\mathbf{91.18\pmstd{2.86}}$ \\

\midrule

\multirow{9}{*}{B}
& 3D ViT + Longitudinal Transformer
& $80.97\pmstd{2.49}$ & $87.98\pmstd{1.55}$ \\

& P2P-$\text{SL}$ + GRU
& $88.65\pmstd{2.17}$ & $94.19\pmstd{1.90}$ \\

& P2P-$\text{SL}$ + Transformer without age
& $86.46\pmstd{2.61}$ & $90.15\pmstd{2.55}$ \\

& P2P-$\text{SL}$ + Transformer with relative age
& $\underline{88.98\pmstd{3.42}}$ & $94.20\pmstd{2.36}$ \\

& P2P-$\text{SL}$ + Transformer with \texttt{CLS} token
& $88.31\pmstd{2.93}$ & $94.09\pmstd{2.34}$ \\

& P2P-$\text{SL}$ + Transformer (b=1) + mean-pool
& $87.95\pmstd{2.23}$ & $93.61\pmstd{2.44}$ \\

& \cellcolor{blue!8} \textbf{P2P-$\text{SL}$ + Transformer (b=2) + mean-pool}
& \cellcolor{blue!8} $\mathbf{89.20\pmstd{2.21}}$ & \cellcolor{blue!8} $\mathbf{94.84\pmstd{1.66}}$ \\

& P2P-$\text{SL}$ + Transformer (b=3) + mean-pool
& $88.05\pmstd{2.49}$ & $94.04\pmstd{2.05}$ \\

& P2P-$\text{PL}$ + GRU
& $87.10\pmstd{2.94}$ & $93.42\pmstd{2.77}$ \\

& \cellcolor{blue!8} P2P-$\text{PL}$ + Longitudinal Transformer
& \cellcolor{blue!8} $88.27\pmstd{3.80}$ & \cellcolor{blue!8} $\underline{94.58\pmstd{1.57}}$ \\

\bottomrule
\end{tabular}
\vspace{18em}
\label{tab:combined_adni_ablations}
\end{wraptable}
\textbf{Stage~A Ablations (single--scan):}
\begin{wrapfigure}{r}{0.55\textwidth}
  \centering
  \vspace{-15em}
  \includegraphics[width=0.55\textwidth]{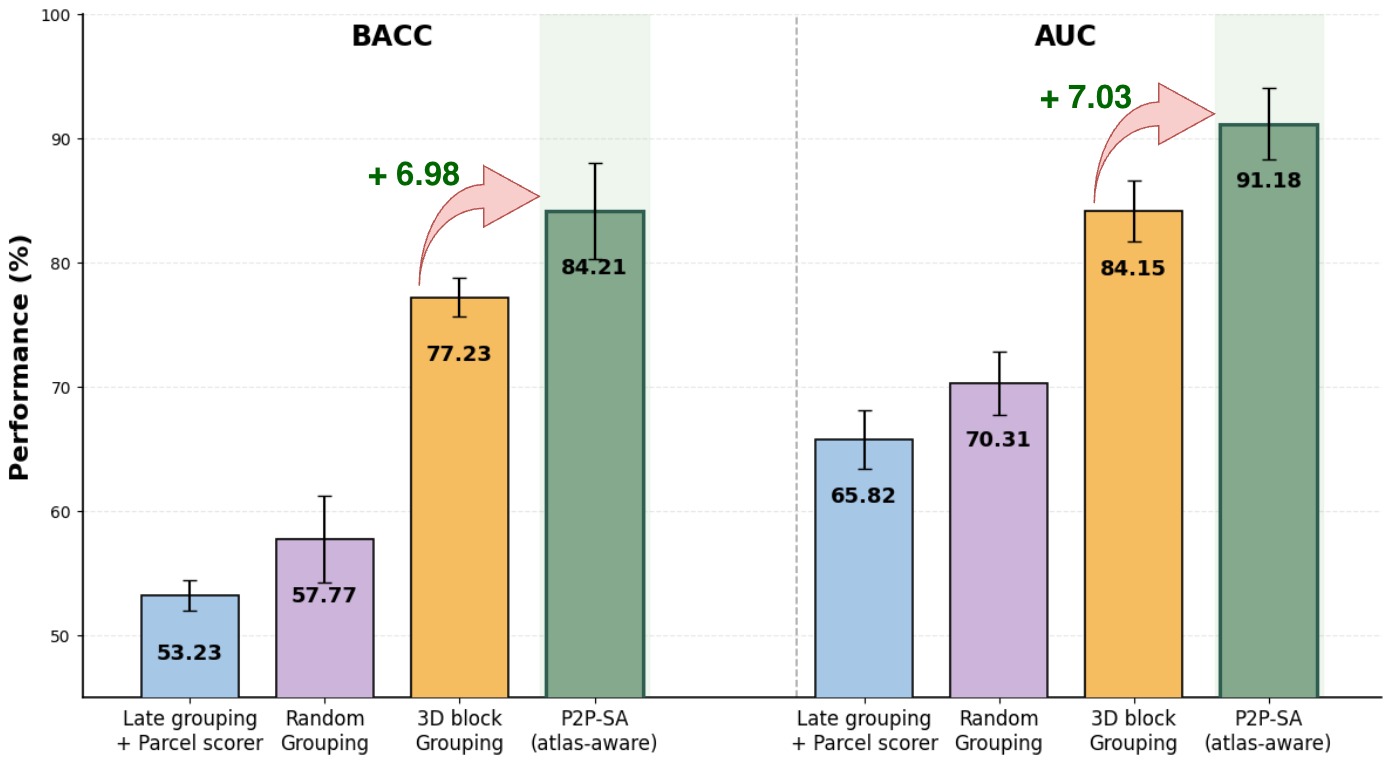}
  \vspace{-1.5em}
  \caption{\textbf{Atlas-aware grouping of P2P-SA} consistently achieves the best BACC and AUC, showing that anatomically meaningful grouping is more effective than late, random, or 3D block grouping.}
  \label{fig:grouping}
  \vspace{-0.5em}
\end{wrapfigure}

We investigate encoder design choices for constructing a scan embedding from a single T1w volume. As baselines, we evaluate 3D~ViT with \emph{CLS}, mean token pooling and other grouping strategies (Appendix Sec~\ref{Asec:sa_ablation}). We then introduce atlas--guided \emph{Parcel encoder} (PE) and study (i) generalized mean (GeM) token pooling within each parcel (\(p{=}1,2\)), (ii) strategies to form the scan embedding from parcel embeddings (mean vs.\ learned \emph{parcel scorer}), and (iii) \emph{parcel–drop} regularization. 
Results summarized in Table~\ref{tab:combined_adni_ablations} show that, firstly, replacing global 3D ViT with Parcel encoder coupled with a learned \emph{Parcel scorer} substantially improves both BACC and AUC, demonstrating the need for identifying the most anomalous parcel for better diagnosis. 
Second, comparison with random and block grouping in Fig~\ref{fig:grouping} shows that the gain is not merely due to token grouping, but arises from anatomically meaningful grouping, which provides a useful inductive bias for aggregating information within biologically coherent regions.
Third, the empirical collapse of the late-grouping baseline indicates that parcel identity must be preserved before attention-based contextual mixing.
Finally, a small parcel–drop rate (\(\rho{=}0.10\)) discourages over–reliance on a few parcels and yields the best Stage~A configuration:
\emph{Parcel Encoder + GeM (\(p{=}2\)) + Parcel scorer + parcel–drop ($\rho=0.1$)}.


\textbf{Stage~B Ablations (longitudinal modeling):}
Given cached parcel embeddings from Stage~A, we compare longitudinal heads, (i) GRU versus Transformer, together with (ii) age conditioning, and (iii) Transformer depth (number of blocks \(b\)). 
We also include a baseline that pairs a Longitudinal Transformer with a plain 3D~ViT encoder to isolate the effect of atlas-guided representations. Results are summarized in Table~\ref{tab:combined_adni_ablations}. 
Firstly, for the same longitudinal head, \emph{3D~ViT + Transformer} underperforms P2P, indicating that atlas–guided Parcel embeddings are crucial for better spatio-temporal learning than scan-level embeddings obtained from global token pooling.
Second, transformer heads consistently outperform GRU across both scan and parcel–level variants, suggesting that attention-based temporal aggregation better captures irregular longitudinal evidence.
Third, removing age encoding leads to poorer performance, reinforcing the relevance of chronological context in progression modeling. Also, using relative age as temporal conditioning confirms that age is not a confounding factor in longitudinal modeling (Appendix Sec~\ref{supp:age_effect}). 
Finally, a lightweight transformer ($b{=}2$) is optimal, indicating that performance gains are not driven by model depth but by the representation and structure of evidence aggregation.

\begin{figure*}[t]
  \centering
 \includegraphics[width=\linewidth]{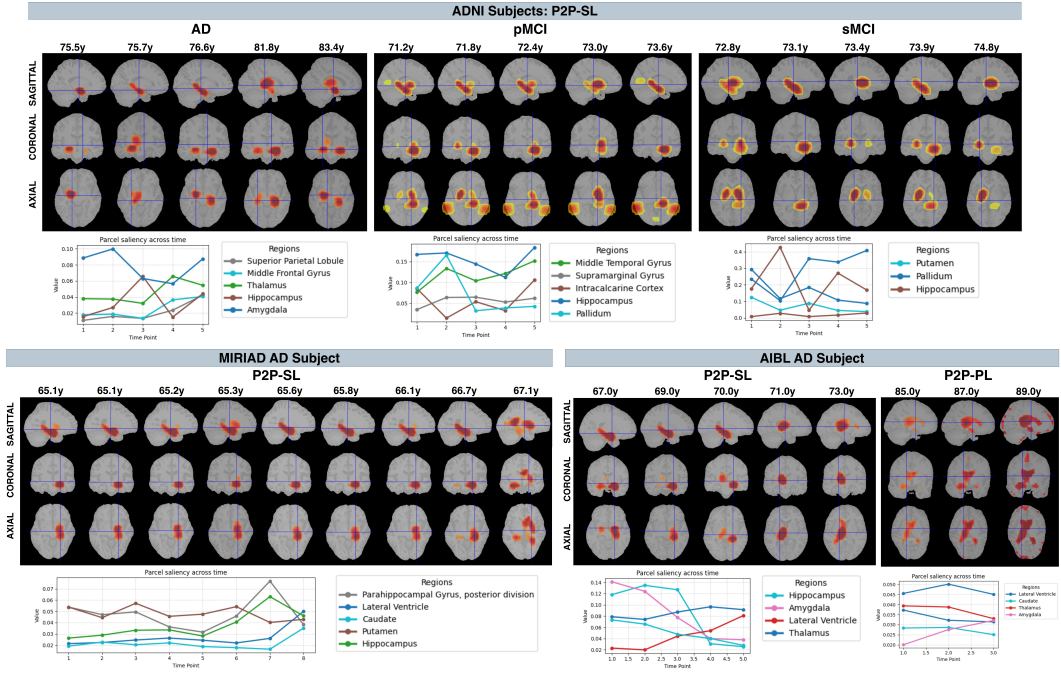}
  \vspace{-1.0em}
  \caption{Attention Rollout visualizations across all three views (coronal, sagittal, \& axial) of correctly predicted longitudinal scans of AD/pMCI/sMCI subjects from ADNI/AIBL/MIRIAD with irregular \& varying visits. Parcel saliencies are also plotted across the respective timepoints. }
  \vspace{-1em}
  \label{fig:attnroll}
\end{figure*}

\begin{figure*}[t]
  \centering
 \includegraphics[width=\linewidth]{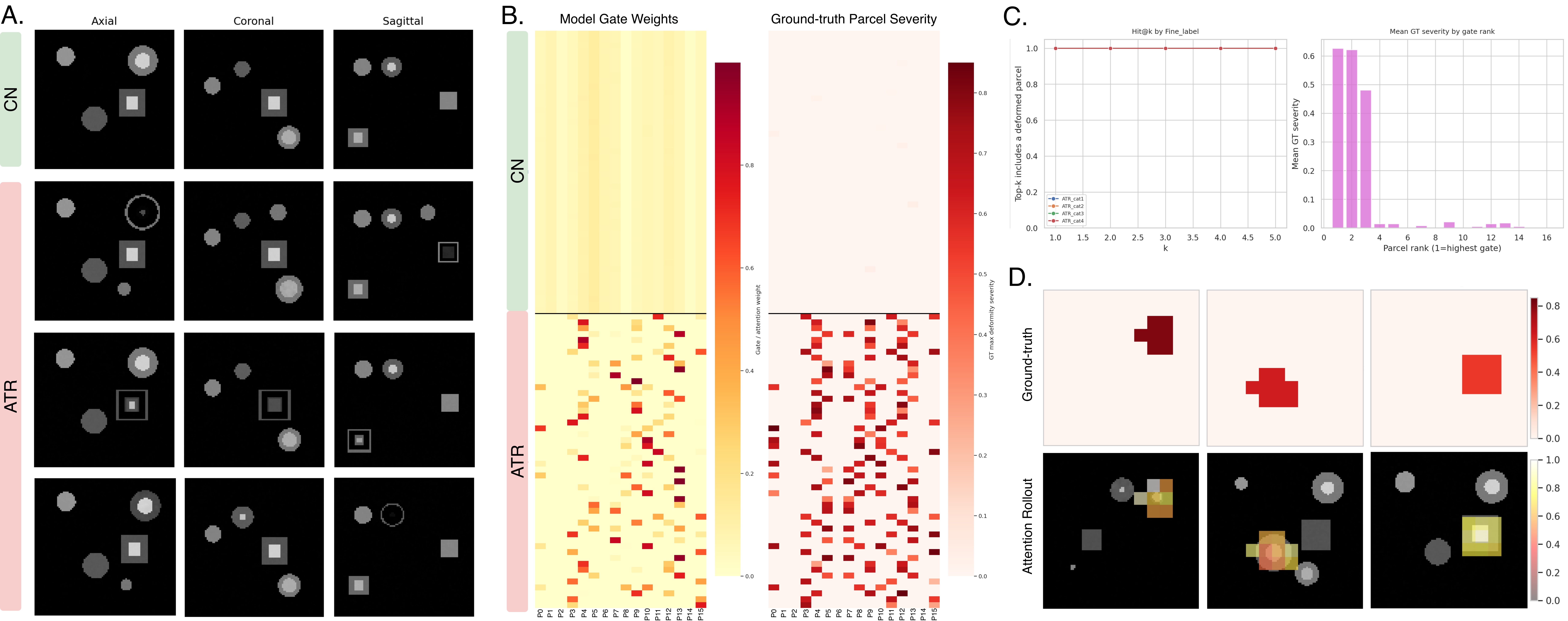}
  \vspace{-1.0em}
  \caption{A. Synthetic Dataset with control normal, and atrophy samples. B. Heatmap of model gate weights versus the ground-truth (GT) parcel severity for control and atrophy samples. C. Stats of Hit@k and GT severity versus parcel rank plot. D. Attention rollouts of atrophy sample. }
  \vspace{-1em}
  \label{fig:synthetic_control}
\end{figure*}

\section{Interpretability}
By design, P2P encodes brain regions into parcel embeddings, enabling stronger, more direct region-specific interpretability. The parcels contributing to a prediction can be easily identified by analysing the Parcel-scorer's weights. We perform this analysis at the subject-level.


\subsection{Brain Samples}
Using Attention rollout~\cite{abnar-zuidema-2020-quantifying} 
in Fig~\ref{fig:attnroll} on the end-to-end framework, temporal attention heatmaps are generated for subjects across the ADNI/AIBL/MIRIAD datasets. These visualizations, combined with the parcel saliency plots, provide better interpretability aligned with model predictions, demonstrating strong coherence with established Alzheimer's literature~\cite{Planche2022,Braak2006,Whitwell2008,Jack2004,Poulin2011,Byun2015,fan2008spatialpatterns}.
A detailed interpretation of Fig~\ref{fig:attnroll}, and extended qualitative cases are discussed in Appendix Sec~\ref {supp:more-explainability}. For cohort-level analysis and additional implementation details of parcel saliencies and attention rollouts, refer to Appendix Sec~\ref{Asec:cohort-saliency},~\ref{Asec:attnroll-details}.

\subsection{Synthetic Data Samples}
\paragraph{Controlled evaluation of parcel-level interpretability.}
The synthetic phantom provides an explicit control for testing whether P2P can recover the spatial origin of an abnormal prediction, since the altered parcels and their severities are known independently by construction. Figure~\ref{fig:synthetic_control}B compares the learned parcel gate weights with this ground truth. Gate weights are comparatively diffuse for CN samples, whereas in ATR samples the highest weights concentrate on the sparse set of parcels containing the introduced abnormalities and qualitatively track their severity. We quantify this correspondence in Fig.~\ref{fig:synthetic_control}C by ranking parcels according to their gate weights. The top-ranked parcels consistently contain a ground-truth altered parcel (Hit@$k$), while the mean ground-truth severity decreases sharply with gate rank, showing that highly weighted parcels preferentially correspond to the regions carrying the strongest abnormal signal. Figure~\ref{fig:synthetic_control}D further compares attention rollout with the known spatial support of the alterations; attention is concentrated within the regions in which the abnormalities were introduced. Importantly, the parcel identities and severity maps are used only for this post-hoc evaluation and are not provided as parcel-level supervision during CN-versus-ATR training. Together, these results provide a controlled validation that P2P identifies \emph{which} parcels contain the discriminative abnormality and localizes supporting evidence within them.

\section{Conclusion and Future Directions}
\label{limitations,conclusion}

We introduced P2P, a framework that combines atlas-guided anatomical representations with variable-length longitudinal modeling to capture region-specific progression patterns from structural MRI. By operating on parcel-level representations, P2P enables scalable modeling of arbitrary-length longitudinal histories while preserving spatial specificity and interpretability. Across cohorts and tasks, the framework demonstrates improved performance in AD diagnosis and MCI-to-AD conversion, suggesting its potential for evidence-based progression modeling aligned with known clinical patterns.
While promising, P2P is currently limited to single-modality T1w MRI and choice of the HO atlas. Future work can extend this framework by integrating complementary modalities like \textit{FLAIR}/\textit{DTI}/\textit{PET} to capture disease heterogeneity, reduce registration dependency, and evaluate on other neurodegenerative conditions like Parkinson’s disease to assess broader generalizability.

{\small
\bibliographystyle{unsrt} 
\bibliography{main} 
}







\newpage
\appendix


\clearpage
\appendix

\begin{center}
    \Large\bfseries Appendix
\end{center}

\startcontents[appendix]
\section*{Contents}

\printcontents[appendix]{}{1}{\setcounter{tocdepth}{3}}


\setcounter{section}{0}
\setcounter{figure}{0}
\setcounter{table}{0}
\renewcommand{\thefigure}{A\arabic{figure}}
\renewcommand{\thetable}{A\arabic{table}}


\section{Datasets and Preprocessing}
\label{Asec:dataset}

\subsection{Public Datasets and Cohort Overview.}

\begin{figure*}[t]
  \centering
 \includegraphics[width=\linewidth]{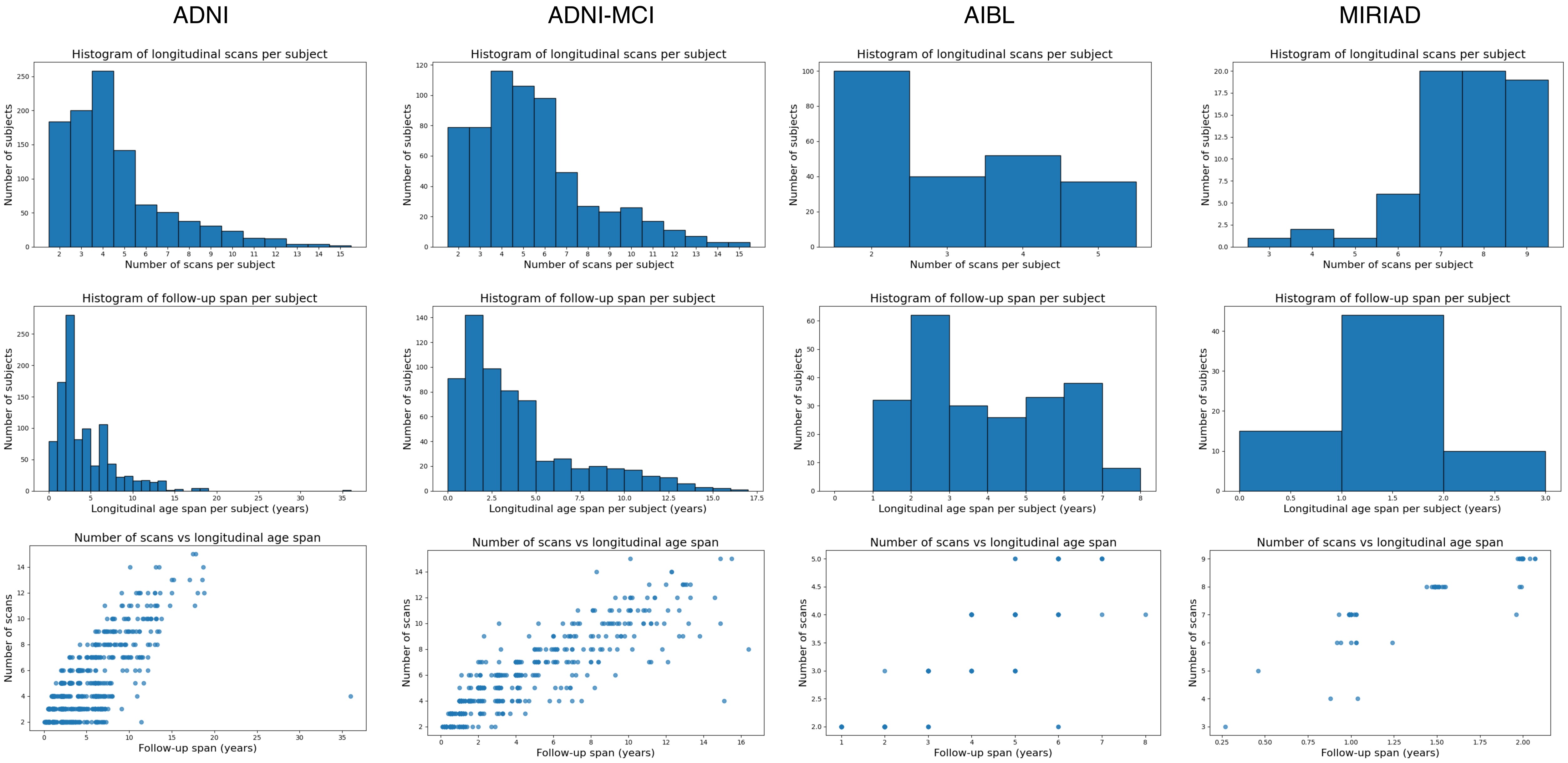}
  \caption{Longitudinal stats of the datasets.}
  \label{fig:long_stats}
\end{figure*}

Our experimental evaluation spans three longitudinal cohorts i.e, ADNI, AIBL, and MIRIAD chosen to collectively stress-test the model across the full spectrum of real-world acquisition variability. The detailed longitudinal statistics for each cohort, including scan frequency distributions, follow-up span distributions, and scan-versus-span scatter, are visualised in Fig.~\ref{fig:long_stats}.

\begin{itemize}
    \item \textbf{ADNI} (ADNI 1/GO/2/3/4) is a large multi-center study spanning dozens of North American sites (U.S./Canada) with contributions across Siemens, GE, and Philips scanners at 1.5T and 3T\footnote{We include scans with 2.9T, but count them under 3T category for simplicity.}, and protocol families such as MPRAGE. It offers broad diagnostic coverage (CN/MCI/AD) with multi-year follow-ups, yielding substantial inter-site and inter-scanner variability in acquisition parameters and subject demographics. As reflected in Fig.~\ref{fig:long_stats}, the ADNI cohort (full) exhibits a long tail extending to 15 scans, indicating a heterogeneous follow-up density.  Crucially, this variability is representative of real-world longitudinal neuroimaging registry data, and P2P(our model) is designed to accommodate such irregular temporal sampling without requiring interpolation or fixed visit schedules.
    
    \item \textbf{AIBL} (Australian Imaging, Biomarkers and Lifestyle) is a prospective Australian cohort with imaging conducted primarily at multiple sites in \emph{Melbourne} and \emph{Perth}, predominantly on Siemens platforms (1.5T/3T). It provides serial T1 weighted scans with rich clinical and biomarker metadata and comparatively fewer longitudinal visits per subject than ADNI. 
    
    \item \textbf{MIRIAD} (Minimal Interval Resonance Imaging in Alzheimer’s Disease) is a single-center cohort from London (UCL), acquired on a fixed scanner/protocol with dense revisit intervals (weeks–months). It enables high sensitivity to within-subject change under no site/scanner variability. 

    \item \textbf{NIFD} (Neuroimaging in Frontotemporal Dementia) is a multi-site NIH-funded study designed to characterise structural and functional neuroimaging signatures across the FTD spectrum. Data were acquired across three academic medical centres: the University of California San Francisco (UCSF), Mayo Clinic (Rochester), and Massachusetts General Hospital (MGH).

\end{itemize}

\begin{table*}[b]
\caption{Cohort statistics of all datasets used in our experiments.}
\centering
\scriptsize
\setlength{\tabcolsep}{3pt}
\renewcommand{\arraystretch}{1}
\begin{tabular}{lcccccccc}

\toprule
\textbf{Dataset} & \textbf{\#scans} & \textbf{\#subjects} & \textbf{\#Max}& \textbf{Age} & \textbf{Gender} & \textbf{Field strength} & \textbf{Manufacturer} \\
 &  &  & \textbf{visits} & \textbf{(mean$\pm$std,y)} & \textbf{(M / F)} & \textbf{(1.5T / 3.0T)} & \textbf{(Siemens/GE/Philips)} \\
\midrule
ADNI(AD/CN)   & 5{,}627 & 1{,}703 & 15 & $72.93 \pm 7.87$ & 734 / 969 & 429 / 1{,}411 & 1{,}063 / 476 / 289 \\
ADNI(MCI)   & 5{,}308 & 687 & 15 & $74.18 \pm 7.44$ & 408 / 279 & 275 / 512 & 389 / 258 / 134 \\
AIBL   & 1{,}084 &   599   &  5  & $73.36 \pm 6.42$ & 255 / 343 & 107 / 541      & 599 / 0 / 0 \\
MIRIAD &   708   &    69   &  9  &  $69.89 \pm 6.98$ & 31 / 38   & 69 / 0         & 0 / 69 / 0 \\
NIFD & 962 & 332 & 7 & $64.18 \pm 8.16$ & 176 / 156 & 0 / 332 & 317 / 16 / 0\\
\bottomrule
\end{tabular}
\label{tab:cohort_adni_stats}
\end{table*}

\paragraph{\textbf{Protocol and Subject-wise splits:}}\label{subject-wise_split} 
All experiments adopt strict \emph{subject-wise} partitioning, thereby eliminating any possibility of scan-level leakage across sets.  Stratification across splits is enforced over all known sources of variability, including scanner platform, field strength, and age distribution, to guard against spurious performance attributable to demographic or acquisition imbalance. All reported results are obtained under 5-fold cross-validation, with 20\% held out as the test set and 10\% validation set within each fold, and final performance metrics are reported as the mean$_{\pm std}$ over test set from all five folds.

\paragraph{\textbf{ADNI MCI Curation (sMCI/pMCI):}}
In the context of Alzheimer's disease progression, Mild Cognitive Impairment (MCI) 
represents an intermediate clinical state between normal aging and dementia. Within 
this cohort, subjects may follow two distinct longitudinal trajectories: 
\textbf{stable MCI (sMCI)}, in which the subject exhibits no progression to dementia 
throughout the entire follow-up period, and \textbf{progressive MCI (pMCI)}, in which 
the subject subsequently converts to a dementia diagnosis at a later timepoint. 
Distinguishing between these two groups at the earliest possible stage is of 
significant clinical importance, as it enables timely intervention for high-risk 
individuals.

\begin{figure*}[t]
  \centering
 \includegraphics[width=0.7\linewidth]{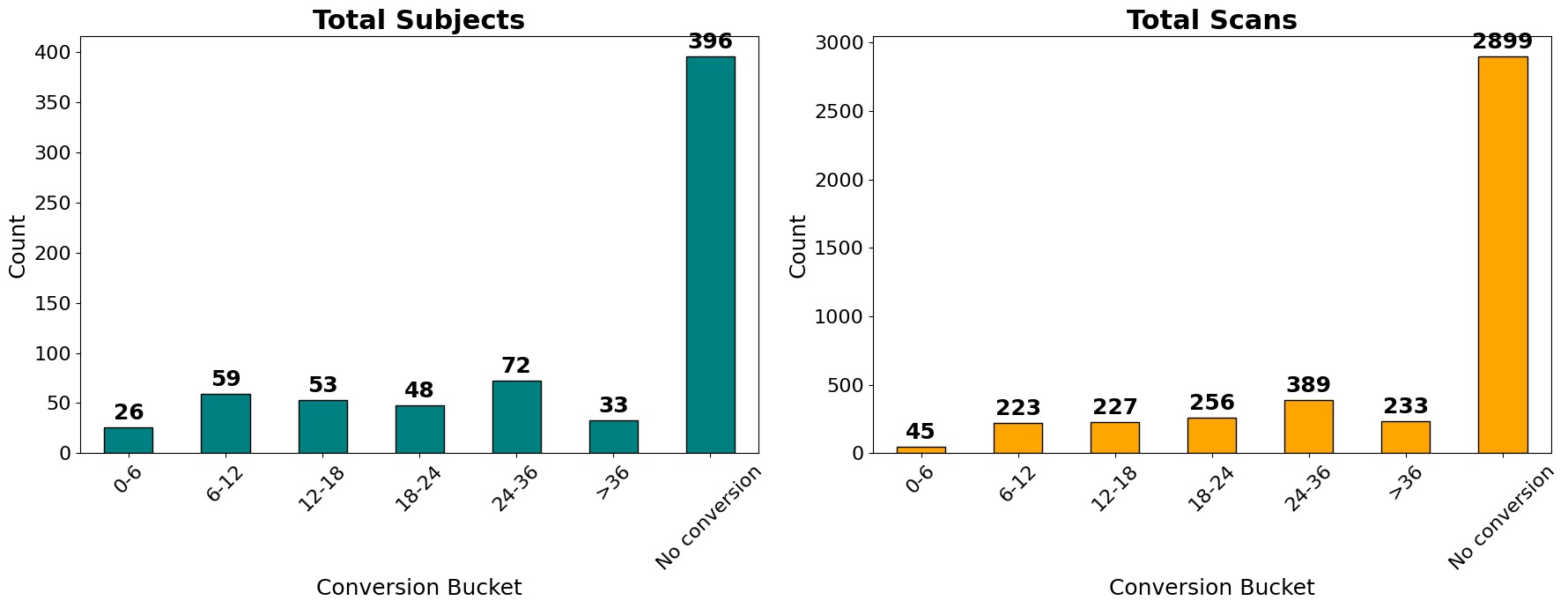}
  \caption{MCI cohort statistics of ADNI under each conversion category (in months).}
  \label{fig:mci_cohort_stats}
\end{figure*}

We define the \emph{baseline} visit as the timepoint corresponding to the subject's 
first MCI diagnosis. For \textbf{sMCI} subjects, all longitudinal scans acquired from 
baseline onward are assigned the \texttt{sMCI} label. For \textbf{pMCI} subjects, all 
scans from baseline up to, but \emph{not including}, the visit at which a dementia 
diagnosis is first recorded are labeled \texttt{pMCI}. The scan acquired at the point 
of dementia conversion is explicitly excluded from the dataset, as it reflects 
dementia-stage pathology rather than MCI-stage biomarkers.
A more detailed stats of the conversion times for ADNI subjects are presented in Fig~\ref{fig:mci_cohort_stats}.

\begin{figure*}[t]
  \centering
 \includegraphics[width=\linewidth]{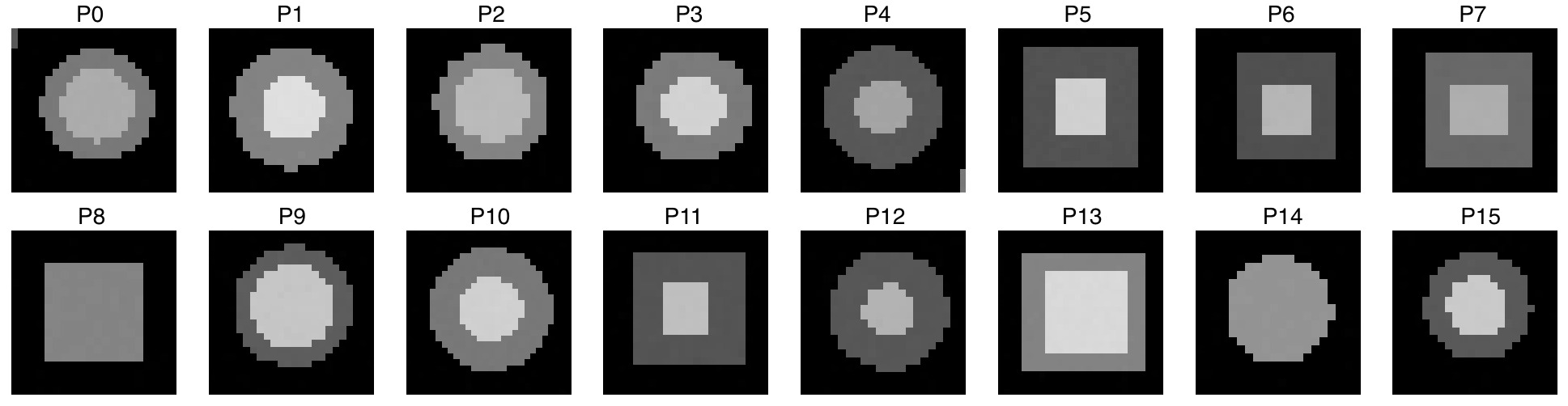}
  \caption{16 Parcels generated in the Synthetic Phantom Dataset.}
  \label{fig:synth_parcel}
\end{figure*}

\paragraph{\textbf{Harvard-Oxford Atlas:}}
A brain atlas partitions the brain into named regions (``parcels”) so we can compare scans at the level of anatomy rather than individual voxels. A \emph{probabilistic} atlas assigns each voxel a soft membership over regions, which better handles inter-subject variability and boundary uncertainty. We use the Harvard–Oxford atlas (69 parcels here), from both cortical and subcortical atlases of 1mm resolution, because its granularity suits T1w sMRI and AD-relevant anatomy (e.g., hippocampus, parahippocampal, amygdala, PCC/ precuneus). Its soft labels let us aggregate tokens into anatomically coherent sets while retaining uncertainty near borders. This yields stable, interpretable parcel embeddings.
We note that the HO atlas does not delineate certain cytoarchitectural structures relevant to AD as independent parcels. Specifically, the entorhinal cortex and perirhinal cortex are anatomically subsumed within the anterior division of the Parahippocampal Gyrus parcel and are therefore implicitly represented as part of the Parahippocampal Gyrus. Similarly, hippocampal subfields (CA1, subiculum, dentate gyrus) are collectively subsumed within the global Hippocampus parcel.
Consequently, while these sub-regions are not individually resolved, their cumulative signal and associated morphometric variations remain captured within these broader macro-anatomical boundaries.

\paragraph{\textbf{Preprocessing:}}
All scans of shape $182\times218\times182$ are preprocessed with a common pipeline comprising: reorientation to canonical (\texttt{fslreorient2std}), bias-field correction (ANTs \texttt{N4BiasFieldCorrection}), skull stripping (SynthStrip~\cite{Andrew2022}), affine+nonlinear registration to MNI152 1\,mm (FSL \texttt{FLIRT}$\rightarrow$\texttt{FNIRT}), intensity z-scoring within the brain mask and final masking. Finally, the scans are padded to be divisible by the patch volumes. The resulting MNI-aligned, skull-stripped, z-scored volumes are saved per scan and used for tokenization in Parcel encoder of Stage~A.

\subsection{Controlled Synthetic Phantom Dataset}
\label{supp:syth_data}

We construct a synthetic 3D phantom dataset to provide an explicit ground truth for evaluating parcel-level localization. The objective of this experiment is not to approximate the appearance of a real neuroimaging cohort, but to create a controlled setting in which the parcels responsible for an abnormal sample are known exactly. This allows us to test whether P2P assigns high importance to the regions in which the discriminative signal was actually introduced, which cannot be established unambiguously from diagnostic labels alone in real datasets.

Each sample is a $96\times96\times96$ volume containing 16 non-overlapping parcels distributed throughout the volume. Parcels are instantiated as axis-aligned spheres or cubes with heterogeneous sizes and internal intensity structure. The same parcel layout is shared across subjects, providing an explicit correspondence between parcels across samples. The volumes are divided into non-overlapping $8\times8\times8$ patches, yielding 1,728 patches per sample, and patch-to-parcel membership is derived from the known phantom geometry. The 16 generated parcels are presented in Fig~\ref{fig:synth_parcel}.

The dataset contains 10,000 samples, equally divided into CN ($n=5{,}000$) and ATR ($n=5{,}000$). CN volumes retain the nominal parcel geometry and intensity structure. To prevent the classification problem from reducing to exact template matching, mild scan-level intensity noise and smooth bias variation are introduced across samples. In addition, a small proportion of CN samples may contain a very weak local intensity or texture variation affecting at most one parcel; these perturbations are deliberately constrained to nuisance-level magnitudes and do not introduce the strong abnormality signal used for ATR samples.

For each ATR sample, strong alterations are introduced only within a sparse subset of parcels (2--4 parcels), while the remaining parcels retain their nominal appearance apart from the same global nuisance variability. Alterations encompass complementary geometric and appearance changes, including parcel contraction, expansion, intensity reduction, local texture perturbation, and loss of internal contrast. One or more alterations may be applied to an affected parcel, with their magnitudes sampled independently. Importantly, the generator records the affected parcel identities and the applied severity for every sample. We therefore obtain a parcel-level ground-truth severity vector
\begin{equation}
    \mathbf{s} = [s_1,\ldots,s_P], \qquad P=16,
\end{equation}
where $s_p=0$ for an unaffected parcel and $s_p>0$ quantifies the magnitude of the synthetic abnormality in an affected parcel. These labels are used only for evaluating localization and are not provided to P2P as parcel-level supervision. The model is trained for the binary CN-versus-ATR task, allowing us to determine whether parcel-level importance emerges from the classification objective and corresponds to the true spatial origin of the abnormal signal.

\subsection{Registration Quality and Parcel Coverage Robustness}
\label{supp:registration_qc}

To assess whether disease-related performance could be confounded by poorer registration or atlas alignment, we performed a patch-level parcel coverage quality-control analysis. For each registered scan, we derived a brain mask from the preprocessed image and converted it into the same patch space used for parcel tokenization. We then quantified: (i) the number of non-empty parcels, (ii) mean parcel coverage relative to the CN reference distribution, (iii) the number of low-coverage parcels, and (iv) the percentage of valid brain patches assigned to any atlas-defined parcel.

Table~\ref{tab:registration_qc} summarizes the main QC metrics across CN, AD, sMCI, and pMCI scans. All disease/progression groups retained complete parcel coverage, with a median of 69/69 non-empty parcels and zero low-coverage parcels. 
Although mean parcel coverage showed some group differences, the absolute differences were small. Importantly, AD, sMCI, and pMCI did not show parcel dropout or systematic failure of atlas assignment.

\begin{table*}[t]
\centering
\caption{Patch-level registration and parcel coverage QC across clinical groups. Values are reported as mean $\pm$ standard deviation, with medians shown in parentheses where relevant. All groups retained nearly complete parcel coverage, indicating that downstream disease/progression effects are unlikely to be driven by group-specific registration failure.}
\label{tab:registration_qc}
\scriptsize
\setlength{\tabcolsep}{4pt}
\renewcommand{\arraystretch}{1.15}
\begin{tabular}{lccccc}
\toprule
\textbf{Group} 
& \textbf{N} 
& \textbf{Non-empty parcels} 
& \textbf{Mean parcel coverage} 
& \textbf{Low-coverage parcels}  \\
\midrule
CN 
& 4144 
& $69.00 \pm 0.00$ $(69)$ 
& $0.9925 \pm 0.0177$ $(0.9958)$ 
& $0.018 \pm 1.075$ $(0)$ \\

AD 
& 1485 
& $69.00 \pm 0.00$ $(69)$ 
& $0.9869 \pm 0.0122$ $(0.9899)$ 
& $0.010 \pm 0.170$ $(0)$ \\

sMCI 
& 3915 
& $69.00 \pm 0.00$ $(69)$ 
& $0.9913 \pm 0.0088$ $(0.9940)$ 
& $0.002 \pm 0.045$ $(0)$ \\

pMCI 
& 1393 
& $69.00 \pm 0.00$ $(69)$ 
& $0.9891 \pm 0.0110$ $(0.9927)$ 
& $0.006 \pm 0.117$ $(0)$ \\
\bottomrule
\end{tabular}
\end{table*}

The Kruskal--Wallis test~\cite{kruskal1952use} showed no significant group difference in the number of non-empty parcels ($p=0.65$), indicating no evidence of systematic parcel dropout across CN, AD, sMCI, and pMCI. The number of low-coverage parcels was also negligible in all groups, with a median of zero across all cohorts. 
We also observed a small reduction in valid brain-mask patches and total brain-mask voxels following the expected ordering CN $>$ sMCI $>$ pMCI $>$ AD. This trend is biologically plausible in neurodegenerative cohorts and likely reflects disease-related tissue loss or atrophy rather than registration failure. Crucially, this reduction did not translate into missing parcels or poor atlas assignment. Overall, the QC analysis supports that the observed disease and progression-related effects are unlikely to be explained by poorer registration quality or reduced parcel coverage in AD, sMCI, or pMCI scans.

\subsection{Robustness to preprocessing variations.}
\label{supp:robustness}
We evaluate preprocessing-related robustness by perturbing the bias-field
log-amplitude, contrast scale, Gibbs $k$-space truncation fraction, image rotation,
and translation for the P2P-SA model on the AD/CN task. Results are averaged over
five folds and summarized in Table~\ref{tab:preprocessing_robustness}.

\begin{wraptable}{r}{0.5\linewidth}
\centering
\vspace{-2em}
\caption{Robustness of P2P-SA to preprocessing perturbations on ADNI AD/CN classification.}
\label{tab:preprocessing_robustness}
\resizebox{0.85\linewidth}{!}{
\begin{tabular}{lccc}
\toprule
\textbf{Perturbation} &
\textbf{Level} &
\textbf{AUC (\%)} &
\textbf{$\Delta$AUC} \\
\midrule

Original (P2P-SA) & --
& $91.18$
& -- \\

Bias field & 0.1
& $90.60$
& $-0.58$ \\

Bias field & 0.2
& $90.54$
& $-0.64$ \\

Bias field & 0.3
& $90.32$
& $-0.86$ \\

Contrast & 0.5
& $87.80$
& $-3.38$ \\

Contrast & 1.5
& $92.29$
& $+1.11$ \\

Contrast & 2.0
& $92.72$
& $+1.54$ \\

Gibbs & 0.1
& $90.84$
& $-0.34$ \\

Gibbs & 0.2
& $90.91$
& $-0.27$ \\

Gibbs & 0.3
& $90.85$
& $-0.33$ \\

Rotation & $1^{\circ}$
& $90.01$
& $-1.17$ \\

Rotation & $2^{\circ}$
& $88.79$
& $-2.39$ \\

Translation & 1 voxel
& $88.43$
& $-2.75$ \\

Translation & 2 voxels
& $87.19$
& $-3.99$ \\

\bottomrule
\end{tabular}
}
\end{wraptable}

The model is largely stable under moderate intensity inhomogeneity, Gibbs
artifacts, and small geometric perturbations, with degradation becoming more
noticeable only for stronger contrast changes or larger spatial misalignment.
Importantly, the probabilistic atlas is registered to each preprocessed scan,
and parcel membership is computed at the $8^3$-voxel patch level rather than
requiring exact voxel-wise correspondence with parcel boundaries. Consequently,
small residual registration errors need not induce abrupt changes in parcel
assignment, since boundary patches can retain graded membership across adjacent
parcels.

\begin{figure*}[t]
  \centering
 \includegraphics[width=\linewidth]{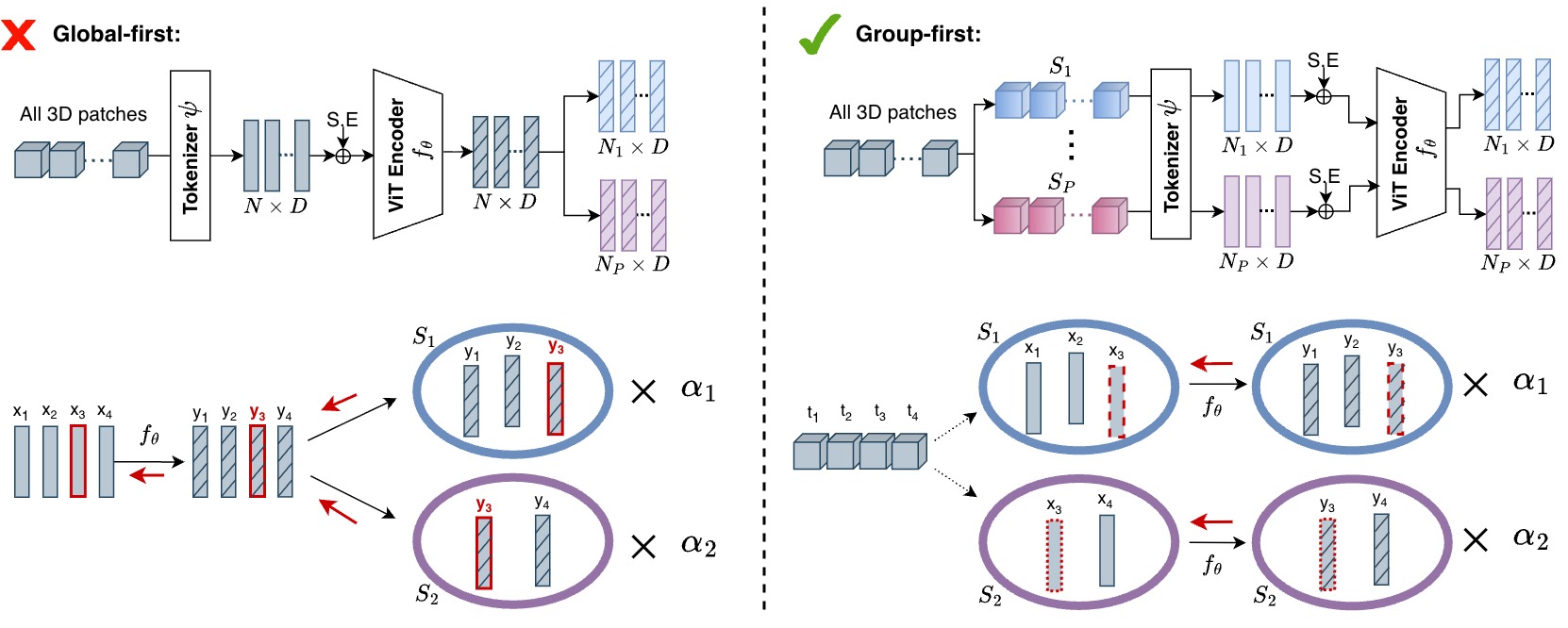}
  \caption{Collapse scenario with late grouping (global-first). A toy example with four tokens and two parcels shows that when attention is applied globally before grouping (left), token mixing causes the parcel weights $\alpha$ to behave as a token-level gate rather than parcel-specific importance weights. This leads to ambiguous parcel attribution and collapse. In contrast, the group-first design (right), used in P2P, performs attention within parcels, preserving parcel identity and enabling interpretable parcel weighting.}
  \label{fig:collapse}
\end{figure*}

\vspace{1em}
\section{Collapse scenario}
\label{Asec:collapse}

Our goal is to learn a scan representation as a parcel-weighted combination
\begin{equation*}
s \;=\; \sum_{p=1}^{P} \alpha_p z_p ,
\end{equation*}
where \(z_p\) is the embedding of parcel \(p\) from set \(S_p\) and \(\alpha_p\) indicates
its importance.  
\noindent
For specificity of parcel representations, the desired optimization behaviour is
\begin{equation*}
\frac{\partial s}{\partial \alpha_p} \;\propto\; z_p ,
\end{equation*}
so that each parcel weight is updated using evidence originating
from its own parcel.

Let \(X \in \mathbb{R}^{N\times D}\) denote token features of a scan and  
\(M \in \mathbb{R}_{\ge0}^{N\times P}\) be the token-to-parcel assignment matrix.

\paragraph{\textbf{Global-first architecture:}}
In the global-first design, self-attention is applied to all tokens first
\begin{equation*}
Y \;=\; f_{\theta}(X) \in \mathbb{R}^{N\times D}.
\end{equation*}
\noindent
Tokens are then pooled into parcels
\begin{equation*}
U \;=\; M^\top Y \in \mathbb{R}^{P\times D},
\end{equation*}
\noindent
and the scan representation becomes: $s \;=\; \alpha^\top U.$
\noindent
Substituting \(U = M^\top Y\),
\begin{equation*}
s
=
\alpha^\top M^\top Y
=
(M\alpha)^\top Y .
\end{equation*}
\noindent
Define the induced token gate,
$b = M\alpha.$
Thus the representation becomes
\begin{equation*}
s \;=\; b^\top Y .
\end{equation*}
\paragraph{Non-identifiability of parcel weights.}
The model depends on the parcel weights only through the
token weights \(b\).
If two parcel weight vectors satisfy
\begin{equation*}
M\alpha = M\tilde{\alpha},
\quad \Rightarrow \quad
s(\alpha) = s(\tilde{\alpha}).
\end{equation*}
Hence different parcel weight configurations can produce exactly the same scan representation.
Consequently, the parcel weights are not uniquely identifiable:
the loss function cannot distinguish between multiple
\(\alpha\) that induce the same token gate \(b\).

\paragraph{Gradient structure.}

The dependence of the representation on parcel weights becomes
\begin{equation*}
s(\alpha) = (M\alpha)^\top Y
\quad \Rightarrow \quad
\frac{\partial s}{\partial \alpha}
=
M^\top Y .
\end{equation*}
Component-wise:
$
\frac{\partial s}{\partial \alpha_p}
=
\sum_{i=1}^{N} M_{ip} Y_i .
$
Thus each parcel update is driven by a sum of globally mixed token
evidence. When parcels share tokens or tokens become correlated after
global attention, these gradients become similar across parcels,
encouraging diffuse parcel weights.

\paragraph{Illustrative example (Fig.\ref{fig:collapse}):}

Consider four tokens \(x_1,x_2,x_3,x_4\) and two parcels
\begin{equation*}
S_1 = \{x_1,x_2,x_3\},
\qquad
S_2 = \{x_3,x_4\}.
\end{equation*}
After global attention, let token evidences be \(y_1,y_2,y_3,y_4\).
Parcel pooling gives
\begin{equation*}
U_1 = y_1 + y_2 + y_3,
\qquad
U_2 = y_3 + y_4.
\end{equation*}
The representation becomes
\begin{align*}
s
&=
\alpha_1 U_1 + \alpha_2 U_2 \\
&=
\alpha_1 y_1
+
\alpha_1 y_2
+
(\alpha_1+\alpha_2)y_3
+
\alpha_2 y_4 .
\end{align*}
The shared token \(y_3\) depends on \(\alpha_1+\alpha_2\).
Thus, multiple parcel weight configurations produce the same
representation, preventing unique parcel attribution.

\paragraph{\textbf{Group-first architecture:}}

In the group-first design, tokens are first separated by parcel.
Let \(X^{(p)}\) denote tokens belonging to parcel \(p\).
Parcel embeddings are computed independently:
$
z_p = f_{\theta}(X^{(p)}).$
The scan representation is: $
s = \sum_{p=1}^{P} \alpha_p z_p. $
The gradient becomes:
$
\frac{\partial s}{\partial \alpha_p} = z_p$ ,
which depends only on parcel-specific features.

\paragraph{Empirical implication.}

Global-first models' optimum
is defined only up to the induced token gate \(b=M\alpha\).
This ambiguity causes parcel weights to become diffuse during training.
In practice we observe this behaviour as a \emph{collapse} of parcel
importance and substantially degraded performance (Table~\ref{tab:combined_adni_ablations}).
In contrast, group-first preserves parcel identity before feature mixing, ensuring that convergence aligns with the desired
parcel-wise decomposition.

\begin{figure}[t]
  \centering
   \includegraphics[width=0.6\linewidth]{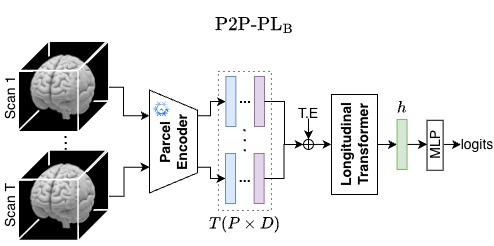}
   \caption{A parcel level variant using an Age-aware Longitudinal Transformer framework (P2P-$\textbf{PL}_{\textbf{B}}$). Here, T.E. denotes the visit age encoding of the subject.}
   \label{fig:parcel_variant}
\end{figure}
\vspace{1em}

\section{Methods \& Experiments}

\subsection{Longitudinal Variant}
\label{Asec:long_variantB}
We have also designed another parcel level variant:
\paragraph{\textbf{Parcel level - Type B (P2P-$\textbf{PL}_{\textbf{B}}$)}: Joint parcel only Longitudinal transformer.}
This variant learns \emph{joint} cross-parcel and cross-time interactions in a single transformer, capturing coordinated anatomical changes (Fig.~\ref{fig:parcel_variant}). We build a single sequence that interleaves \emph{all} parcel–time tokens and add their respective age-based temporal embedding:
\vspace{-0.2em}
\begin{align*}
\tilde{z}_{p,t} &= z_{p,t} + \phi_t(A_s(t)),\\
h &= \mathcal{T}_{parcelB}\big(\tilde{z}_{1:P,1:T}\big)\in\mathbb{R}^{D}
\end{align*}
A Longitudinal transformer $\mathcal{T}_{parcelB}(\cdot)$ jointly models cross-parcel and cross-time interactions. The \texttt{[CLS]} token is fed to an MLP head for logits: $\hat{y}=\text{softmax}\big(\text{MLP}(h)\big)$.

\subsection{Prior work and baseline comparison}


\begin{table*}[h]
\centering
\caption{
Comparison of P2P with single-scan and longitudinal baselines. 
While several baselines operate on high-dimensional scan inputs, most either ignore longitudinal structure, require fixed-length scan sequences, or do not explicitly model anatomy-aware parcel-level progression. 
P2P is designed to support high-dimensional longitudinal modelling with anatomy-aware representations, variable-length visits, and irregular follow-up intervals.
}
\label{tab:p2p_baseline_comparison}
\scriptsize
\setlength{\tabcolsep}{3.2pt}
\renewcommand{\arraystretch}{1.18}

\begin{tabular}{
    lccclccccc
}
\toprule
\textbf{Model} &
\textbf{Long.} &
\textbf{T} &
\textbf{FLOPs} &
\textbf{Spatial Backbone} &
\textbf{Anat.-aware} &
\textbf{High-dim.} &
\textbf{Folds} &
\textbf{Var. length} &
\textbf{Irreg. interval} \\
& & & \textbf{(e+12)} & & & & \textbf{(original)} & & \\
\midrule

\multicolumn{10}{l}{\textit{Single-scan}} \\
\midrule
DenseNet121
& \xmark & 1T & 1.37
& 3D CNN
& \xmark & \cmark
& -- 
& \xmark & \xmark \\

ResNet152
& \xmark & 1T & 1.97
& 3D CNN
& \xmark & \cmark
& --
& \xmark & \xmark \\

ViT
& \xmark & 1T & 10.12
& 3D Transformer
& \xmark & \cmark
& --
& \xmark & \xmark \\

M3T
& \xmark & 1T & 7.01
& Hybrid (slice-based)
& \xmark & \xmark
& 1
& \xmark & \xmark \\

AAGN
& \xmark & 1T & 8.09
& 3D CNN
& \cmark & \xmark
& 5
& \xmark & \xmark \\

\rowcolor{green!6}
\textbf{P2P-SA}
& \xmark & 1T & 2.48
& 3D Transformer
& \cmark & \cmark
& \textbf{5}
& \xmark & \xmark \\

\midrule
\multicolumn{10}{l}{\textit{Multi-scan / longitudinal }} \\
\midrule

Longformer
& \cmark & 2T & 5.37
& 3D CNN
& \xmark & \cmark
& 1
& \xmark & \xmark \\

DenseNet121+GRU
& \cmark & 15T & 20.55
& 3D CNN
& \xmark & \cmark
& --
& \cmark & \cmark \\

DenseNet121+LT
& \cmark & 15T & 20.55
& 3D CNN
& \xmark & \cmark
& --
& \cmark & \cmark \\

ViT+LT
& \cmark & 15T & 131.56
& 3D Transformer
& \xmark & \cmark
& --
& \cmark & \cmark \\

\rowcolor{green!6}
\textbf{P2P-SL}
& \cmark & 2T & 4.96
& 3D Transformer
& \cmark & \cmark
& \textbf{5}
& \cmark & \cmark \\

\rowcolor{green!6}
\textbf{P2P-SL}
& \cmark & 15T & 37.30
& 3D Transformer
& \cmark & \cmark
& \textbf{5}
& \cmark & \cmark \\

\rowcolor{green!6}
\textbf{P2P-PL}
& \cmark & 15T & 37.23
& 3D Transformer
& \cmark & \cmark
& \textbf{5}
& \cmark & \cmark \\

\bottomrule
\end{tabular}

\vspace{2pt}
\begin{flushleft}
\footnotesize
\textbf{Note.} \textit{Long.} denotes whether the model explicitly uses longitudinal scans. \textit{FLOPs} denotes full FLOPs reported for the respective visits considered (T) using ptflops.
\textit{Anat.-aware} indicates whether the method explicitly incorporates anatomical or parcel-level structure. 
\textit{High-dim.} indicates whether the method directly operates on high-dimensional scan-derived representations or downsampled inputs. ``\textit{Folds (original)}'' denotes the original paper's settings for reporting results. ``--'' indicates that the corresponding entry is a baseline.
\textit{Var. length} denotes support for variable numbers of visits across subjects. 
\textit{Irreg. interval} denotes whether irregular follow-up intervals can be explicitly incorporated. 
\cmark: supported; \xmark: not supported. 
\end{flushleft}
\end{table*}

\begin{table*}[t]
\centering
\scriptsize
\caption{Performance of other methods across datasets and tasks on the full dataset as P2P. Values are reported as $\mathrm{mean}_{\pm \mathrm{std}}$ over 5-folds (in \%).}
\label{tab:densenet_bilstm_lda}
\setlength{\tabcolsep}{3pt}
\begin{tabular}{lcccccccc}
\toprule
& \multicolumn{2}{c}{ADNI (AD/CN)} 
& \multicolumn{2}{c}{ADNI (sMCI/pMCI)} 
& \multicolumn{2}{c}{AIBL (AD/CN)} 
& \multicolumn{2}{c}{MIRIAD (AD/CN)} \\
\cmidrule(lr){2-3} \cmidrule(lr){4-5} \cmidrule(lr){6-7} \cmidrule(lr){8-9}
Method & BACC & AUC & BACC & AUC & BACC & AUC & BACC & AUC \\
\midrule
\multirow{2}{*}{\begin{tabular}{l}Jomeiri \\et al.\cite{Jomeiri2024}\end{tabular}}
& $77.43_{\pm 3.95}$ 
& $87.11_{\pm 1.09}$ 
& $62.24_{\pm 1.47}$ 
& $73.06_{\pm 4.24}$ 
& $69.95_{\pm 3.85}$ 
& $85.23_{\pm 4.90}$ 
& $81.70_{\pm 3.52}$ 
& $85.45_{\pm 4.99}$ \\
&&&&&&&& \\
\multirow{2}{*}{\begin{tabular}{l}Ouyang \\et al.\cite{ouyang2022disentangle} \end{tabular}}
& $74.61_{\pm 2.21}$ 
& $82.45_{\pm 2.37}$ 
& $60.17_{\pm 2.93}$ 
& $69.74_{\pm 3.13}$ 
& $65.47_{\pm 2.69}$ 
& $82.06_{\pm 4.32}$ 
& $86.66_{\pm 3.87}$ 
& $88.82_{\pm 4.44}$ \\
&&&&&&&& \\
\bottomrule
\end{tabular}
\end{table*}

A detailed comparison of the prior works and baselines is presented in Table~\ref{tab:p2p_baseline_comparison}. Notably, all the SOTA prior works and baselines, that were trained for comparison, were originally trained on smaller data cohorts. Longformer~\cite{chen2024longformer} reported their results on a single fold only in the original paper.
We retrained these architectures on the same inputs as P2P
and reported their results for all 5 folds using the same splits. 
All these chosen architectures could consume the full dataset, like P2P, without any constraints on dataset size. For the other works that use fixed histories like~\cite{Jomeiri2024} or compromise the spatial resolution to $64 \times 64 \times 64$ like~\cite{ouyang2022disentangle}, we have retrained these models and reported their results in Table~\ref{tab:densenet_bilstm_lda}. To enable~\cite{Jomeiri2024} to handle variable-length inputs, we introduced age as the temporal embedding, like the other Longitudinal baselines in the main paper. Also,~\cite{ouyang2022disentangle} uses only a pair of scans for the longitudinal modelling. It is observed that P2P-SL with paired scans (reported in Table~\ref{tab:sota_comparison_adni_split}) outperforms these methods.




\subsection{More Ablations for P2P-SA}
\label{Asec:sa_ablation}
To validate that the gains of our anatomy-aware grouping are attributable to biologically meaningful parcel structure rather than merely to token partitioning itself, we introduce two non-anatomical grouping baselines. These baselines preserve the same overall framework, including the encoder, parcel-wise aggregation mechanism using the scorer, classifier, and the total number of groups, while changing only the grouping strategy. This enables a controlled ablation in which the effect of anatomical priors can be isolated from the effect of grouped processing alone.

\textit{(i) Random grouping:}
In the first baseline, all foreground brain tokens are reassigned into a fixed set of $69$ groups using a random partition. Background tokens are excluded, and each valid brain token is assigned to exactly one random group. The random grouping is generated once and then kept fixed for all samples, ensuring that the comparison reflects a stable non-anatomical partition. This baseline tests whether the proposed framework derives its benefit simply from dividing tokens into multiple groups, irrespective of whether those groups carry anatomical meaning.

\begin{wraptable}{r}{0.55\textwidth}
  \centering
  \vspace{-2em}
\caption{More Ablations for P2P-SA (Stage~A, single scan with Parcel Encoder) based on grouping strategy for ADNI on AD vs. CN task (mean$_{\pm\text{std}}$ over 5 folds).}
\centering
\tiny
\setlength{\tabcolsep}{5pt}
\renewcommand{\arraystretch}{1}
\begin{tabular}{lccc}
\toprule
\textbf{Model} & \textbf{Acc. (\%)} & \textbf{BAcc. (\%)} & \textbf{AUC (\%)} \\
\midrule

Random Grouping  & $75.58_{\pm 1.89}$ & $57.77_{\pm 3.53}$ & $70.31_{\pm 2.58}$ \\
3D block Grouping   & $78.2_{\pm 2.09}$ & $77.23_{\pm 1.54}$ & $84.15_{\pm 2.45}$ \\
\rowcolor{blue!8} P2P-SA    & $\mathbf{88.54_{\pm 2.12}}$ & $\mathbf{84.21_{\pm 3.84}}$ & $\mathbf{91.18_{\pm 2.86}}$ \\

\bottomrule
\end{tabular}
\vspace{-2em}
\label{tab:grouping_ablation}

\end{wraptable}
\textit{(ii) 3D block grouping:}
In the second baseline, all foreground brain tokens are partitioned into a fixed set of $69$ spatially contiguous groups using a geometric block-based strategy. Starting from the complete set of valid brain tokens, the token set is recursively split along the spatial axis with the largest extent, using median cuts, until exactly $69$ groups are obtained. This produces groups that are spatially coherent but not aligned to neuroanatomical boundaries. The block-grouping baseline therefore tests whether coarse spatial locality alone is sufficient, or whether explicit anatomical organization is necessary for the observed performance gains.

Together, these two baselines serve as complementary controls: random grouping removes both anatomical and spatial structure, whereas block grouping preserves spatial contiguity but discards anatomical meaning. As shown in Table~\ref{tab:combined_adni_ablations},~\ref{tab:grouping_ablation} and Fig~\ref{fig:grouping}, the anatomy-aware grouping consistently outperforms both baselines, indicating that the gains of the proposed framework are not simply due to token partitioning or locality. Rather, the results suggest that anatomically meaningful grouping provides a beneficial inductive bias, enabling the model to aggregate information within biologically coherent regions and learn representations that better reflect brain organization.

\subsection{Ablations on Age effect}
\label{supp:age_effect}

\begin{wraptable}{r}{0.5\linewidth}
\centering
\vspace{-1.8em}
\caption{Ablation of age and temporal encoding on ADNI.}
\label{tab:age_ablation}
\resizebox{\linewidth}{!}{
\begin{tabular}{llcc}
\toprule
\textbf{Model} &
\textbf{Age encoding} &
\multicolumn{2}{c}{\textbf{ADNI [AD/CN]}} \\
\cmidrule(lr){3-4}
& &
\textbf{BAcc (\%)} &
\textbf{AUC (\%)} \\
\midrule

Age-alone & Age features only
& $57.01\pmstd{0.93}$
& $75.99\pmstd{2.30}$ \\

P2P-SL & None
& $86.46\pmstd{2.61}$
& $90.15\pmstd{2.55}$ \\

P2P-SL & Relative time
& $88.98\pmstd{3.42}$
& $94.20\pmstd{2.36}$ \\

\rowcolor{blue!8} P2P-SL & Absolute age
& $89.20\pmstd{2.21}$
& $94.84\pmstd{1.66}$ \\

P2P-PL & None
& $87.18\pmstd{3.41}$
& $92.41\pmstd{3.41}$ \\

P2P-PL & Relative time
& $88.16\pmstd{3.19}$
& $93.98\pmstd{2.86}$ \\

\rowcolor{blue!8} P2P-PL & Absolute age
& $88.27\pmstd{3.80}$
& $94.58\pmstd{1.57}$ \\

\bottomrule
\end{tabular}
}
\end{wraptable}
\paragraph{Effect of age and temporal encoding.}
To disentangle the contribution of age from longitudinal imaging information, 
we evaluate age alone, MRI representations without temporal encoding, relative 
time from the first visit, and absolute-age encoding. The results are summarized 
in Table~\ref{tab:age_ablation}.
Age alone is only weakly discriminative, whereas both P2P variants remain strongly 
predictive even when age information is removed entirely. Relative temporal encoding 
recovers nearly all of the benefit obtained from absolute-age encoding, indicating 
that the temporal signal is primarily useful for representing visit ordering and 
inter-visit intervals rather than serving as a proxy for disease status. Absolute 
age therefore provides only modest additional biological context and is not the 
primary source of predictive performance.

\begin{wraptable}{r}{0.60\linewidth}
\centering
\vspace{-1.8em}
\caption{External-cohort generalization under relative-time and absolute-age encoding.}
\label{tab:age_external_generalization}
\resizebox{\linewidth}{!}{
\begin{tabular}{llcccc}
\toprule
\textbf{Model} &
\textbf{Age encoding} &
\multicolumn{2}{c}{\textbf{AIBL [AD/CN]}} &
\multicolumn{2}{c}{\textbf{MIRIAD [AD/CN]}} \\
\cmidrule(lr){3-4}
\cmidrule(lr){5-6}
& &
\textbf{BAcc (\%)} &
\textbf{AUC (\%)} &
\textbf{BAcc (\%)} &
\textbf{AUC (\%)} \\
\midrule

P2P-SL & Relative time
& $83.13\pmstd{3.12}$
& $91.02\pmstd{2.32}$
& $89.14\pmstd{3.48}$
& $94.31\pmstd{3.17}$ \\

P2P-SL & Absolute age
& $82.75\pmstd{3.07}$
& $90.58\pmstd{2.18}$
& $90.83\pmstd{4.19}$
& $93.42\pmstd{3.62}$ \\

P2P-PL & Relative time
& $82.78\pmstd{3.62}$
& $90.67\pmstd{2.86}$
& $89.01\pmstd{4.01}$
& $94.46\pmstd{3.09}$ \\

P2P-PL & Absolute age
& $82.02\pmstd{3.13}$
& $90.04\pmstd{3.01}$
& $90.21\pmstd{3.15}$
& $94.02\pmstd{3.52}$ \\

\bottomrule
\end{tabular}
}
\end{wraptable}
\paragraph{Generalization across age distributions.}
We additionally examine whether this behavior persists under cohort shift. The 
evaluated cohorts span different age distributions, from 
$69.89\pmstd{6.98}$ years in MIRIAD to $74.18\pmstd{7.44}$ years in ADNI-MCI. 
Using the trained checkpoints directly for inference on the external AIBL and 
MIRIAD cohorts, we obtain the results in Table~\ref{tab:age_external_generalization}.
Performance remains consistent across the external cohorts under both temporal 
encoding strategies, despite differences in their age distributions. Importantly, 
relative-time encoding transfers comparably to absolute-age encoding, and is superior 
for several metrics. 

Taken together with the age-only and no-age ablations, this 
supports that P2P primarily exploits longitudinal MRI information and transferable 
temporal structure, rather than relying on absolute age as a predictive shortcut.

\subsection{Role of soft parcel assignment}
For each patch $i$, the probabilistic atlas provides a parcel-membership weight 
$m_{ip}$ for parcel $p$. A patch is included in the context of parcel $p$ when
$m_{ip}>\tau$. Consequently, a boundary patch that overlaps multiple parcels can
participate in more than one parcel-specific context. For example, if a patch
overlaps parcels $p$ and $q$, it is contextualized independently within both
corresponding token sets. Because self-attention is performed independently within each parcel context before aggregation, the resulting representations are parcel-conditioned rather than identical copies of the same patch representation. Information from one parcel context is therefore not directly mixed into another. This differs from spatial smoothing, in which neighboring features are explicitly averaged or blended. We further evaluated a late-grouping variant in which self-attention is performed first over a shared token context and parcel grouping is applied only afterward using the atlas-membership matrix. As reported in
Table~\ref{tab:soft_assignment_ablation}, this formulation performs substantially
worse, supporting the importance of performing parcel-specific contextualization
before regional aggregation.

Soft assignment is particularly relevant at the $8^3$ patch resolution because
a boundary patch may genuinely contain voxels from more than one anatomical
region due to partial-volume effects, inter-subject anatomical variability, and
small residual registration differences. Hard argmax assignment discards this
information by forcing each patch into a single parcel, even when multiple parcel
memberships are substantial. This may also distort the available context for
small-volume parcels.

\begin{wraptable}{r}{0.5\linewidth}
\centering
\vspace{-2em}
\caption{Effect of progressively removing low-probability parcel memberships and
replacing soft assignment with hard argmax assignment on ADNI AD/CN classification.}
\label{tab:soft_assignment_ablation}
\resizebox{\linewidth}{!}{
\begin{tabular}{lccc}
\toprule
\textbf{Assignment} &
\textbf{Membership removed (\%)} &
\textbf{BAcc (\%)} &
\textbf{AUC (\%)} \\
\midrule

Soft & 0
& $84.21\pmstd{3.84}$
& $91.18\pmstd{2.86}$ \\

Soft & 10
& $84.21\pmstd{3.84}$
& $91.18\pmstd{2.86}$ \\

Soft & 20
& $84.21\pmstd{3.84}$
& $91.18\pmstd{2.86}$ \\

Soft & 30
& $75.82\pmstd{3.99}$
& $84.49\pmstd{2.98}$ \\

Soft & 50
& $68.85\pmstd{2.96}$
& $78.51\pmstd{2.99}$ \\

Soft & 90
& $52.00\pmstd{1.41}$
& $60.88\pmstd{2.98}$ \\

Hard argmax & --
& $72.76\pmstd{3.66}$
& $85.10\pmstd{3.22}$ \\

\bottomrule
\end{tabular}
}
\end{wraptable}

Removing up to 20\% of the weakest memberships leaves performance unchanged,
showing that the model does not benefit from indiscriminately retaining negligible
parcel overlaps. Performance deteriorates only once increasingly substantial
memberships are discarded, while hard argmax assignment also produces a clear
drop. These findings indicate that the benefit of soft assignment is not explained
by generic boundary smoothing. Rather, it arises from preserving anatomically
meaningful information in boundary patches and allowing those patches to be
contextualized within each parcel to which they substantially contribute. This
is especially relevant for boundary regions that may contain disease-related
atrophy cues or contribute important context to small anatomical parcels.

\begin{table}[t]
\centering
\scriptsize

\begin{minipage}[t]{0.48\textwidth}
\centering
\scriptsize
\caption{Training and hyperparameter configuration details for Stage A}
\label{tab:train_hparams1}
\begin{tabular}{ll}
\toprule
\textbf{Aspect} & \textbf{Value} \\
\midrule
Backbone & ViT-base, pre-norm \\
Patch size & $8 \times 8 \times 8$ \\
Encoder layers & $12$ \\
Encoder heads & $12$ \\
Embed dim & $768$ \\
FFN hidden dim & 2048\\
Dropout & $0.10$ \\
GeM pooling & $p{=}2$ \\
Parcel drop prob & $0.10$ \\
Optimizer & AdamW \\
LR & $1\!\times\!10^{-4}$ \ \\
Weight decay & $1\!\times\!10^{-4}$ \\
Scheduler & Cosine , min LR: $10^{-6}$ \\
Label smoothing & $0.05$ \\
Accumulation steps & $5$ \\
Batch size & $12$ \\
Epochs & $20$ \\
\bottomrule
\end{tabular}
\end{minipage}
\hfill
\begin{minipage}[t]{0.48\textwidth}
\caption{Training and hyperparameter configuration details for Stage B}
\label{tab:train_hparams2}
\scriptsize
\begin{tabular}{ll}
\toprule
\textbf{Aspect} & \textbf{Value} \\
\midrule
Backbone & 2-layer transformer \\
Encoder layers & $2$ \\
Encoder heads & $8$ \\
MLP ratio & $8$ \\
Embed dim & $768$ \\
Dropout & $0.10$ \\
Optimizer & AdamW \\
LR & $3\!\times\!10^{-4}$ \ \\
Weight decay & $3\!\times\!10^{-3}$ \\
Scheduler & Cosine , min LR: $10^{-6}$ \\
Batch size & $64$ \\
Epochs & $20$ \\
\bottomrule
\end{tabular}
\end{minipage}

\end{table}

\subsection{Training setup and Model details}
\label{Asec:train_details}
The training details of the experiments conducted for Stage A and Stage B are in Table~\ref{tab:train_hparams1} and Table~\ref{tab:train_hparams2}, respectively.  The full end-to-end longitudinal model has 
$\mathbf{77\text{M}}$ trainable parameters and integrates a per-scan 
parcel-level encoder with a temporal Transformer operating over 
variable-length visit sequences. Using the max timepoints ($T=15$), ptflops estimates a per-subject forward-pass cost of 
$\mathbf{3.72\times 10^{13}}$ FLOPs. The inference time FLOPs presented in Fig~\ref{fig:overview}B plots for all baselines, prior works and P2P variants are reported in Table~\ref{tab:p2p_baseline_comparison}. 
Note that the 2-stage framework of P2P is only for training and not for inference. This was undertaken to overcome the batch-size differences to accomodate the varying lengths of scans during training.

\begin{figure}[t]
  \centering
   \includegraphics[width=\linewidth]{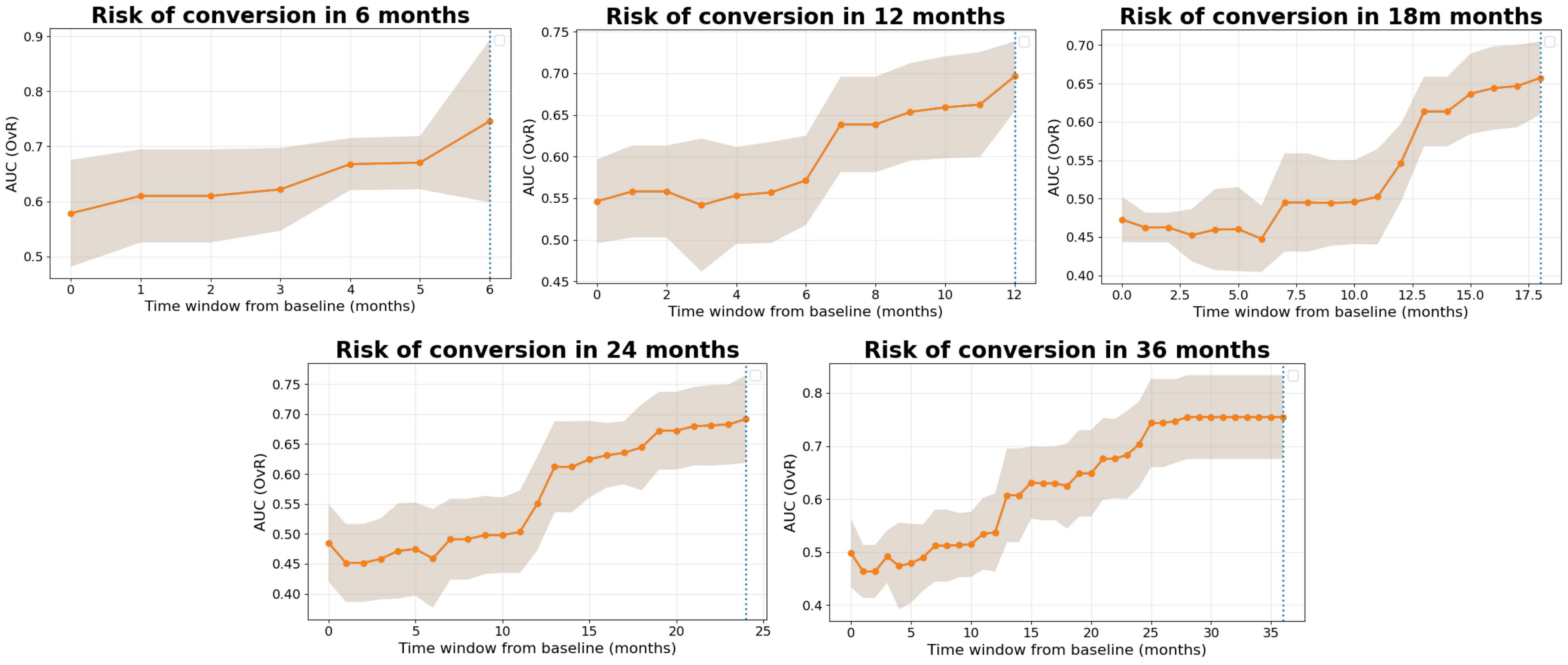}
   \caption{\textbf{Risk of conversion time.} AUC (mean$_{\pm \text{std}}$ over 5 folds) plots for MCI-to-AD conversion-time prediction (in months) using ADNI's pMCI subjects with increasing window from baseline visit. Across the conversion horizons, discriminability increases with added follow-up, with larger gains closer to conversion.}
   \label{fig:mci_risk}
   \vspace{-0.8em}
\end{figure}
\vspace{1em}

\subsection{Experiments and Results} 
A detailed report of P2P's performance of all the longitudinal variants including P2P-SL, P2P-PL and P2P-PL$_\text{B}$ is presented in Table~\ref{tab:dataset_detailed_metrics_models_tok_tasks}. We have also reported the results of the P2P using different tokenizers like: MLP and 3D Densenet. All the results are reported as mean$_{\pm \text{std}}$ across all 5-folds.
Note: The comparatively higher standard deviations in all tables arise from fold-to-fold variation in scan counts induced by subject-level partitioning.

\begin{table*}[t]
\centering
\caption{Detailed results by dataset and task for the three Stage~B variants with the Stage~A tokenizer. Metrics are mean$_{\pm\text{std}}$ over 5 folds.}
\scriptsize
\setlength{\tabcolsep}{4pt}
\renewcommand{\arraystretch}{1.05}
\begin{tabular}{cllcccc}
\toprule
\textbf{Dataset \& Task} & \textbf{Model} & \textbf{Tokenizer} & \textbf{Acc (\%)} & \textbf{Bal. Acc (\%)} & \textbf{F1 (\%)} & \textbf{AUC (\%)} \\
\midrule
\multirow{6}{*}{\begin{tabular}{c} \textbf{ADNI}\\ \text{[AD vs CN]} \end{tabular}}
& P2P-PL  & MLP      & $92.96_{\pm 1.61}$ & $88.27_{\pm 3.80}$ & $84.39_{\pm 3.55}$ & $94.58_{\pm 1.57}$ \\
& P2P-$\text{PL}_B$  & MLP      & $92.05_{\pm 2.47}$ & $87.38_{\pm 2.83}$ & $81.36_{\pm 4.33}$ & $94.00_{\pm 2.02}$ \\
& P2P-$\text{SL}$        & MLP      & $\mathbf{93.91_{\pm 1.68}}$ & $89.20_{\pm 2.21}$ & $84.81_{\pm 4.05}$ & $\mathbf{94.84_{\pm 1.66}}$ \\
& P2P-PL  & 3D Densenet & $92.14_{\pm 1.22}$ & $87.93_{\pm 2.23}$ & $82.88_{\pm 4.53}$ & $94.24_{\pm 1.17}$ \\
& P2P-$\text{PL}_B$ & 3D Densenet & $91.46_{\pm 1.25}$ & $87.54_{\pm 1.47}$ & $82.42_{\pm 2.76}$ & $93.55_{\pm 1.22}$ \\
& P2P-$\text{SL}$        & 3D Densenet & $93.01_{\pm 1.15}$ & $\mathbf{89.90_{\pm 1.74}}$ & $\mathbf{85.25_{\pm 3.19}}$ & $94.71_{\pm 0.78}$ \\
\midrule
\multirow{6}{*}{\begin{tabular}{c} \textbf{ADNI}\\ \text{[sMCI vs pMCI]} \end{tabular}}
& P2P-PL  & MLP      & $84.61_{\pm 1.13}$ & $76.65_{\pm 1.90}$ & $67.02_{\pm 1.98}$ & $85.86_{\pm 0.60}$ \\
& P2P-$\text{PL}_B$ & MLP      & $82.73_{\pm 1.65}$ & $76.94_{\pm 2.52}$ & $67.08_{\pm 3.11}$ & $82.02_{\pm 1.41}$ \\
& P2P-$\text{SL}$            & MLP      & $\mathbf{84.62_{\pm 1.00}}$ & $\mathbf{78.26_{\pm 0.91}}$ & $\mathbf{68.55_{\pm 1.53}}$ & $\mathbf{85.96_{\pm 1.18}}$ \\
& P2P-PL  & 3D Densenet & $84.27_{\pm 1.57}$ & $76.09_{\pm 1.90}$ & $66.20_{\pm 2.68}$ & $83.72_{\pm 3.73}$ \\
& P2P-$\text{PL}_B$ & 3D Densenet & $81.97_{\pm 1.60}$ & $73.36_{\pm 2.65}$ & $62.07_{\pm 3.54}$ & $83.15_{\pm 2.29}$ \\
& P2P-$\text{SL}$        & 3D Densenet & $84.54_{\pm 1.36}$ & $77.51_{\pm 2.01}$ & $67.71_{\pm 2.06}$ & $85.52_{\pm 1.88}$ \\
\midrule
\multirow{6}{*}{\begin{tabular}{c} \textbf{AIBL} \\ \text{[AD vs CN]} \end{tabular}}
& P2P-PL  & MLP      & $\mathbf{95.39_{\pm 0.57}}$ & $\mathbf{88.38_{\pm 3.09}}$ & $\mathbf{81.53_{\pm 4.24}}$ & $92.21_{\pm 3.06}$ \\
& P2P-$\text{PL}_B$ & MLP      & $94.96_{\pm 1.43}$ & $86.57_{\pm 4.42}$ & $79.66_{\pm 4.43}$ & $91.14_{\pm 4.17}$ \\
& P2P-$\text{SL}$             & MLP      & $95.13_{\pm 0.59}$ & $86.86_{\pm 3.47}$ & $80.37_{\pm 3.19}$ & $90.89_{\pm 4.69}$ \\
& P2P-PL  & 3D Densenet & $93.68_{\pm 1.78}$ & $84.82_{\pm 3.80}$ & $75.88_{\pm 4.38}$ & $92.13_{\pm 3.36}$ \\
& P2P-$\text{PL}_B$ & 3D Densenet & $93.36_{\pm 0.98}$ & $81.84_{\pm 3.47}$ & $72.46_{\pm 4.81}$ & $\mathbf{93.11_{\pm 1.71}}$ \\
& P2P-$\text{SL}$             & 3D Densenet & $93.82_{\pm 1.59}$ & $84.85_{\pm 3.16}$ & $75.41_{\pm 4.65}$ & $90.53_{\pm 4.39}$ \\
\midrule
\multirow{6}{*}{\begin{tabular}{c} \textbf{MIRIAD}\\ \text{[AD vs CN]} \end{tabular}}
& P2P-PL  & MLP      & $96.97_{\pm 4.41}$ & $96.54_{\pm 3.58}$ & $97.57_{\pm 4.15}$ & $98.33_{\pm 3.00}$ \\
& P2P-$\text{PL}_B$ & MLP      & $92.93_{\pm 4.31}$ & $93.16_{\pm 3.81}$ & $94.14_{\pm 3.08}$ & $96.96_{\pm 3.56}$ \\
& P2P-$\text{SL}$             & MLP      & $97.41_{\pm 3.45}$ & $96.90_{\pm 3.55}$ & $98.14_{\pm 3.89}$ & $98.05_{\pm 3.22}$ \\
& P2P-PL  & 3D Densenet & $95.73_{\pm 4.69}$ & $95.47_{\pm 3.98}$ & $96.57_{\pm 3.83}$ & $98.00_{\pm 2.58}$ \\
& P2P-$\text{PL}_B$ & 3D Densenet & $97.20_{\pm 3.93}$ & $96.43_{\pm 4.05}$ & $97.86_{\pm 3.00}$ & $\mathbf{98.40_{\pm 1.11}}$ \\
& P2P-$\text{SL}$            & 3D Densenet & $\mathbf{98.05_{\pm 2.48}}$ & $\mathbf{98.44_{\pm 2.00}}$ & $\mathbf{98.39_{\pm 2.17}}$ & $98.33_{\pm 1.11}$ \\
\bottomrule
\end{tabular}
\label{tab:dataset_detailed_metrics_models_tok_tasks}
\end{table*}

\subsubsection{MCI Prognosis}
We perform prognosis of MCI to AD conversion as a binary sMCI vs.\ pMCI classification task (Table~\ref{tab:sota_comparison_adni_split}). Further, we investigate how early the conversion trajectory within the pMCI population becomes detectable from accumulated longitudinal data. Across five conversion horizons (6, 12, 18, 24, and 36 months), we plot AUC against longitudinal history from baseline (Fig.~\ref{fig:mci_risk}). Across all horizons, while absolute AUC values are moderate, discriminability improves as more longitudinal context accumulates, with notable jumps in the curves before the conversion window, indicating that model captures meaningful cues about the conversion trajectory ahead of the clinical event.

\subsubsection{Survival analysis over P2P longitudinal representations}
\label{Asec:survival}

The primary MCI prognosis experiment in the main paper evaluates progression using the conventional binary formulation of stable MCI versus progressive MCI. While this setting is widely used, it compresses a clinically time-dependent process into a static label. In practice, MCI-to-AD progression is naturally a time-to-event problem i.e., a subject may remain stable for a certain follow-up duration, convert later, or leave the study before conversion is observed. Therefore, we performed an additional censoring-aware survival analysis to examine whether P2P representations contain prognostic information about not only whether a subject may convert, but also when conversion is likely to occur using the P2P-SL framework.

\begin{figure}[t]
  \centering
   \includegraphics[width=\linewidth]{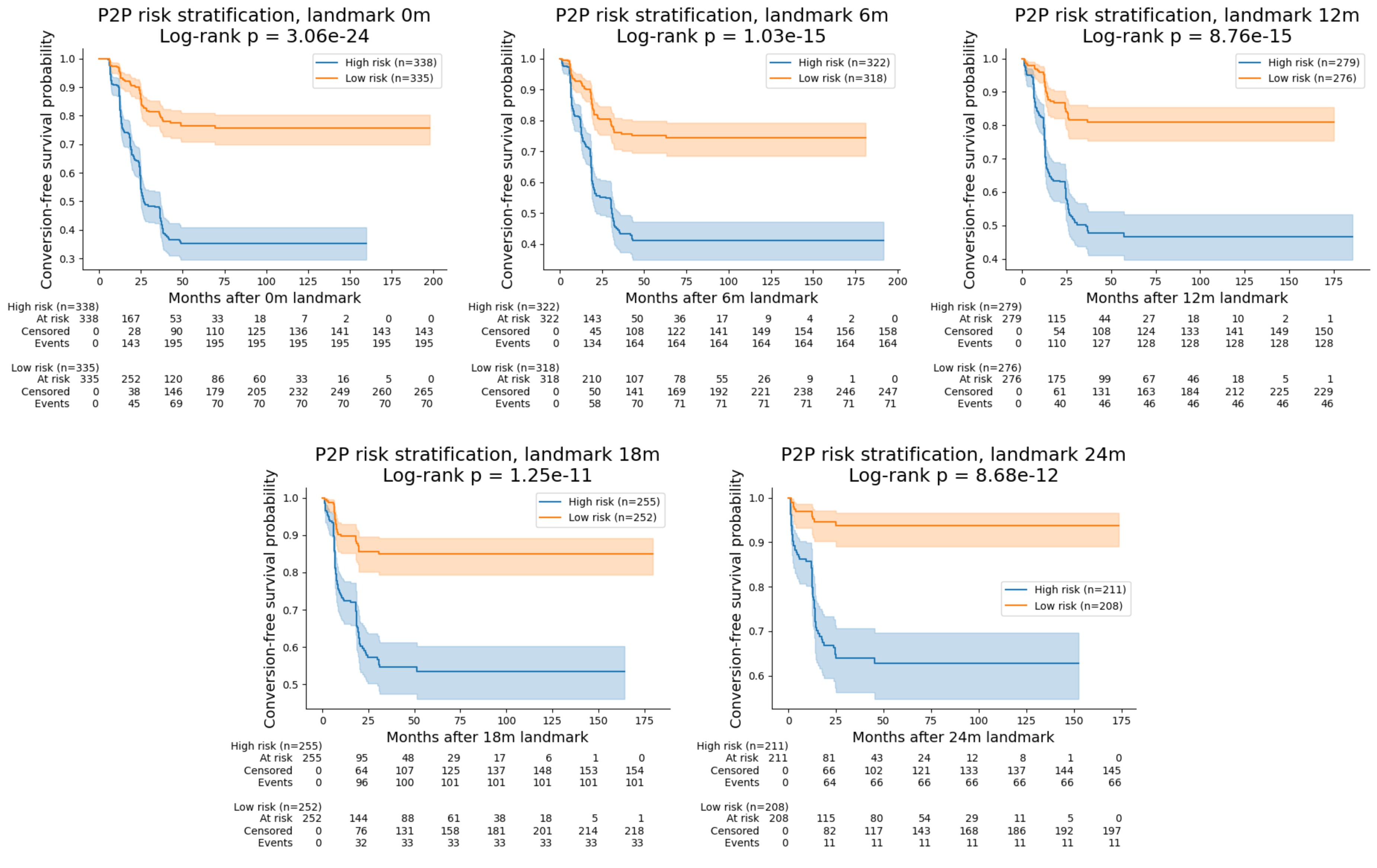}
   \caption{\textbf{Kaplan-Meier curves} at landmarks 0, 6, 12, 18, 24 months showing the separation of the high and low risk MCI subjects from ADNI.}
   \label{fig:KM_curve}
   \vspace{-0.8em}
\end{figure}
\vspace{1em}

\paragraph{Setting:}
For each subject, we constructed a subject-level survival record from the scan-level ADNI metadata. The model uses longitudinal MRI information available up to a predefined landmark month and predicts future conversion risk after that landmark. At a landmark $L$, all visits with visit month less than or equal to $L$ are used as model input. Subjects who converted before or at the landmark are excluded, because their event has already occurred before the prediction origin. For subjects with observed conversion, the survival duration is computed as conversion time minus landmark time. For censored subjects, the duration is computed as last known follow-up or censoring time minus landmark time.


\paragraph{Methodology:}
At each landmark $t_L \in \{0, 6, 12, 18, 24\}$ months, subjects not yet converted form the at-risk cohort; non-converters are right-censored. We extract the \emph{same frozen} P2P parcel embeddings used for classification. After PCA (95\% variance retained), a Cox proportional hazards~\cite{cox1972regression} model is fit per landmark within 5-fold CV:
\begin{equation}
    \lambda(t \mid \mathbf{z}_i) = \lambda_0(t) \exp(\boldsymbol{\beta}^\top \mathbf{z}_i)
\end{equation}
All reported metrics use out-of-fold validation predictions. For KM stratification, subjects are split into high/low risk using the median within each fold before pooling, avoiding fold-specific scaling artifacts. We evaluate using: (i)~Harrell's C-index~\cite{harrell1982evaluating}, (ii)~cumulative/dynamic time-dependent AUC at horizons $\Delta t \in \{12,18,24,30,36\}$ months, and (iii)~Kaplan-Meier~\cite{kaplan1958nonparametric} stratification with log-rank testing. Note that AUC@$\Delta t$ at landmark $t_L$ predicts conversion by $t_L + \Delta t$ from study baseline.
 
 
\paragraph{Concordance index:}
Table~\ref{tab:cindex_landmarks} reports C-index across landmarks. The C-index measures whether the model correctly ranks subject pairs by survival time, where higher predicted risk should correspond to earlier conversion. Performance generally increases with landmark time from $69.1\%$ at baseline to $76.7\%$ at 24m, indicating that additional longitudinal history improves prognostic ranking. 
 
\begin{table}[h]
\centering
\small
\caption{C-index across landmarks (mean$_{\pm \text{std}}$, 5-fold CV).}
\label{tab:cindex_landmarks}
\begin{tabular}{ccccc}
\toprule
Landmark & $N$ at risk & Events & Censored & C-index (\%) \\
\midrule
0m  & 640 & 235 & 405 & $69.1_{\pm 4.7}$ \\
12m & 555 & 174 & 381 & $70.6_{\pm 6.1}$ \\
24m & 419 & 77  & 342 & $76.7_{\pm 8.3}$ \\
\bottomrule
\end{tabular}
\end{table}
 
 
\paragraph{Time-dependent AUC:}
While the C-index evaluates overall ranking, time-dependent AUC assesses discrimination at specific future horizons: can the predicted risk separate subjects who convert within $\Delta t$ months from those who remain event-free beyond that horizon? Table~\ref{tab:tdauc_full} reports AUC across all landmark--horizon combinations. Baseline from short-term prediction (AUC@12m ${\sim}66\%$), improves to ${\sim}73\%$ at longer horizons. Later landmarks achieve consistently higher AUCs: 70--75\% at 18m and 77--81\% at 24m. Comparing the same absolute endpoint across landmarks (e.g., conversion by month 42 from baseline: landmark 18m AUC@24m $= 75.5\%$ vs.\ landmark 24m AUC@18m $= 79.3\%$) confirms that additional MRI history improves prediction of the same clinical outcome.
 
\begin{table}[h]
\centering
\small
\caption{Time-dependent AUC (\%) across landmarks and prediction horizons (mean$_{\pm \text{std}}$, 5-fold CV). All results use frozen P2P embeddings without clinical covariates.}
\label{tab:tdauc_full}
\begin{tabular}{cccccc}
\toprule
Landmark & AUC@12m & AUC@18m & AUC@24m & AUC@30m & AUC@36m \\
\midrule
0m  & $66.2_{\pm 5.4}$ & $70.3_{\pm 4.2}$ & $72.6_{\pm 4.8}$ & $72.9_{\pm 7.1}$ & $72.9_{\pm 7.0}$ \\
18m & $69.5_{\pm 8.6}$ & $70.1_{\pm 8.6}$ & $75.5_{\pm 9.0}$ & $75.2_{\pm 8.8}$ & $75.3_{\pm 11.0}$ \\
24m & $77.4_{\pm 15.6}$ & $79.3_{\pm 8.1}$ & $79.4_{\pm 7.5}$ & $79.8_{\pm 8.5}$ & $80.6_{\pm 8.8}$ \\
\bottomrule
\end{tabular}
\begin{flushleft}

\end{flushleft}
\end{table}
 
 
\paragraph{Kaplan-Meier stratification:}
Finally, we assessed whether the continuous risk scores translate to clinically meaningful patient-group separation. Figure~\ref{fig:KM_curve} shows KM curves at landmarks 0, 6, 12, 18, and 24 months. The high-risk group exhibits rapid decline in conversion-free survival probability soon after each landmark, while the low-risk group maintains substantially higher event-free probability.
Table~\ref{tab:km_summary} quantifies the separation: the event ratio between high- and low-risk groups increases monotonically from 2.3:1 at 6m to 6.0:1 at 24m, with all log-rank p-values below $10^{-11}$. At landmark 24m, only 5\% of low-risk subjects converted versus ${\sim}$31\% in the high-risk group, the low-risk survival curve remains above 0.90 throughout follow-up.
 
\begin{table}[h]
\centering
\small
\caption{KM stratification summary across landmarks. Event ratio = high-risk events / low-risk events.}
\label{tab:km_summary}
\begin{tabular}{cccccccc}
\toprule
Landmark & $N$ & High $n$ & Low $n$ & High events & Low events & Ratio & Log-rank $p$ \\
\midrule
6m  & 640 & 322 & 318 & 164 & 71 & 2.3:1 & $1.0{\times}10^{-15}$ \\
12m & 555 & 279 & 276 & 128 & 46 & 2.8:1 & $8.8{\times}10^{-15}$ \\
18m & 507 & 255 & 252 & 101 & 33 & 3.1:1 & $1.3{\times}10^{-11}$ \\
24m & 419 & 211 & 208 & 66  & 11 & 6.0:1 & $8.7{\times}10^{-12}$ \\
\bottomrule
\end{tabular}
\end{table}
 
 
\begin{table}[t]
\centering
\caption{Landmark-based survival evaluation on ADNI. Left: C-index across landmarks.
Right: time-dependent AUC at 0- and 24-month landmarks. Results are mean over 5 folds.}
\label{tab:landmark_survival}
\begin{minipage}[t]{0.40\linewidth}
\centering
\resizebox{\linewidth}{!}{
\begin{tabular}{lccc}
\toprule
\textbf{Model} &
\textbf{0m} &
\textbf{12m} &
\textbf{24m} \\
\midrule

DenseNet121+GRU
& $63.13$
& $66.88$
& $63.53$ \\

DenseNet121+LT
& $64.19$
& $66.22$
& $64.51$ \\

3D ViT+LT
& $64.04$
& $65.15$
& $63.85$ \\

\rowcolor{blue!8} \textbf{P2P-SL}
& $\mathbf{69.15}$
& $\mathbf{70.62}$
& $\mathbf{76.65}$ \\

\bottomrule
\end{tabular}
}

\vspace{2pt}
\small
\textit{C-index across landmarks}
\end{minipage}
\hfill
\begin{minipage}[t]{0.57\linewidth}
\centering
\resizebox{\linewidth}{!}{
\begin{tabular}{llccc}
\toprule
\textbf{Landmark} &
\textbf{Model} &
\textbf{AUC@12m} &
\textbf{AUC@24m} &
\textbf{AUC@36m} \\
\midrule

0m & DenseNet121+GRU
& $65.58$
& $66.26$
& $64.92$ \\

0m & DenseNet121+LT
& $66.06$
& $67.88$
& $65.66$ \\

0m & 3D ViT+LT
& $65.90$
& $66.33$
& $65.05$ \\

\rowcolor{blue!8} 0m & \textbf{P2P-SL}
& $\mathbf{66.16}$
& $\mathbf{72.64}$
& $\mathbf{72.93}$ \\

\midrule

24m & DenseNet121+GRU
& $61.42$
& $67.74$
& $72.10$ \\

24m & DenseNet121+LT
& $60.94$
& $69.57$
& $70.13$ \\

24m & 3D ViT+LT
& $59.09$
& $64.51$
& $69.14$ \\

\rowcolor{blue!8} 24m & \textbf{P2P-SL}
& $\mathbf{77.41}$
& $\mathbf{79.42}$
& $\mathbf{80.62}$ \\

\bottomrule
\end{tabular}
}

\vspace{2pt}
\small
\textit{Time-dependent AUC}
\end{minipage}
\end{table}

Table~\ref{tab:landmark_survival} demonstrates that across landmarks, the longitudinal baselines exhibit flat or non-monotonic
C-index trajectories, whereas P2P-SL improves monotonically from the 0-month
to the 24-month landmark. Its margin over the strongest baseline increases
from $4.96$ percentage points at 0 months to $12.14$ percentage points at
24 months. This indicates that P2P-SL increasingly benefits from accumulated
longitudinal history: when anatomical correspondence is preserved across
visits, evidence can accumulate within each parcel to form a more informative
disease trajectory.

Overall, these results suggest that P2P representations support a time-to-event formulation of MCI prognosis beyond static sMCI/pMCI classification. The model provides meaningful early risk stratification, while additional longitudinal history appears to refine prognostic ranking and horizon-specific discrimination under censoring-aware evaluation.

\begin{figure}[t]
  \centering
   \includegraphics[width=\linewidth]{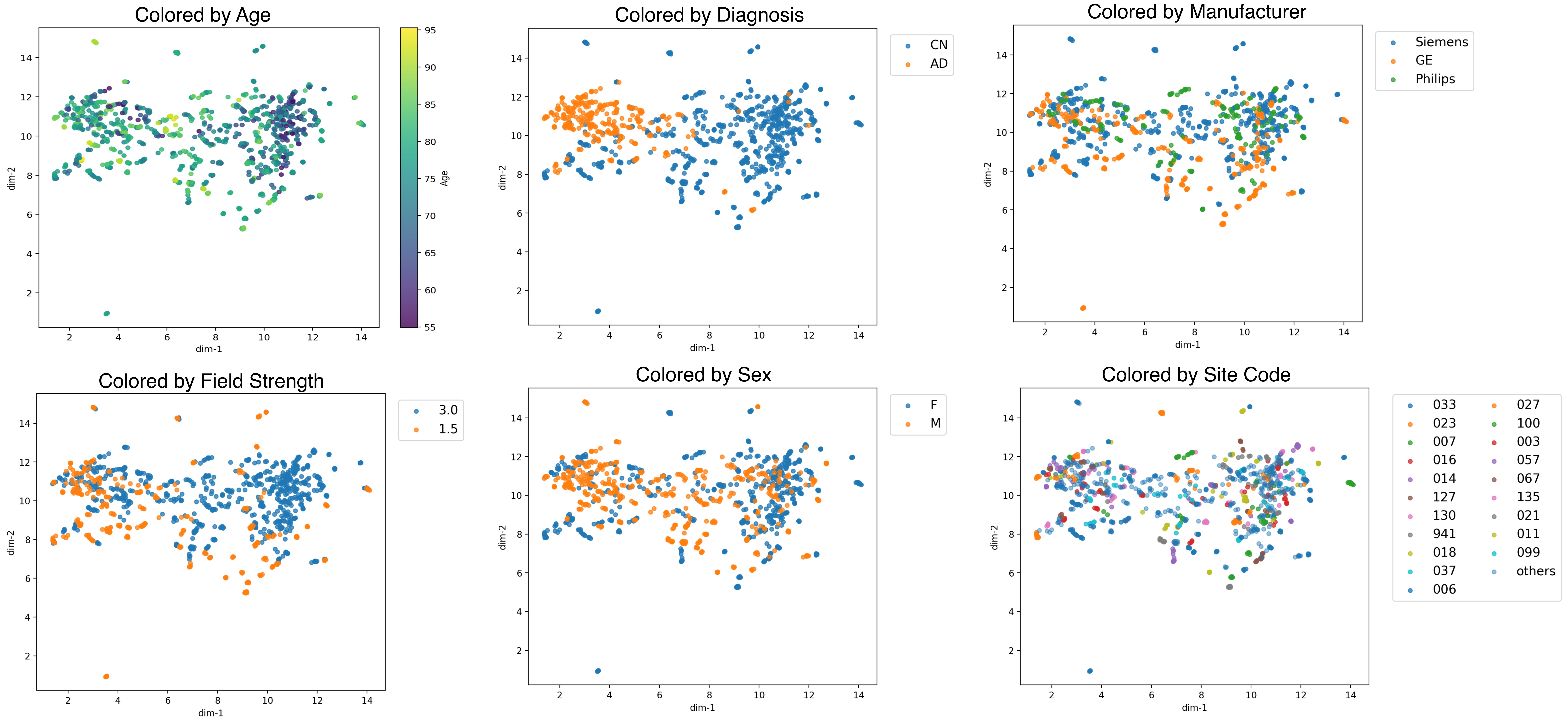}
   \caption{UMAPs of frozen scan-level embeddings from the parcel encoder on the test set, coloured by diagnosis, demographic or acquisition variables, including age, sex, field strength, manufacturer, and site.}
   \label{fig:umap_embed}
   \vspace{-0.8em}
\end{figure}
\vspace{1em}

\begin{table*}[!htb]
\centering
\scriptsize

\begin{minipage}[t]{\textwidth}
\centering
\setlength{\tabcolsep}{3pt}
\scriptsize
\caption{Diagnosis \& confound predictability from embeddings, and embedding confound dependence before and after confound residualization. Results are reported as mean$_{\pm\mathrm{std}}$ over 5 folds. Classification tasks use BACC and AUC, age uses MAE, and Pearson $r$, and dependence is measured using HSIC. Each row in the table corresponds to a separate prediction task, where the target is the diagnosis/confounder itself (for example, sex, field strength, or manufacturer), evaluated before and after residualization.}
\label{tab:diag_confound_hsic_combined}
\begin{tabular}{llccc}
\toprule
Task & Metric & Before & After & $\Delta$ \\
\midrule
\multirow{2}{*}{Diagnosis} & BACC & $0.84_{\pm0.04}$ & $0.84_{\pm0.04}$ & $-0.00_{\pm0.02}$ \\
 & AUC  & $0.91_{\pm0.03}$ & $0.91_{\pm0.03}$ & $-0.00_{\pm0.01}$ \\
\midrule
\multirow{2}{*}{Age} & MAE $\downarrow$ & $4.70_{\pm0.18}$ & $5.94_{\pm0.04}$ & $+1.24_{\pm0.19}$ \\
 & Pearson $r$ & $0.63_{\pm0.02}$ & $-0.52_{\pm0.06}$ & $-1.15_{\pm0.06}$ \\
\midrule
\multirow{2}{*}{Sex} & BACC & $0.85_{\pm0.01}$ & $0.22_{\pm0.03}$ & $-0.63_{\pm0.03}$ \\
 & AUC  & $0.93_{\pm0.01}$ & $0.14_{\pm0.03}$ & $-0.79_{\pm0.03}$ \\
\midrule
\multirow{2}{*}{Manufacturer} & BACC & $0.83_{\pm0.04}$ & $0.08_{\pm0.02}$ & $-0.76_{\pm0.03}$ \\
 & AUC  & $0.95_{\pm0.02}$ & $0.20_{\pm0.04}$ & $-0.75_{\pm0.02}$ \\
\midrule
\multirow{2}{*}{Field Strength} & BACC & $0.96_{\pm0.01}$ & $0.35_{\pm0.02}$ & $-0.61_{\pm0.02}$ \\
 & AUC  & $0.99_{\pm0.00}$ & $0.28_{\pm0.03}$ & $-0.71_{\pm0.03}$ \\
\midrule
\multirow{1}{*}{Site} & BACC & $0.05_{\pm0.02}$ & $0.00_{\pm0.00}$ & $-0.05_{\pm0.02}$ \\
\midrule
Sex & HSIC & $0.003_{\pm0.001}$ & $0.001_{\pm0.000}$ & $-0.001_{\pm0.000}$ \\
Manufacturer & HSIC & $0.002_{\pm0.001}$ & $0.001_{\pm0.000}$ & $-0.001_{\pm0.001}$ \\
Field Strength & HSIC & $0.008_{\pm0.003}$ & $0.001_{\pm0.000}$ & $-0.007_{\pm0.003}$ \\
\bottomrule
\end{tabular}
\end{minipage}
\hfill

\begin{minipage}[t]{0.49\textwidth}
\centering
\caption{Diagnosis performance from embeddings across demographic and acquisition subgroups.}
\label{tab:subgroup_diag}
\setlength{\tabcolsep}{4pt}
\begin{tabular}{llcc}
\toprule
Group & Subgroup & BACC & AUC \\
\midrule
\multirow{2}{*}{Sex} & Female & $0.84_{\pm0.03}$ & $0.91_{\pm0.03}$ \\
 & Male & $0.84_{\pm0.05}$ & $0.90_{\pm0.04}$ \\
\midrule
\multirow{3}{*}{Manufacturer} & GE & $0.83_{\pm0.02}$ & $0.90_{\pm0.02}$ \\
 & Philips & $0.81_{\pm0.09}$ & $0.89_{\pm0.08}$ \\
 & Siemens & $0.85_{\pm0.04}$ & $0.92_{\pm0.04}$ \\
\midrule
\multirow{2}{*}{Field Strength} & 1.5T & $0.81_{\pm0.05}$ & $0.89_{\pm0.04}$ \\
 & 3.0T & $0.85_{\pm0.04}$ & $0.91_{\pm0.04}$ \\
\midrule
\multirow{4}{*}{Age bin} & $\leq 65$ & $0.84_{\pm0.11}$ & $0.90_{\pm0.05}$ \\
 & 66--75 & $0.86_{\pm0.03}$ & $0.93_{\pm0.02}$ \\
 & 76--85 & $0.82_{\pm0.04}$ & $0.90_{\pm0.04}$ \\
 & $>85$ & $0.79_{\pm0.04}$ & $0.90_{\pm0.04}$ \\
\bottomrule
\end{tabular}
\end{minipage}
\hfill
\begin{minipage}[t]{0.49\textwidth}

\centering
\caption{Diagnosis prediction using only metadata confounders.}
\label{tab:metadata_diag}
\setlength{\tabcolsep}{3pt}
\begin{tabular}{lcc}
\toprule
Model & BACC & AUC \\
\midrule
Metadata confounders only & $0.56_{\pm0.02}$ & $0.58_{\pm0.03}$ \\
\bottomrule
\end{tabular}

 \vspace{8pt}

\centering
\caption{Equalized odds gaps from embeddings across confounder groups.}
\label{tab:fairness_embed}
\setlength{\tabcolsep}{4pt}
\begin{tabular}{lcc}
\toprule
Confounder & TPR gap & FPR gap \\
\midrule
Sex & $0.07_{\pm0.04}$ & $0.04_{\pm0.02}$ \\
Manufacturer & $0.14_{\pm0.10}$ & $0.09_{\pm0.03}$ \\
Field Strength & $0.05_{\pm0.03}$ & $0.11_{\pm0.06}$ \\
Age bin & $0.21_{\pm0.16}$ & $0.20_{\pm0.04}$ \\
\bottomrule
\end{tabular}

\end{minipage}

\end{table*}

\begin{table*}[!htb]
\centering

\begin{minipage}{0.49\linewidth}
\centering
\scriptsize
\caption{Subgroup performance of the final diagnosis predictions across demographic and acquisition factors. Values are reported as mean$_{\pm\mathrm{std}}$ over 5 folds.}
\label{tab:subgroup_perf}
\begin{tabular}{clcc}
\toprule
Group & Subgroup & BACC & AUC \\
\midrule

\multirow{2}{*}{Sex} 
& Female & $0.82_{\pm 0.02}$ & $0.91_{\pm 0.02}$ \\
& Male   & $0.84_{\pm 0.06}$ & $0.91_{\pm 0.05}$ \\

\midrule
\multirow{3}{*}{Manufacturer}
& GE      & $0.81_{\pm 0.03}$ & $0.89_{\pm 0.02}$ \\
& Philips & $0.82_{\pm 0.08}$ & $0.90_{\pm 0.09}$ \\
& Siemens & $0.85_{\pm 0.03}$ & $0.92_{\pm 0.04}$ \\

\midrule
\multirow{2}{*}{Field Strength}
& 1.5T & $0.82_{\pm 0.04}$ & $0.89_{\pm 0.04}$ \\
& 3.0T & $0.83_{\pm 0.04}$ & $0.92_{\pm 0.03}$ \\

\midrule
\multirow{4}{*}{Age Bin}
& $\leq65$ & $0.73_{\pm 0.12}$ & $0.87_{\pm 0.09}$ \\
& 66--75   & $0.85_{\pm 0.02}$ & $0.95_{\pm 0.02}$ \\
& 76--85   & $0.84_{\pm 0.05}$ & $0.90_{\pm 0.04}$ \\
& $>85$    & $0.82_{\pm 0.03}$ & $0.86_{\pm 0.04}$ \\

\bottomrule
\end{tabular}
\end{minipage}
\hfill
\begin{minipage}{0.49\linewidth}
\centering
\scriptsize
\caption{Prediction-level fairness summaries measured using Equalized Odds gaps across confounder groups. Lower values indicate smaller disparity.}
\label{tab:fairness}
\begin{tabular}{lcc}
\toprule
Confounder & FPR Gap & TPR Gap \\
\midrule
Sex & $0.03_{\pm 0.02}$ & $0.11_{\pm 0.05}$ \\
Manufacturer & $0.06_{\pm 0.03}$ & $0.16_{\pm 0.09}$ \\
Field Strength & $0.06_{\pm 0.04}$ & $0.07_{\pm 0.04}$ \\
Age Bin & $0.16_{\pm 0.03}$ & $0.19_{\pm 0.22}$ \\
\bottomrule
\end{tabular}
\end{minipage}

\end{table*}

\subsection{Confounder Analysis}
\label{Asec:confounders}

We examined confounding at two complementary levels on the ADNI dataset (AD vs CN task), which covers a wide variability of confounders like sex, manufacturer, age, and field-strength. First, we analyzed the \emph{frozen parcel-level embeddings} produced by the parcel encoder to determine the extent to which the learned representation retained information about non-diagnostic variables and whether it influenced the diagnosis. Second, we evaluated the \emph{final prediction probabilities} to determine whether any dependence on these variables translated into subgroup disparities in classification behaviour. All quantities are reported as mean $\pm$ standard deviation across the 5 folds. Note that the number of sites in the ADNI dataset is varied and high. There are certain sites that contributed very few scans in ADNI cohort. Hence, only BACC results could be reported in Table~\ref{tab:diag_confound_hsic_combined}.

\paragraph{Embedding-level analysis:}
For the representation analysis, parcel embeddings from each scan were stacked and applied PCA to get a single frozen scan-level embedding to avoid loss of embedding information during aggregations. Linear probes were trained on the training-fold embeddings only. Using these embeddings, we evaluated: (i) diagnosis prediction, (ii) prediction of individual confounders, and (iii) a diagnosis baseline based only on confound metadata. 
To ensure that our model’s performance was driven by true pathological signatures rather than demographic or acquisition biases, we performed linear residualization. By regressing out the influence of available confounds from the embeddings, we created a `clean' latent space and then re-evaluated the predictability of both the diagnosis and the confounds. This allowed us to confirm that while the diagnostic signal remained, the embeddings no longer contained information that could reliably identify the confounding variables.

\noindent
The embedding-based diagnosis probe achieved strong performance (Table~\ref{tab:diag_confound_hsic_combined}), confirming that the frozen representation preserves substantial disease-relevant information. In contrast, a diagnosis model trained only on metadata confounders performed worse (Table~\ref{tab:metadata_diag}), indicating that the diagnostic signal captured by the embeddings cannot be explained by demographic and acquisition variables alone.
To examine whether the learned representation encodes diagnostic signal independently of demographic and acquisition-related confounders, we performed a series of post-hoc analyses.

First, we applied linear residualization by fitting a ridge regression from metadata confounders (age, sex, manufacturer, field strength, and site) to the embedding vectors on the training split, and subtracting the predicted confound component from both splits. Table~\ref{tab:diag_confound_hsic_combined} evaluates the prediction of the respective confounder/diagnosis before and after residualization. Nuisance decodability dropped sharply following this adjustment: sex, manufacturer, and field-strength probes showed degraded performance, age-prediction error increased substantially, and the age correlation reversed sign. This reflects strong attenuation of confound structure. In contrast, diagnostic performance remained unchanged, confirming that diagnostic separability in the embeddings is not driven by the removed confound-related variance.
Second, we assessed statistical dependence (non-linearities) between the embedding vectors and the confounder label space using the Hilbert-Schmidt Independence Criterion (HSIC) (Table~\ref{tab:diag_confound_hsic_combined}). 
HSIC, a kernel-based measure of statistical dependence, makes no assumptions about the functional form of the relationship between variables and equals zero if and only if the two distributions are statistically independent, making it a principled and complete test. We applied an RBF kernel over the embedding space and a delta kernel over the categorical confounder labels. Across all tested confounders, the Hilbert--Schmidt Independence Criterion (HSIC) values were uniformly small and decreased further after residualization (Table~\ref{tab:diag_confound_hsic_combined}).  Overall, the embeddings exhibit negligible non-linear confound-related dependence for diagnosis, both before and after linear residualization.
Third, we evaluated diagnostic performance across demographic and acquisition subgroups, (Table~\ref{tab:subgroup_diag}). Performance remained broadly consistent across sex, manufacturer, field strength, and age strata, with only modest variation. This suggests diagnostic behavior does not disproportionately depend on specific demographic or scanner subpopulations.
 Finally, we evaluated equalized odds gaps. This metric directly quantifies whether the model's error structure is consistent across demographic and acquisition conditions, providing a more clinically grounded assessment of fairness.
 Equalized-odds gaps remained small across sex and field strength , confirming stable prediction behaviour across these demographic and acquisition axes 
 in Table~\ref{tab:fairness_embed}. Thus
indicating that prediction behavior is broadly stable across demographic and acquisition conditions.

Taken together, these analyses indicate that although the representation contains measurable demographic and acquisition-related structure, diagnostic performance cannot be explained primarily by these confounders. Diagnostic separability remains stable after confound attenuation, confounders alone provide only weak diagnostic signal, and performance remains broadly consistent across major demographic and acquisition strata.
This qualitative trend is further supported visually by the embedding projections in Fig.~\ref{fig:umap_embed}, where diagnosis remains well structured despite variation across demographic and acquisition-related factors.

\paragraph{Prediction-level analysis:}
To complement the representation analysis, we next examined the final model outputs at the level of predicted probabilities. We evaluated subgroup behaviour across sex, manufacturer, field strength, and age bins using subgroup performance and Equalized Odds gaps.

The subgroup analysis in Tables~\ref{tab:subgroup_perf} and~\ref{tab:fairness} indicates that the final predictions are broadly stable across demographic and acquisition-related factors. Performance remains comparable across sex, scanner manufacturer, field strength, and age bins, suggesting good generalization across clinically and technically relevant subgroups. The final predictions are not dominated by any single confounding axis.

\vspace{1em}
\section{Explainability Analysis in detail}

\subsection{Cohort-level analysis}
\label{Asec:cohort-saliency}

\begin{figure*}[t]
  \centering
  \includegraphics[width=\linewidth]{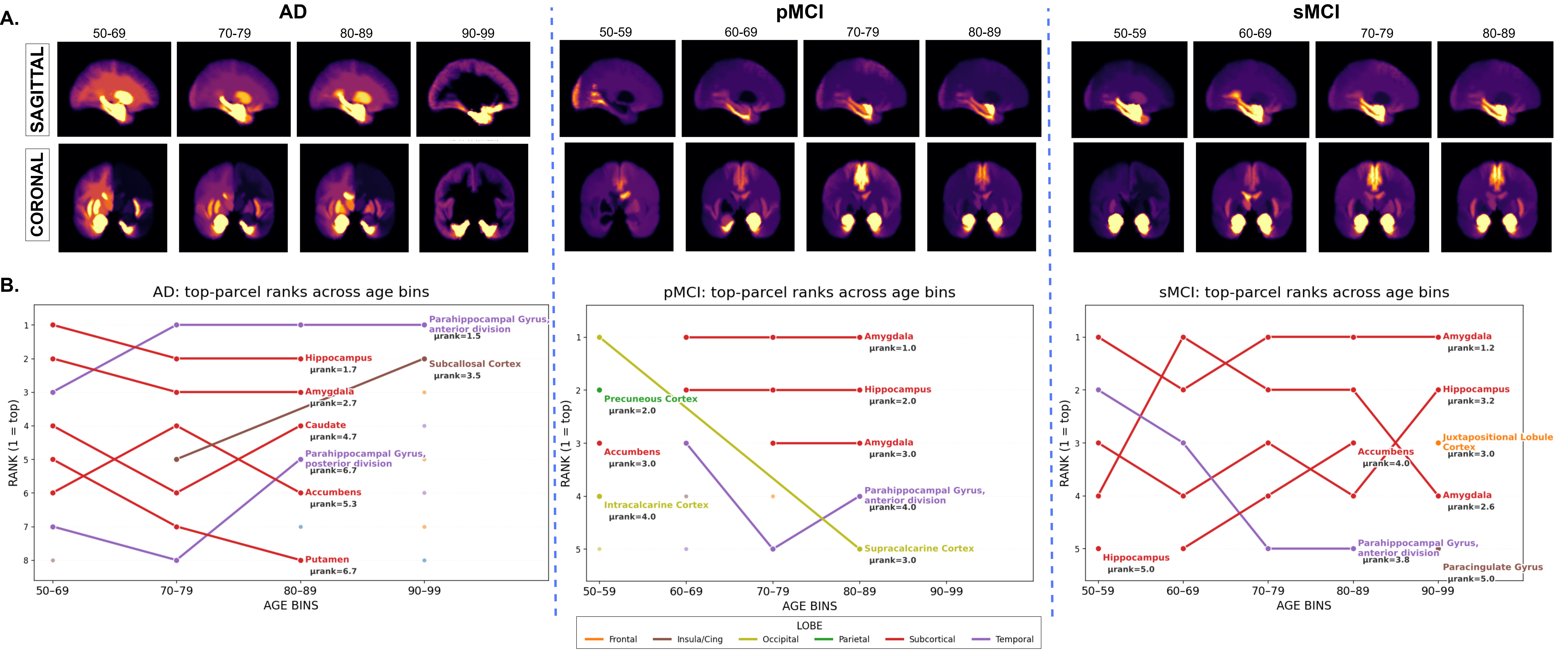}
  \vspace{-0.2em}
  \caption{\textbf{A. Cohort-level parcel saliency in AD/pMCI/sMCI subjects (test set) across age strata.}
  Each panel is a sagittal (slice:64) / coronal (slice:122) view of a \emph{voxel-space heatmap} constructed using \emph{cohort-mean parcel saliencies}. Bright regions indicate parcels that contributed most to the model’s prediction within the corresponding age bin. \textbf{B. Top-parcel rank trajectories across age bins.} The chart shows \emph{relative} importance of parcels across the age bins. Parcels, colored by lobe, are ranked within each age bin by absolute mean saliency score \(|\bar{s}|\) (1=top). \(\mu\)rank indicates the mean rank over all bins. 
  }
  \label{fig:saliency}
\end{figure*}

\noindent
\textit{\textbf{Cohort level Age-binned parcel saliencies:}}
For each test scan, we compute a normalized, non-negative parcel saliency vector \(\hat{\mathbf{s}}\in\mathbb{R}^P\) and average these across the cohort $\mathcal{I}$ within diagnosis \(d\) and age bin \(b\),
$\bar{\mathbf{s}}_{d,b}=\frac{1}{|\mathcal{I}_{d,b}|}\sum_{i\in\mathcal{I}_{d,b}}\hat{\mathbf{s}}^{(i)}.$
A sagittal and coronal view of this volume is visualized across all the bins in Fig~\ref{fig:saliency}A for AD, pMCI, and sMCI test subjects. Fig~\ref{fig:saliency}B shows top-parcel rank trajectories over age. Cohort-level explainability analysis reveals that our model's classification decisions are anchored in anatomically plausible and disease-consistent patterns. Across age strata, saliency concentrates on medial temporal structures (e.g., hippocampus, amygdala, parahippocampal regions) in AD and MCI, consistent with established AD-related atrophy patterns~\cite{Planche2022,Braak1991}.
These cohort-aggregated maps provide an auditable summary of the anatomical regions most responsible for predictions across diagnoses and age.

\paragraph{Per-scan parcel saliency:}
Given a trained model and a target class logit \(f_y\), let \(\text{scores}_p\) be the pre-softmax gate score for parcel \(p\in\{1,\ldots,P\}\).
Define parcel weights \(w_p=\operatorname{softmax}(\text{scores}/\tau)_p\). Then the gradient-based parcel saliency becomes:
\[
s_p \;=\; \mathrm{ReLU}\!\Bigl( w_p \,\cdot\, \frac{\partial f_y}{\partial \text{scores}_p} \Bigr),
\qquad
\hat{s}_p \;=\; \frac{s_p}{\sum_{q=1}^{P} s_q + \varepsilon},
\]
yielding a per-scan vector \(\hat{\mathbf{s}}\in\mathbb{R}^P\) that is nonnegative and \(L_1\)-normalized.

\vspace{-1em}
\paragraph{Age binning and cohort mean:}
Assign each scan \(i\) to an age bin \(b\) (in years). For diagnosis \(d\) and bin \(b\), compute the cohort mean
\[
\bar{s}_{d,b} \;=\; \frac{1}{|\mathcal{I}_{d,b}|}\sum_{i\in\mathcal{I}_{d,b}}\hat{s}^{(i)}.
\]

\vspace{-1em}
\paragraph{Atlas-weighted 3D maps:}
A \emph{voxel-space heatmap} constructed by linearly mixing Harvard–Oxford (HO) probabilistic parcel maps with \emph{cohort-mean parcel saliencies}. Using the Harvard--Oxford probabilistic atlas \(\{A_k(x)\}_{k=1}^{P}\) (voxel \(x\)), form the explanation volume
\[
H_{d,b}(x) \;=\; \sum_{k=1}^{P}\bar{s}_{d,b}[k]\;A_k(x).
\]
For Fig~\ref{fig:saliency}, we apply light smoothing and robust normalization (shared across panels in a row) and render sagittal/coronal views.
Within each bin \(b\), rank parcels by descending \(|\bar{s}_{d,b}[k]|\) and keep the top \(N\) with integer ranks (1=highest). 


\vspace{1em}
\subsection{Attention Rollout visualizations}
\label{Asec:attnroll-details}
\paragraph{\textbf{Context:}} In the best performing scan-level longitudinal variant, each visit $t$ produces parcel embeddings $\{z_{p,t}\}_{p=1}^P$ and post-softmax parcel scores $\alpha_{p,t}$, and a gated scan embedding
$s_t=\sum_{p=1}^P \alpha_{p,t}\, z_{p,t}$. A Longitudinal transformer then models the longitudinal sequence
$(s_1,\ldots,s_T)$.

\paragraph{\textbf{Attention Rollout heatmaps:}} For this, we compute (i) a \emph{scan-level} token importance from the Parcel encoder, (ii) a \emph{temporal} importance over visits from the Longitudinal transformer, and (iii) \emph{parcel scores} per token. For visit $i$ and token $k$:
\[
h_i[k] \;=\; r_{\text{time}}[i]\;\cdot\; r_{\text{scan}}[k]\;\cdot\;\pi_i[k]
\]
where
\begin{align*}
r_{\text{scan}}[k] &= \frac{1}{K}\sum_{u=1}^{K}
\Big(\prod_{\ell=1}^{L} \mathrm{row\_norm}(A_\ell{+}I)\Big)[u,k],\\
r_{\text{time}}[i] &= \frac{1}{t}\sum_{u=1}^{t}
\Big(\prod_{b=1}^{B} \mathrm{row\_norm}(T_b{+}I)\Big)[u,i],\\
\pi_i[k] &= \sum_{p=1}^{P} M_i[k,p]\; \alpha_{p,i},
\end{align*}

\noindent
Here, $\{A_\ell\}_{\ell=1}^{L}$ are within-scan Parcel encoder attentions over $K$ tokens, $\{T_b\}_{b=1}^{B}$ are Longitudinal Transformer's temporal attentions over $t$ visits, and $M_i\!\in\!\mathbb{R}^{K\times P}$ are the token–parcel soft memberships for visit $i$.

\begin{figure*}[t]
  \centering
 \includegraphics[width=0.8\linewidth]{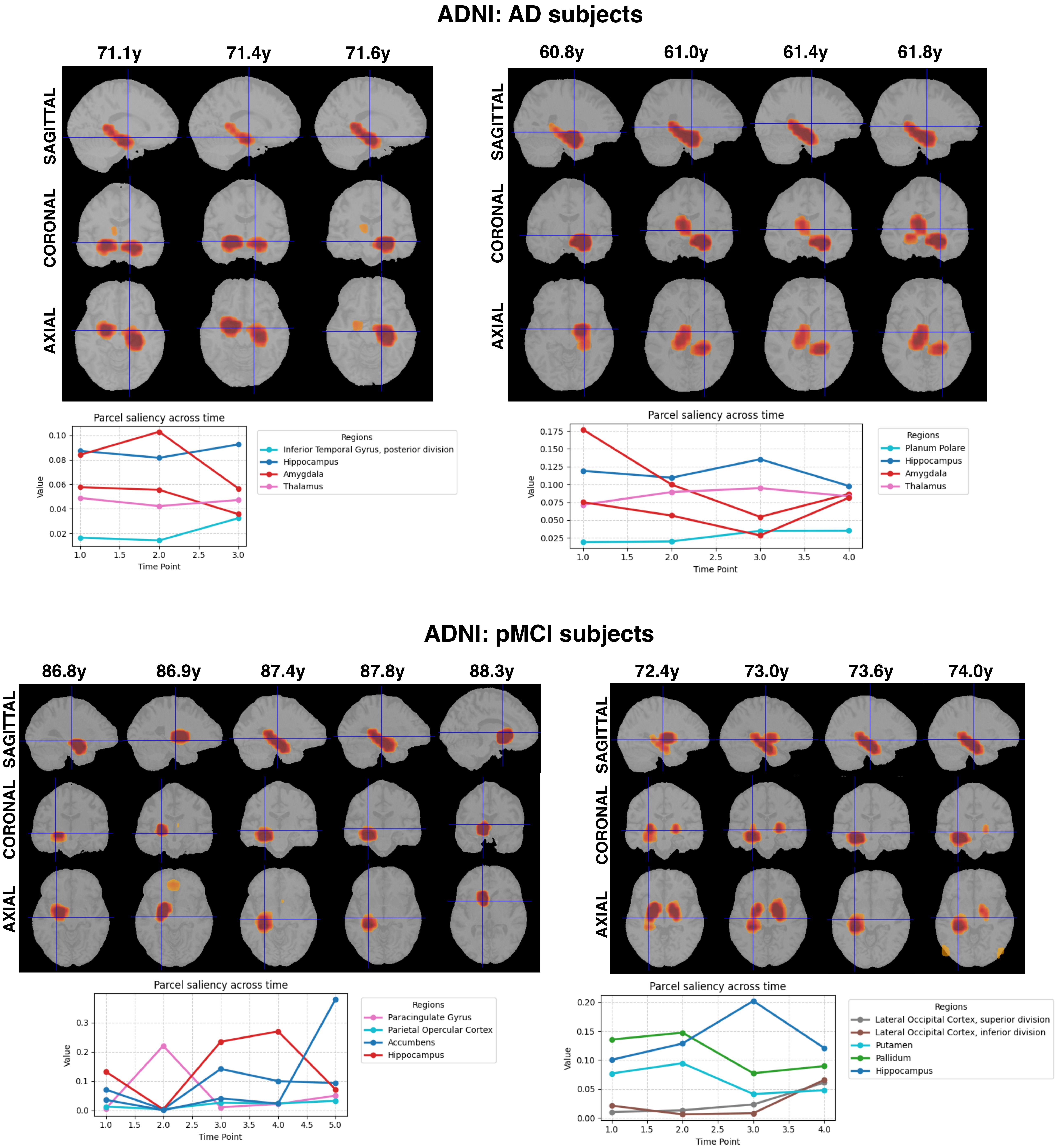}
  \caption{Attention Rollout visualizations across all three views (coronal, sagittal, and axial) along with their parcel saliency plots of correctly predicted longitudinal scans of \textbf{AD/pMCI} subjects of the ADNI datasets.}
  \vspace{-0.8em}
  \label{fig:attnroll_AD_pMCI_sMCI}
\end{figure*}

\begin{figure*}[t]
  \centering
 \includegraphics[width=0.8\linewidth]{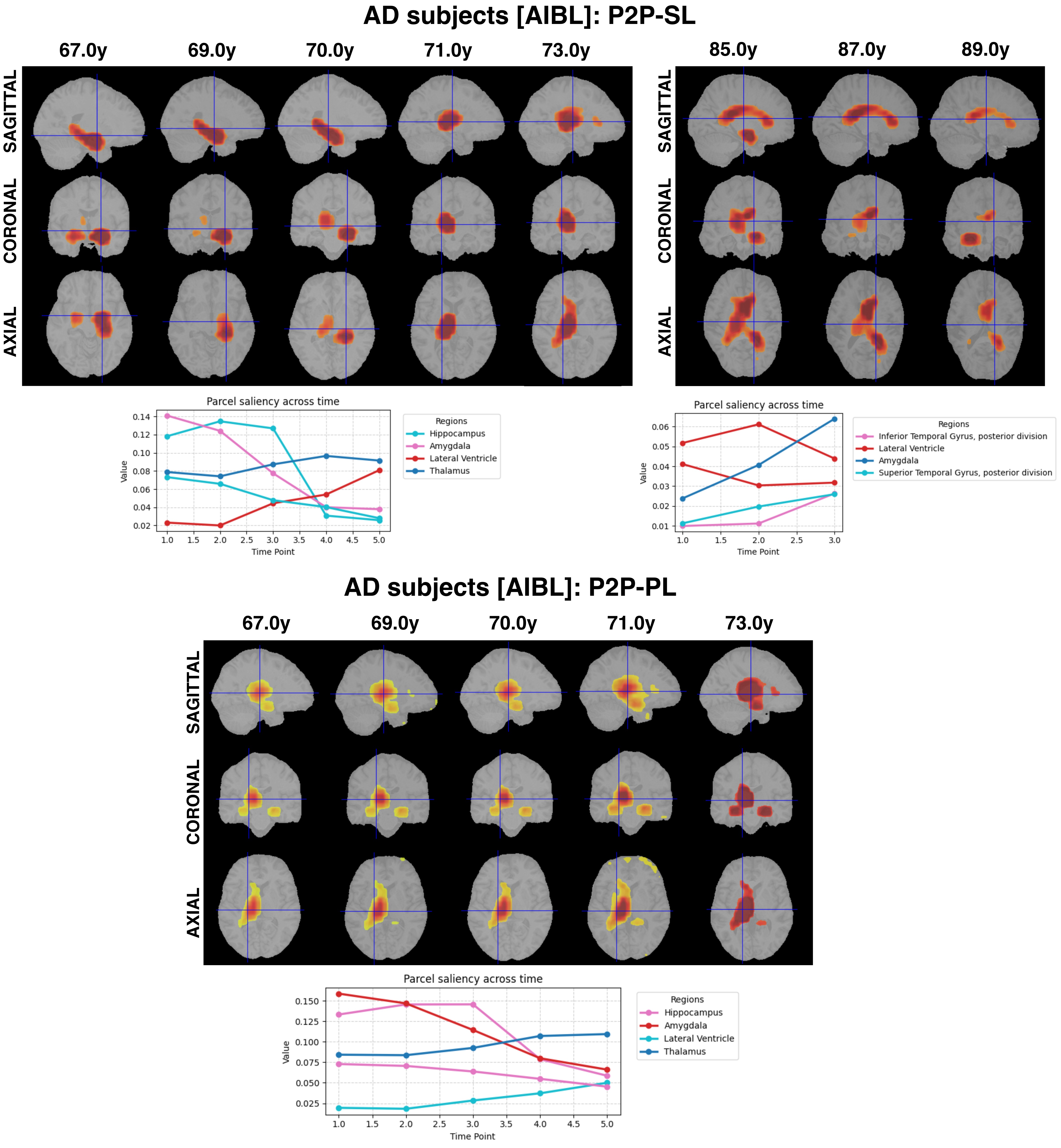}
  \caption{Attention Rollout visualizations across all three views (coronal, sagittal, and axial) along with their parcel saliency plots of correctly predicted longitudinal scans of \textbf{AD} subjects of the AIBL dataset. Attention rollouts are performed for both the variants: P2P-PL \& P2P-SL.}
  \vspace{-0.8em}
  \label{fig:attnroll_MCI_aibl_adni}
\end{figure*}

\subsection{More Explainability Analysis}
\label{supp:more-explainability}

We conduct a two-step explainability analysis spanning cohort-level aggregate attribution and subject-level longitudinal trajectories to investigate the interpretability of P2P. These complementary views assess whether the model's classification decisions are grounded in anatomically plausible, disease-consistent spatiotemporal patterns. 

\textbf{Interpreting Fig~\ref{fig:saliency}.} Cohort-level explainability analysis reveals that our model's classification decisions are anchored in anatomically plausible and disease- consistent patterns (Fig~\ref{fig:saliency}). Across age-stratified bins, gradient-based attribution concentrates persistently on medial temporal structures such as hippocampus ~\cite{Pennanen2004}, amygdala ~\cite{Poulin2011}, and parahippocampal gyrus in AD and pMCI~\cite{Pennanen2004}, directly mirroring the canonical Braak staging trajectory~\cite{Braak1991,Braak2006,Planche2022,Lam2013,blinkouskaya2021brain,Poulin2011} and established MRI volumetry findings~\cite{Braak1991,Braak2006,Jack1999,Dickerson2009}. Critically, pMCI exhibits a directional shift with age: strong limbic emphasis~\cite{Ferreira2017} in younger bins (50-69 years) transitions toward posterior association cortices (precuneus, supracalcarine )~\cite{Doody2010,Jellinger2022} in older bins (70-89 years), recapitulating the known spread of classical AD Pathlogy along with limbic region~\cite{Jack2010,Whitwell2007,Buckner2005}. In contrast, sMCI shows unstable Medial Temporal Lobe attribution (hippocampus), reflecting its heterogeneous, non-progressive etiology and providing specificity confirmation that the model discriminates AD-trajectory cases~\cite{Petersen2004,Chetelat2005}. 

\paragraph{Note for Fig~\ref{fig:saliency}:} Parcel labels besides denote parcels with high persistence (appear in many bins with large summed \(|\bar{s}|\)). Parcels appearing twice in the chart correspond to the left and right hemispheres of the Brain.\\

\textbf{Interpreting Fig~\ref{fig:attnroll}.} Subject-level longitudinal parcel attribution trajectories demonstrate coherence with cohort-level analysis. AD exhibits sustained, convergent Medial Temporal Lobe (MTL) weighting, such as amygdala and Hippocampus, rising to dominate and thalamus peaking (at timepoint 4), mirroring progressive limbic neurodegeneration and secondary degeneration~\cite{Braak2006,Poulin2011,DeJong2008}. 

pMCI displays the canonical hippocampal and Medial Temporal Gyrus, recapitulating maximal atrophy rates during MCI-to-AD conversion~\cite{Jack2004,Henneman2009}. It is followed by posterior/lateral cortical parcels (intracalcarine) progressively reflecting disease progression~\cite{Whitwell2008,Buckner2005}. 
In contrast, sMCI shows high-amplitude oscillations without stable rankings, i.e, hippocampus fluctuates, striatal parcels intermittently dominate, thus reflecting etiological heterogeneity~\cite{Petersen2004,Stern2012}. These distinct dynamics, AD's confident convergence, pMCI's sequential progression tracking, and sMCI's exploratory heterogeneity, reflect that models' behaviour replicate the findings of the extant pathophysiological trajectory of AD.

\paragraph{More analysis:}
We visualize attention rollouts for representative AD/pMCI subjects from ADNI in Fig~\ref{fig:attnroll_AD_pMCI_sMCI} and AD subjects from AIBL and MIRIAD in Fig~\ref{fig:attnroll_MCI_aibl_adni} using the P2P-SL variant. Also, attention rollouts of P2P-$\text{PL}$ variant which performed best with AIBL is visualized in Fig~\ref{fig:attnroll_MCI_aibl_adni}. Additionally, we also provide their respective parcel saliency plots. These complementary views show that model routing (attention rollouts) and model reliance (saliency) are largely co-localized in anatomically plausible regions. 



\paragraph{ADNI analysis (Fig~\ref{fig:attnroll_AD_pMCI_sMCI}):}
In AD cases, P2P consistently allocates saliency to canonical medial temporal hubs such as hippocampus, amygdala, and posterior inferior temporal cortex, mirroring established staging and reinforcing that the framework recovers core loci of Alzheimer’s pathology in a visit-consistent manner. In pMCI, the trajectories emphasize limbic and posterior association cortices whose evolving engagement is characteristic of elevated conversion risk. Notably, older converters often exhibit progressive involvement of paracingulate and parietal opercular regions, whereas younger converters show a complementary posterior–subcortical pattern involving lateral occipital parcels and limbic nuclei. 


\paragraph{AIBL/MIRIAD analysis:}
In \textsc{AIBL} (Fig~\ref{fig:attnroll_MCI_aibl_adni}), hippocampal saliency remains persistently high with increasing emphasis on hippocampus, amygdala, and lateral ventricles, the atrophy-consistent signals, whereas \textsc{MIRIAD} (Fig~\ref{fig:attnroll_MCI_aibl_adni}) subject frequently accentuates the posterior division of the parahippocampal gyrus. In Fig~\ref{fig:attnroll_MCI_aibl_adni}, which visualizes the P2P-$\text{PL}$ variant's attention rollouts, clinically aligned parcel activations (i.e., hippocampus, Amygdala, Lateral ventricle) are visible across the scans of all time points, but with increasing focus towards the latest scans. Together, these visualizations indicate that P2P learns physiologically meaningful spatiotemporal signatures rather than relying on confounds or scan-specific artifacts.

\begin{wrapfigure}{r}{0.35\linewidth}
    \centering
    \vspace{-1.5em}
    \includegraphics[width=\linewidth]{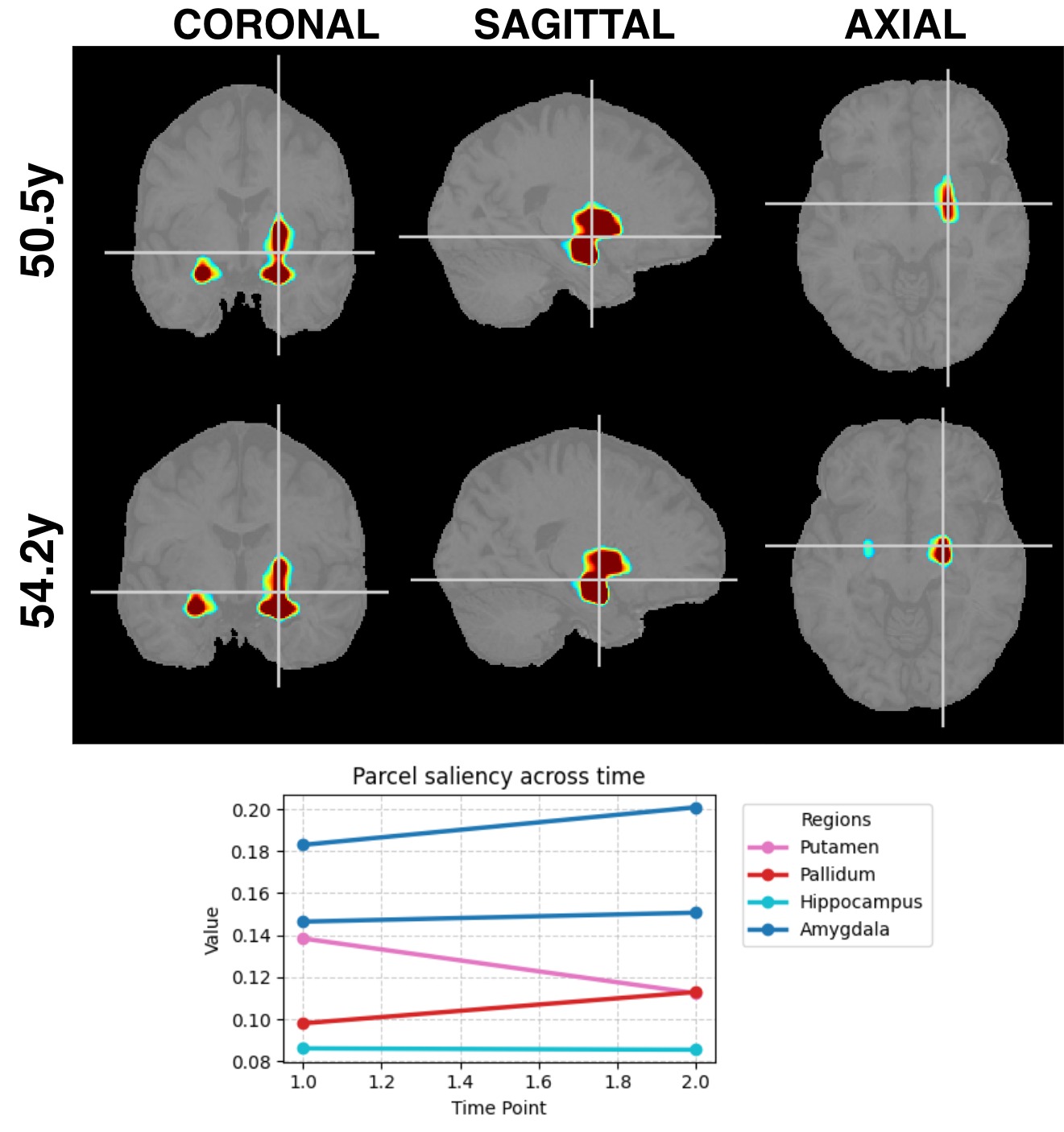}
    \vspace{-1.4em}
    \caption{Attention rollout visualizations across the coronal, sagittal, and
    axial views, together with parcel-saliency plots, for a misclassified ADNI
    subject.}
    \label{fig:failure}
    
\end{wrapfigure}
\paragraph{Failure Case:}
Fig~\ref{fig:failure} shows a failure case of a misclassified CN subject predicted as AD. This indicates that the false-positive prediction was driven by a consistent pattern of medial-temporal and subcortical evidence that the model interpreted
as AD-like, rather than by unstable or spatially scattered attention.

Critically, across all cohorts, variants, and diagnostic categories, attention rollouts and gradient saliency maps are largely co-localized in anatomically plausible regions. Since rollouts characterize where the network routes information and saliency characterizes what the prediction depends upon, their spatial agreement constitutes a high-faithfulness criterion: the model attends to the right regions and relies upon them for the right reasons. These converging lines of evidence, taken together, demonstrate that P2P has learned physiologically meaningful spatiotemporal signatures of neurodegeneration, distinguishing AD, pMCI, and sMCI through structurally distinct, temporally coherent attribution dynamics that replicate established neuropathological trajectories without explicit anatomical supervision.

\clearpage
\newpage


\end{document}